\documentstyle[12pt,epsf,fullname]{book}

\newtheorem{theorem}{Theorem}
\newtheorem{proposition}{Proposition}
\newcommand{\tab}{\hspace{5mm}}

\begin{document}


\thispagestyle{empty}

\ \ 

\vspace{30mm}

\begin{center}
{\LARGE A Probabilistic Approach to Lexical Semantic Knowledge Acquisition
and Structural Disambiguation}
\end{center}

\vspace{5mm}

\begin{center}
{\Large
Hang LI} \\
\end{center}

\vspace{50mm}

\begin{center}
{\large A Dissertation \\
Submitted to the Graduate School of Science \\
of the University of Tokyo \\
in Partial Fulfillment of the Requirements \\
for the Degree of Doctor of Science \\
in Information Science}
\end{center}

\begin{center}
{\large July 1998}
\end{center}

\begin{center}
\copyright 1998 by Hang LI\\
All rights reserved.
\end{center}

\newpage
\pagenumbering{roman}

\addcontentsline{toc}{chapter}{\numberline{}Abstract}

\chapter*{Abstract}

Structural disambiguation in sentence analysis is still a central
problem in natural language processing. Past researches have verified
that using lexical semantic knowledge can, to a quite large extent,
cope with this problem. Although there have been many studies
conducted in the past to address the lexical knowledge acquisition
problem, further investigation, especially that based on a {\em
principled methodology} is still needed, and this is, in fact, the
problem I address in this thesis.

The problem of acquiring and using lexical semantic knowledge,
especially that of case frame patterns, can be formalized as follows. 
A learning module acquires case frame patterns on the basis of some
case frame instances extracted from corpus data. A processing
(disambiguation) module then refers to the acquired knowledge and
judges the degrees of acceptability of some number of new case frames,
including {\em previously unseen} ones.

The approach I adopt has the following characteristics: (1) dividing
the problem into three subproblems: case slot generalization, case
dependency learning, and word clustering (thesaurus construction). (2)
viewing each subproblem as that of statistical estimation and defining
probability models for each subproblem, (3) adopting the Minimum
Description Length (MDL) principle as learning strategy, (4) employing
efficient learning algorithms, and (5) viewing the disambiguation
problem as that of statistical prediction.

The need to divide the problem into subproblems is due to the
complicatedness of this task, i.e., there are too many relevant
factors simply to incorporate all of them into a single model. The use
of MDL here leads us to a theoretically sound solution to the `data
sparseness problem,' the main difficulty in a statistical approach to
language processing.

In Chapter 3, I define probability models for each subproblem: (1) the
hard case slot model and the soft case slot model; (2) the word-based
case frame model, the class-based case frame model, and the slot-based
case frame model; and (3) the hard co-occurrence model and the soft
co-occurrence model. These are respectively the probability models for
(1) case slot generalization, (2) case dependency learning, and (3)
word clustering. Here the term `hard' means that the model is
characterized by a type of word clustering in which a word can only
belong to a single class alone, while `soft' means that the model is
characterized by a type of word clustering in which a word can belong
to several different classes.

In Chapter 4, I describe one method for learning the hard case slot
model, i.e., generalizing case slots. I restrict the class of hard
case slot models to that of tree cut models by using an existing
thesaurus. In this way, the problem of generalizing the values of a
case slot turns out to be that of estimating a model from the class of
tree cut models for some fixed thesaurus tree. I then employ an
efficient algorithm, which provably obtains the optimal tree cut model
in terms of MDL. This method, in fact, conducts generalization in the
following way. When the differences between the frequencies of the
nouns in a class are not large enough (relative to the entire data
size and the number of the nouns), it generalizes them into the class.
When the differences are especially noticeable, on the other hand, it
stops generalization at that level.

In Chapter 5, I describe one method for learning the case frame model,
i.e., learning dependencies between case slots. I restrict the class
of case frame models to that of dependency forest models. Case frame
patterns can then be represented as a dependency forest, whose nodes
represent case slots and whose directed links represent the
dependencies that exist between these case slots. I employ an
efficient algorithm to learn the optimal dependency forest model in
terms of MDL. This method first calculates a statistic between all
node pairs and sorts these node pairs in descending order with respect
to the statistic. It then puts a link between the node pair highest in
the order, provided that this value is larger than zero. It repeats
this process until no node pair is left unprocessed, provided that
adding that link will not create a loop in the current dependency
graph.

In Chapter 6, I describe one method for learning the hard
co-occurrence model, i.e., automatically conducting word clustering. I
employ an efficient algorithm to repeatedly estimate a suboptimal MDL
model from a class of hard co-occurrence models. The clustering method
iteratively merges, for example, noun classes and verb classes in
turn, in a bottom up fashion. For each merge it performs, it
calculates the decrease in empirical mutual information resulting from
merging any noun (or verb) class pair, and performs the merge having
the least reduction in mutual information, provided that this
reduction in mutual information is less than a threshold, which will
vary depending on the data size and the number of classes in the
current situation.

In Chapter 7, I propose, for resolving ambiguities, a new method which
combines the use of the hard co-occurrence model and that of the tree
cut model. In the implementation of this method, the learning module
combines with the hard co-occurrence model to cluster words with
respect to each case slot, and it combines with the tree cut model for
generalizing the values of each case slot by means of a hand-made
thesaurus. The disambiguation module first calculates a likelihood
value for each interpretation on the basis of hard co-occurrence
models and outputs the interpretation with the largest likelihood
value; if the likelihood values are equal (most particularly, if all
of them are 0), it uses likelihood values calculated on the basis of
tree cut models; if the likelihood values are still equal, it makes a
default decision.

The accuracy achieved by this method is $85.2\%$, which is higher than
that of state-of-the-art methods.

\newpage
\addcontentsline{toc}{chapter}{\numberline{}Acknowledgements}

\chapter*{Acknowledgements}

I would like to express my sincere appreciation to my supervisor,
Prof.~Jun'ichi Tsujii of the University of Tokyo, for his continuous
encouragement and guidance. It was Prof.~Tsujii who guided me in the
fundamentals of natural language processing when I was an
undergraduate student at Kyoto University. His many helpful
suggestions and comments have also been crucial to the completion of
this thesis.

I would also like to express my gratitude to the members of my
dissertation committee: Prof.~Toshihisa Takagi and Prof.~Hiroshi Imai
of the University of Tokyo, Prof.~Yuji Matsumoto of Nara Institute of
Science and Technology (NAIST), and Prof.~Kenji Kita of Tokushima
University, who have been good enough to give this work a very serious
review.

Very special thanks are also due to Prof.~Makoto Nagao of Kyoto
University for his encouragement and guidance, particularly in his
supervision of my master thesis when I was a graduate student at Kyoto
University I learned a lot from him, especially in the skills of
conducting research. He is one of the persons who continue to
influence me strongly in my research carrer, even though the approach
I am taking right now is quite different from his own.

I also would like to express my sincere gratitude to Prof.~Yuji
Matsumoto. He has given me much helpful advice with regard to the
conduct of research, both when I was at Kyoto University and after I
left there. The use of dependency graphs for representation of case
frame patterns was inspired by one of his statements in a personal
conversation.

I would like to thank Prof.~Jun'ichi Nakamura of Kyoto University,
Prof.~Satoshi Sato of the Japan Advance Institute of Science and
Technology, and other members of the Nagao Lab. for their advice and
contributions to our discussions.

The research reported in this dissertation was conducted at C\&C Media
Research Laboratories, NEC Corporation and the Theory NEC Laboratory,
Real World Computing Partnership (RWCP). I would like to express my
sincere appreciation to Mr.~Katsuhiro Nakamura, Mr.~Tomoyuki Fujita,
and Dr.~Shun Doi of NEC.  Without their continuous encouragement and
support, I would not have been able to complete this work.

Sincere appreciation also goes to Naoki Abe of NEC. Most of the
research reported in this thesis was conducted jointly with
him. Without his advice and proposals, I would not have been able to
achieve the results represented here. Ideas for the tree cut model and
the hard co-occurrence model came out in discussions with him, and the
algorithm `Find-MDL' was devised on the basis of one of his ideas.

I am deeply appreciative of the encouragement and advice given me by
Kenji Yamanishi of NEC, who introduced me to the MDL principle; this
was to become the most important stimulus to the idea of conducting
this research. He also introduced me to many useful machine learning
techniques, that have broadened my outlook toward the field.

I also thank Jun-ichi Takeuchi, Atsuyoshi Nakamura, and Hiroshi
Mamitsuka of NEC for their helpful advice and suggestions. Jun-ichi 's
introduction to me of the work of Joe Suzuki eventually leads to the
development in this study of the case-dependency learning method.

Special thanks are also due to Yuuko Yamaguchi and Takeshi Futagami of
NIS who implemented the programs of Find-MDL, 2D-Clustering.

In expressing my appreciation to Yasuharu Den of NAIST, David Carter
of Speech Machines, and Takayoshi Ochiai of NIS, I would like them to
know how much I had enjoyed the enlightening conversations I had with
them.

I am also grateful to Prof.~Mark Petersen of Meiji University for what
he has taught me about the technical writing of English.
Prof.~Petersen also helped correct the English of the text in this
thesis, and without his help, it would be neither so readable nor so
precise.

Yasuharu Den, Kenji Yamanishi, David Carter, and Diana McCarthy of
Sussex University read some or all of this thesis and made many
helpful comments. Thanks also go to all of them, though the
responsibility for flaws and errors it contains remains entirely with
me.

I owe a great many thanks to many people who were kind enough to help
me over the course of this work. I would like to express here my great
appreciation to all of them.

Finally, I also would like to express a deep debt of gratitude to my
parents, who instilled in me a love for learning and thinking, and to
my wife Hsiao-ya, for her constant encouragement and support.

\newpage
\tableofcontents

\listoftables

\listoffigures

\newpage
\pagenumbering{arabic}

\chapter{Introduction}

\begin{tabular}{p{5.5cm}r}
 &
\begin{minipage}{10cm}
\begin{tabular}{p{9cm}}
{\em ... to divide each of the difficulties
under examination into as many parts as
possible, and as might be necessary for
its adequate solution.} \\
\multicolumn{1}{r}{- Ren\'{e} Descartes} \\
\end{tabular}
\end{minipage}
\end{tabular}
\vspace{0.5cm}

\section{Motivation}

Structural (or syntactic) disambiguation in sentence analysis is still
a central problem in natural language processing. To resolve
ambiguities completely, we would need to construct a human language
`understanding' system \cite{Johnson-Laird83,Tsujii87,Altmann88}. The
construction of such a system would be extremely difficult, however,
if not impossible. For example, when analyzing the sentence
\begin{equation}\label{eq:icecream} \mbox{I ate ice cream with a
spoon,} \end{equation} a natural language processing system may obtain
two interpretations: ``I ate ice cream using a spoon'' and ``I ate ice
cream and a spoon.'' i.e., a pp-attachment ambiguity may arise,
because the prepositional phrase `with a spoon' can {\em
  syntactically} be attached to both `eat' and `ice cream.' If a human
speaker reads the same sentence, common sense will certainly lead him
to assume the former interpretation over the latter, because he
understands that: ``a spoon is a tool for eating food,'' ``a spoon is
not edible,'' etc.  Incorporating such `world knowledge' into a
natural language processing system is highly difficult, however,
because of its sheer enormity.

An alternative approach is to make use of only lexical semantic
knowledge, specifically case frame patterns \cite{Fillmore68} (or
their near equivalents: selectional patterns \cite{Katz63}, and
subcategorization patterns \cite{Pollard87}). That is, to represent
the content of a sentence or a phrase with a `case frame' having a
`head'\footnote{I slightly abuse terminology here, as
  `head' is usually used for subcategorization patterns in the
discipline of HPSG, but not in case
  frame theory.} and multiple `slots,' and to incorporate into a
natural language processing system the knowledge of which words can
fill into which slot of a case frame.

For example, we can represent the sentence ``I ate ice cream'' as \[
\mbox{(eat (arg1 I) (arg2 ice-cream)),} \] where the head is `eat,'
the arg1 slot represents the subject and the arg2 slot represents the
direct object. The values of the arg1 slot and the arg2 slot are `I'
and `ice cream,' respectively. Furthermore, we can incorporate as the
case frame patterns for the verb `eat' the knowledge that a member of
the word class $\langle$animal$\rangle$ can be the value of the arg1
slot and a member of the word class $\langle$food$\rangle$ can be the
value of the arg2 slot, etc.

The case frames of the two interpretations obtained in the analysis of
the above sentence (\ref{eq:icecream}), then, become
\[
\begin{array}{l}
\mbox{(eat (arg1 I) (arg2 ice-cream) (with spoon))} \\
\mbox{(eat (arg1 I) (arg2 (ice-cream (with spoon)))).} \\
\end{array}
\] 
Referring to the case frame patterns indicating that `spoon' can be
the value of the `with' slot when the head is `eat,' and `spoon'
cannot be the value of the `with' slot when the head is `ice cream,' a
natural language processing system naturally selects the former
interpretation and thus resolves the ambiguity.

Previous data analyses have indeed indicated that using lexical
semantic knowledge can, to a quite large extent, cope with the
structural disambiguation problem \cite{Hobbs90,Whittemore90}. The
advantage of the use of lexical knowledge over that of world knowledge
is the relative smallness of its amount. By restricting knowledge to
that of {\em relations between words}, the construction of a natural
language processing system becomes much easier. (Although the lexical
knowledge is still unable to resolve the problem completely, past
research suggests that it might be the most realistic path we can take
right now.)

As is made clear in the above example, case frame patterns mainly
include `generalized information,' e.g., that a member of the word
class $\langle$animal$\rangle$ can be the value of the arg2 slot for
the verb `eat.'

Classically, case frame patterns are represented by `selectional
restrictions' \cite{Katz63}, i.e., discretely represented by semantic
features, but it is better to represent them continuously, because a
word can be the value of a slot to a certain probabilistic {\em
degree}, as is suggested by the following list \cite{Resnik93b}: \[
\begin{array}{ll} (1) & \mbox{Mary drank some wine.} \\ (2) &
\mbox{Mary drank some gasoline.} \\ (3) & \mbox{Mary drank some
pencils.} \\ (4) & \mbox{Mary drank some sadness.} \\ \end{array} \]

Furthermore, case frame patterns are not limited to reference to
individual case slots. Dependencies between case slots need also be
considered. The term `dependency' here refers to the relationship that
may exist between case slots and that indicates strong co-occurrence
between the values of those case slots.  For example, consider the
following sentences:\footnote{`*' indicates an
  unacceptable natural language expression.}
\begin{equation}\label{eq:fly}
\begin{array}{ll}
(1) & \mbox{She flies jets.} \\
(2) & \mbox{That airline company flies jets.} \\
(3) & \mbox{She flies Japan Airlines.} \\
(4) & \mbox{*That airline company flies Japan Airlines.} \\
\end{array}
\end{equation}

We see that an `airline company' can be the value of the arg1 slot,
when the value of the arg2 slot is an `airplane' but not when it is an
`airline company.'  These sentences indicate that the possible values
of case slots depend in general on those of others: dependencies
between case slots exist.\footnote{One may argue that `fly' has
different word senses in these sentences and for each of these word
senses there is no dependency between the case slots. Word senses are
in general difficult to define precisely, however. I think that it is
preferable not to resolve them until doing so is necessary in a
particular application. That is to say that, in general, case
dependencies do exist and the development of a method for learning
them is needed.}

Another consensus on lexical semantic knowledge in recent studies is
that it is preferable to learn lexical knowledge automatically from
corpus data. Automatic acquisition of lexical knowledge has the merits
of (1) saving the cost of defining knowledge by hand, (2) doing away
with the subjectivity inherent in human-defined knowledge, and (3)
making it easier to adapt a natural language processing system to a
new domain.

Although there have been many studies conducted in the past (described
here in Chapter 2) to address the lexical knowledge acquisition
problem, further investigation, especially that based on a {\em
principled methodology} is still needed, and this is, in fact, the
problem I address in this thesis.

The search for a mathematical formalism for lexical knowledge
acquisition is not only motivated by concern for logical niceties; I
believe that it can help to better cope with practical problems (for
example, the disambiguation problem). The ultimate outcome of the
investigations in this thesis, therefore, should be a formalism of
lexical knowledge acquisition and at the same time a high-performance
disambiguation method.

\section{Problem Setting}

The problem of acquiring and using lexical semantic knowledge,
especially that of case frame patterns, can be formalized as
follows. A learning module acquires case frame patterns on the basis
of some case frame instances extracted from corpus data. A processing
(disambiguation) module then refers to the acquired knowledge and
judges the degrees of acceptability of some new case frames, including
{\em previously unseen} ones. The goals of learning are to represent
more {\em compactly} the given case frames, and to judge more {\em
correctly} the degrees of acceptability of new case frames.

In this thesis, I propose a probabilistic approach to lexical knowledge
acquisition and structural disambiguation.

\section{Approach}

In general, a machine learning process consists of three elements:
model, strategy (criterion), and algorithm.  That is, when we conduct
machine learning, we need consider (1) what kind of model we are to
use to represent the problem, (2) what kind of strategy we should
adopt to control the learning process, and (3) what kind of algorithm
we should employ to perform the learning task. We need to consider
each of these elements here.

\subsubsection*{Division into subproblems} The lexical semantic
knowledge acquisition problem is a quite complicated task, and there
are too many relevant factors (generalization of case slot values,
dependencies between case slots, etc.) to simply incorporate all of
them into a single model. As a first step, I divide the problem into
three subproblems: case slot generalization, case dependency learning,
and word clustering (thesaurus construction).

I define probability models (probability distributions) for each
subproblem and view the learning task of each subproblem as that of
estimating its corresponding probability models based on corpus data.

\subsubsection*{Probability models} 

We can assume that case slot data for a case slot for a verb are
generated on the basis of a conditional probability distribution that
specifies the conditional probability of a noun given the verb and the
case slot. I call such a distribution a `case slot model.' When the
conditional probability of a noun is defined as the conditional
probability of the noun class to which the noun belongs, divided by
the size of the noun class, I call the case slot model a `hard case
slot model.' When the case slot model is defined as a finite mixture
model, namely a linear combination of the word probability
distributions within individual noun classes, I call it a `soft case
slot model.' 

Here the term `hard' means that the model is characterized by a type
of word clustering in which a word can only belong to a single class
alone, while `soft' means that the model is characterized by a type of
word clustering in which a word can belong to several different
classes.

I formalize the problem of generalizing the values of a case slot as
that of estimating a hard (or soft) case slot model. The
generalization problem, then, turns out to be that of selecting a
model, from a class of hard (or soft) case slot models, which is most
likely to have given rise to the case slot data.

We can assume that case frame data for a verb are generated according
to a multi-dimensional joint probability distribution over random
variables that represent the case slots. I call the distribution a
`case frame model.' I further classify this case frame model into
three types of probability models each reflecting the type of its
random variables: the `word-based case frame model,' the `class-based
case frame model,' and the `slot-based case frame model.'

I formalize the problem of learning dependencies between case slots as
that of estimating a case frame model. The dependencies between case
slots are represented as probabilistic dependencies between random
variables.

We can assume that co-occurrence data for nouns and verbs with respect
to a slot are generated based on a joint probability distribution that
specifies the co-occurrence probabilities of noun verb pairs. I call
such a distribution a `co-occurrence model.' I call this co-occurrence
model a `hard co-occurrence model,' when the joint probability of a
noun verb pair is defined as the product of the following three
elements: (1) the joint probability of the noun class and the verb
class to which the noun and the verb respectively belong, (2) the
conditional probability of the noun given its noun class, and (3)
the conditional probability of the verb given its verb class. When
the co-occurrence model is defined as a double mixture model, namely,
a double linear combination of the word probability distributions
within individual noun classes and those within individual verb
classes, I call it a `soft co-occurrence model.'

I formalize the problem of clustering words as that of estimating a
hard (or soft) co-occurrence model. The clustering problem, then,
turns out to be that of selecting a model from a class of hard (or
soft) co-occurrence models, which is most likely to have given rise to
the co-occurrence data.

\subsubsection*{MDL as strategy} For all subproblems, the learning
task turns out to be that of selecting the best model from among a
class of models. The question now is what the learning strategy (or
criterion) is to be. I employ here the Minimum Description Length
(MDL) principle.  The MDL principle is a principle for both data
compression and statistical estimation (described in Chapter 2).

MDL provides a theoretically way to deal with the `data sparseness
problem,' the main difficulty in a statistical approach to language
processing. At the same time, MDL leads us to an information-theoretic
solution to the lexical knowledge acquisition problem, in which case
frames are viewed as {\em structured data}, and the learning process
turns out to be that of {\em data compression}.

\subsubsection*{Efficient algorithms} In general, there is a trade-off
between model classes and algorithms.  A complicated model class would
be precise enough for representing a problem, but it might be
difficult to learn in terms of learning accuracy and computation time. 
In contrast, a simple model class might be easy to learn, but it would
be too simplistic for representing a problem.

In this thesis, I place emphasis on efficiency and restrict a model
class when doing so is still reasonable for representing the problem
at hand.

For the case slot generalization problem, I make use of an existing
thesaurus and restrict the class of hard case slot models to that of
`tree cut models.'  I also employ an efficient algorithm, which
provably obtains the optimal tree cut model in terms of MDL.

For the case dependency learning problem, I restrict the class of
case frame models to that of `dependency forest models,' and employ
another efficient algorithm to learn the optimal dependency forest
model in terms of MDL.

For the word clustering problem, I address the issue of
estimating the hard co-occurrence model, and employ an efficient
algorithm to repeatedly estimate a suboptimal MDL model from a class
of hard co-occurrence models.

\subsubsection*{Disambiguation methods}

I then view the structural disambiguation problem as that of {\em
statistical prediction}. Specifically, the likelihood value of each
interpretation (case frame) is calculated on the basis of the above
models, and the interpretation with the largest likelihood value is
output as the analysis result.

I have devised several disambiguation methods along this line.

One of them is especially useful when the data size for training is at
the level of that currently available. In implementation of this
method, the learning module combines with the hard co-occurrence model
to cluster words with respect to each case slot, and it combines with
the tree cut model to generalize the values of each case slot by means
of a hand-made thesaurus. The disambiguation module first calculates a
likelihood value for each interpretation on the basis of hard
co-occurrence models and outputs the interpretation with the largest
likelihood value; if the likelihood values are equal (most
particularly, if all of them are 0), it uses likelihood values
calculated on the basis of tree cut models; if the likelihood values
are still equal, it makes a default decision.

The accuracy achieved by this method is $85.2\%$, which is higher than
that of state-of-the-art methods.

\section{Organization of the Thesis}

\begin{figure}[htb]
\begin{center}
\epsfxsize8cm\epsfysize8cm\epsfbox{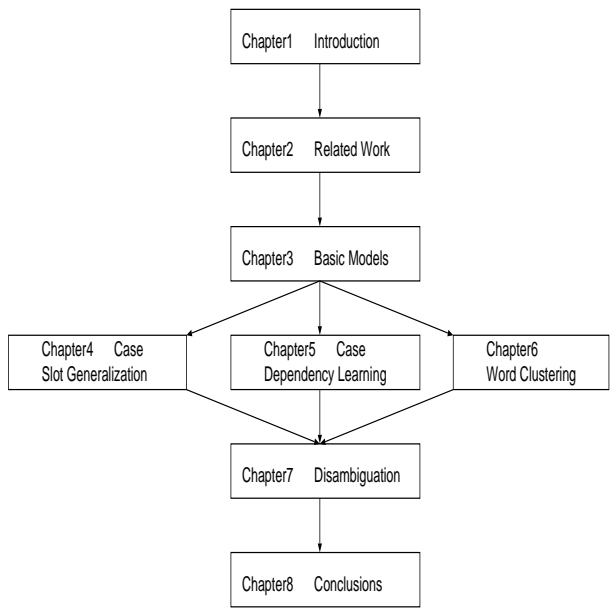}
\end{center}
\caption{Organization of this thesis.}
\label{fig:thesisorg} 
\end{figure}

This thesis is organized as follows. In Chapter 2, I review previous
work on lexical semantic knowledge acquisition and structural
disambiguation. I also introduce the MDL principle. In Chapter 3, I
define probability models for each subproblem of lexical semantic
knowledge acquisition. In Chapter 4, I describe the method of using
the tree cut model to generalize case slots. In Chapter 5, I describe
the method of using the dependency forest model to learn dependencies
between case slots. In Chapter 6, I describe the method of using the
hard co-occurrence model to conduct word clustering. In
Chapter 7, I describe the practical disambiguation method. In Chapter
8, I conclude the thesis with some remarks (see
Figure~\ref{fig:thesisorg}).

\chapter{Related Work}

\begin{tabular}{p{5.5cm}r}
 &
\begin{minipage}{10cm}
\begin{tabular}{p{9cm}}
{\em Continue to cherish old knowledge so as to continue to
  discover new.}\\
\multicolumn{1}{r}{- Confucius} \\
\end{tabular}
\end{minipage}
\end{tabular}
\vspace{0.5cm}

In this chapter, I review previous work on lexical knowledge
acquisition and disambiguation. I also introduce the MDL principle.

\section{Extraction of Case Frames}

Extracting case frame instances {\em automatically} from corpus data
is a difficult task, because when conducting extraction, ambiguities
may arise, and we need to exploit lexical semantic knowledge to
resolve them. Since our goal of extraction is indeed to acquire such
knowledge, we are faced with the problem of which is to come first,
the chicken or the egg.

Although there have been many methods proposed to automatically
extract case frames from corpus data, their accuracies do not seem
completely satisfactory, and the problem still needs investigation.

\namecite{Manning92}, for example, proposes extracting case frames by
using a finite state parser. His method first uses a statistical
tagger (cf.,
\cite{Church88,Kupiec92,Charniak93,Merialdo94,Nagata94,Schutze94,Brill95,Samuelsson95,Ratnaparkhi96,Haruno97})
to assign a part of speech to each word in the sentences of a
corpus. It then uses the finite state parser to parse the sentences
and note case frames following verbs. Finally, it filters out
statistically unreliable extracted results on the basis of hypothesis
testing (see also
\cite{Brent91,Brent93,Smadja93,Chen94,Grefenstette94}).

\namecite{Briscoe97} extracted case frames by using a probabilistic LR
parser. This parser first parses sentences to obtain analyses with
`shallow' phrase structures, and assigns a likelihood value to each
analysis. An extractor then extracts case frames from the most likely
analyses (see also \cite{Hindle91,Grishman92}).

\namecite{Utsuro92} propose extracting case frames from a
parallel corpus in two different languages. Exploiting the fact that a
syntactic ambiguity found in one language may not exist at all in
another language, they conduct pattern matching between case frames of
translation pairs given in the corpus and choose the best matched case
frames as extraction results (see also \cite{Matsumoto93}).

An alternative to the automatic approach is to employ a semi-automatic
method, which can provide much more reliable results. The
disadvantage, however, is its requirement of having disambiguation
decisions made by a human, and how to reduce the cost of human
intervention becomes an important issue. 

\namecite{Carter97} developed an interaction system for effectively
collecting case frames semi-automatically. This system first presents
a user with a range of properties that may help resolve ambiguities in
a sentence.  The user then designates the value of one of the
properties, the system discards those interpretations which are
inconsistent with the designation, and it re-displays only the
properties which remain. After several such interactions, the system
obtains a most likely correct case frame of a sentence (see also
\cite{Marcus93}).

Using any one of the methods, we can extract case frame instances for
a verb, to obtain data like that shown in Table~\ref{tab:cfdata1},
although no method guarantees that the extracted results are
completely correct. In this thesis, I refer to this type of data as
`case frame data.' If we restrict our attention on a specific slot,
we obtain data like that shown in Table~\ref{tab:csdata1}. I refer to
this type of data as `case slot data.'

\begin{table}[htb]
\caption{Example case frame data.}
\label{tab:cfdata1}
\begin{center}
\begin{tabular}{|l|} \hline
(fly (arg1 girl)(arg2 jet)) \\
(fly (arg1 company)(arg2 jet)) \\
(fly (arg1 girl)(arg2 company)) \\ \hline
\end{tabular}
\end{center}
\end{table}

\begin{table}[htb]
\caption{Example case slot data.}
\label{tab:csdata1}
\begin{center}
\begin{tabular}{|lcc|} \hline
Verb & Slot name & Slot value \\ \hline
fly & arg1 & girl \\
fly & arg1 & company \\
fly & arg1 & girl \\ \hline
\end{tabular}
\end{center}
\end{table}

\section{Case Slot Generalization}

One case-frame-pattern acquisition problem is that of generalization
of (values of) case slots; this has been intensively investigated in
the past.

\subsection{Word-based approach and the data sparseness problem}

\begin{figure}[htb]
\begin{center}
{\epsfxsize7.5cm\epsfysize4cm\epsfbox{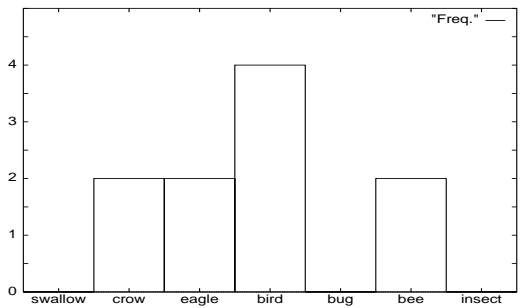}}
\end{center}
\caption{Frequency data for the subject slot for verb `fly.'}
\label{fig:csfreq}
\end{figure}

Table~\ref{tab:csdata2} shows some example cast slot data for the arg1
slot for the verb `fly.' By counting occurrences of each noun at the
slot, we can obtain frequency data shown in Figure~\ref{fig:csfreq}.

\begin{table}[htb]
\caption{Example case slot data.}
\label{tab:csdata2}
\begin{center}
\begin{tabular}{|lcc|} \hline
Verb & Slot name & Slot value \\ \hline
fly & arg1 & bee \\
fly & arg1 & bird \\
fly & arg1 & bird \\
fly & arg1 & crow \\
fly & arg1 & bird \\
fly & arg1 & eagle \\
fly & arg1 & bee \\
fly & arg1 & eagle \\
fly & arg1 & bird \\
fly & arg1 & crow \\ \hline
\end{tabular}
\end{center}
\end{table}

The problem of learning `case slot patterns' for a slot for a verb can
be viewed as the problem of estimating the underlying conditional
probability distribution which gives rise to the corresponding case
slot data.  The conditional distribution is defined as
\begin{equation}\label{eq:word-csmodel}
P(n|v,r),
\end{equation}
where random variable $n$ represents a value in the set of nouns ${\cal
  N}=\{n_1,n_2,\cdots,n_N\}$, random variable $v$ a value in the set
of verbs ${\cal V} = \{v_1,v_2,\cdots,v_V\}$, and random variable $r$
a value in the set of slot names ${\cal R}=\{r_1,r_2,$ $\cdots,r_R\}$. 
Since random variables take on words as their values, this type of
probability distribution is often referred to as a `word-based model.' 
The degree of noun $n$'s being the value of slot $r$ for verb
$v$\footnote{Hereafter, I will sometimes
  use the same symbol to denote both a random variable and one of its
  values; it should be clear from the context, which it is denoting at
  any given time.} is represented by a conditional probability.

Another way of learning case slot patterns for a slot for a verb is
to calculate the `association ratio' measure, as proposed in
\cite{Church89a,Church89b,Church91}. The association ratio is defined
as \begin{equation}\label{eq:assocratio} S(n|v,r) = \log
\frac{P(n|v,r)}{P(n)}, \end{equation} where $n$ assumes a value from
the set of nouns, $v$ from the set of verbs and $r$ from the set of
slot names. The degree of noun $n$ being the value of
slot $r$ for verb $v$ is represented as the ratio between a
conditional probability and a marginal probability.

The two measures in fact represent two different aspects of case slot
patterns. The former indicates the {\em relative frequency} of a
noun's being the slot value, while the latter indicates the {\em
  strength of associativeness} between a noun and the verb with
respect to the slot.  The advantage of the latter may be that it takes
into account of the influence of the marginal probability $P(n)$ on
the conditional probability $P(n|v,r)$. The advantage of the former
may be its ease of use in disambiguation as a likelihood value.

Both the use of the conditional probability and that of the
association ratio may suffer from the `data sparseness problem,' i.e.,
the number of parameters in the conditional distribution defined in
(\ref{eq:word-csmodel}) is very large, and accurately estimating them
is difficult with the amount of data typically available.

\begin{figure}[htb]
\begin{center}
{\epsfxsize7.5cm\epsfysize4cm\epsfbox{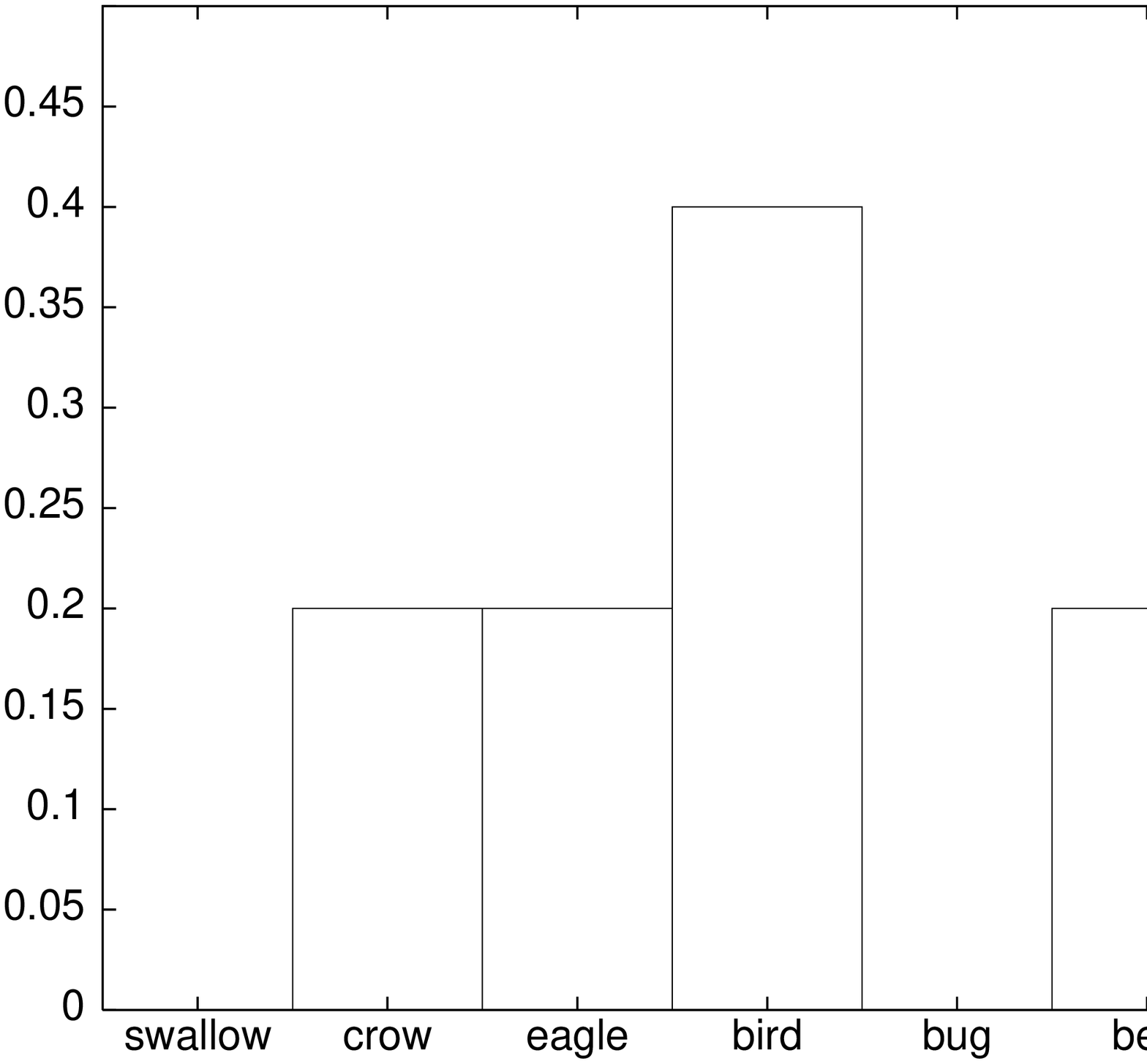}}
\end{center}
\caption{Word-based distribution estimated using MLE.}
\label{fig:csmodel}
\end{figure}

When we employ Maximum Likelihood Estimation (MLE) to estimate the
parameters, i.e., when we estimate the conditional probability
$P(n|v,r)$ as\footnote{Throughout this thesis, $\hat{\theta}$ denotes
an estimator (or an estimate) of $\theta$.}
\[
\hat{P}(n|v,r) = \frac{f(n|v,r)}{f(v,r)},
\] where $f(n|v,r)$ stands for the frequency of noun $n$ being the
value of slot $r$ for verb $v$, $f(v,r)$ the total frequency of $r$
for $v$ (Figure~\ref{fig:csmodel} shows the results for the data in
Figure~\ref{fig:csfreq}), we may obtain quite poor results. Most of the
probabilities might be estimated as 0, for example, just because a
possible value of the slot in question happens not to appear.

To overcome this problem, we can smooth the probabilities by resorting
to statistical techniques \cite{Jelinek80,Katz87,Gale90,Ristad95}. We
can, for example, employ an extended version of the Laplace's Law of
Succession (cf., \cite{Jeffreys61,Krichevskii81}) to estimate
$P(n|v,r)$ as \[ \hat{P}(n|v,r) = \frac{f(n|v,r)+0.5}{f(v,r)+0.5\cdot
N} \] where $N$ denotes the size of the set of nouns.\footnote{This
smoothing method can be justified from the viewpoint of Bayesian
Estimation. The estimate is in fact the Bayesian estimate with
Jeffrey's Prior being the prior probability.}

The results may still not be satisfactory, however. One possible way
to cope better with the data sparseness problem is to exploit {\em
  additional} knowledge or data rather than make use of only related
case slot data. Two such approaches have been proposed previously: one
is called the `similarity-based approach,' the other the `class-based
approach.'

\subsection{Similarity-based approach}

\namecite{Grishman94} propose to estimate conditional probabilities by
using other conditional probabilities under contexts of similar words,
where the similar words themselves are collected on the basis of
corpus data. Their method estimates the conditional probability
$P(n|v,r)$ as \[ \hat{P}(n|v,r) = \sum_{v'} \lambda(v,v') \cdot
\hat{P}(n|v',r), \] where $v'$ represents a verb similar to verb $v$,
and $\lambda(v,v')$ the similarity between $v$ and $v'$. That is, it
smoothes a conditional probability by taking the weighted average of
other conditional probabilities under contexts of similar words using
similarities as the weights. Note that the equation \[ \sum_{v'}
\lambda(v,v') = 1 \] must hold. The advantage of this approach is that
it relies only on corpus data. (Cf.,
\cite{Dagan92,Dagan94,Dagan97}.)

\subsection{Class-based approach}

A number of researchers have proposed to employ `class-based models,'
which use classes of words rather than individual words.

An example of a class-based approach is Resnik's method of
learning case slot patterns by calculating the
`selectional association' measure \cite{Resnik93a,Resnik93b}. The
selectional association is defined as:
\begin{equation}\label{eq:defsa}
  A(n|v,r) = \max_{C \ni n} \left(P(C|v,r) \cdot \log
  \frac{P(C|v,r)}{P(C)}\right), \end{equation} where $n$ represents a
  value in the set of nouns, $v$ a value in the set of verbs and $r$ a
  value in the set of slot names, and $C$ denotes a class of nouns
  present in a given thesaurus. (See also \cite{Framis94,Ribas95}.)
  This measure, however, is based on heuristics, and thus can be
  difficult to justify theoretically.

Other class-based methods for case slot generalization are also
proposed \cite{Almuallim94,Tanaka94,Tanaka96,Utsuro97,Miyata97}.

\section{Word Clustering}

Automatically clustering words or constructing a thesaurus can also be
considered to be a class-based approach, and it helps cope with the
data sparseness problem not only in case frame pattern acquisition but
also in other natural language learning tasks.

\begin{figure}[htb]
\begin{center}
\epsfxsize5cm\epsfysize5cm\epsfbox{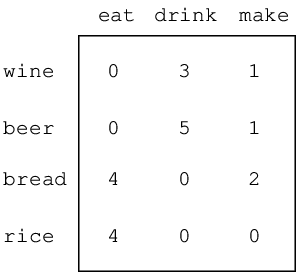}
\end{center}
\caption{Example co-occurrence data.}
\label{fig:co-occurrence}
\end{figure}

If we focus our attention on one case slot, we can obtain
`co-occurrence data' for verbs and nouns with respect to that
slot. Figure~\ref{fig:co-occurrence}, for example, shows such data, in
this case, counts of co-occurrences of verbs and their arg2 slot
values (direct objects). We can classify words by using such
co-occurrence data on the assumption that semantically similar words
have similar co-occurrence patterns.

A number of methods have been proposed for clustering words on the
basis of co-occurrence data. \namecite{Brown92}, for example, propose
a method of clustering words on the basis of MLE in the context of
n-gram estimation. They first define an n-gram class model as \[
P(w_n|w_1^{n-1}) = P(w_n|C_n) \cdot P(C_n|C_1^{n-1}), \] where $C$
represents a word class. They then view the clustering problem as that
of partitioning the vocabulary (a set of words) into a designated
number of word classes whose resulting 2-gram class model has the
maximum likelihood value with respect to a given word sequence (i.e.,
co-occurrence data). Brown et al have also devised an efficient
algorithm for performing this task, which turns out to iteratively
merge the word class pair having the least reduction in empirical
mutual information until the number of classes created equals the
designated number. The disadvantage of this method is that one has to
designate in advance the number of classes to be created, with no
guarantee at all that this number will be optimal.

\namecite{Pereira93} propose a method of clustering words based on
co-occurrence data over two sets of words. Without loss of generality,
suppose that the two sets are a set of nouns ${\cal N}$ and a set of
verbs ${\cal V}$, and that a sample of co-occurrence data is given as
$(n_i,v_i), n_i \in {\cal N}, v_i \in {\cal V}, i=1,\cdots,s$. They
define \[ P(n,v) = \sum_{C} P(C) \cdot P(n|C) \cdot P(v|C) \] as a
model which can give rise to the co-occurrence data, where $C$
represents a class of nouns. They then view the problem of clustering
nouns as that of estimating such a model. The classes obtained in this
way, which they call `soft clustering,' have the following properties:
(1) a noun can belong to several different classes, and (2) each class
is characterized by a membership distribution. They devised an
efficient clustering algorithm based on `deterministic annealing
technique.'\footnote{Deterministic annealing is a computation
  technique for finding the global optimum (minimum) value of a cost
function \cite{Rose90,Ueda98}. The basic idea is to conduct
minimization by using a number of `free energy'
  functions parameterized by `temperatures' for which free energy
  functions with high temperatures {\em loosely} approximate a target
  function, while free energy functions with low temperatures {\em
precisely} approximate it. A deterministic-annealing-based
  algorithm manages to find the global minimum value of the target
  function by continuously finding the minimum values of the free
  energy functions while incrementally decreasing the temperatures. 
  (Note that deterministic annealing is different from the classical
  `simulated annealing' technique \cite{Kirkpatrick83}.) In Pereira et
  al's case, deterministic annealing is used to find the minimum of
  average distortion. They have proved that, in their problem setting,
  minimizing average distortion is equivalent to maximizing likelihood
  with respect to the given data (i.e., MLE).} Conducting
soft clustering makes it possible to cope with structural and word
sense ambiguity at the same time, but it also requires more training
data and makes the learning process more computationally demanding.

\namecite{Tokunaga95} point out that, for disambiguation purposes, it
is necessary to construct one thesaurus for each case slot on the
basis of co-occurrence data concerning to that slot. Their
experimental results indicate that, for disambiguation, the use of
thesauruses constructed from data specific to the target slot is
preferable to the use of thesauruses constructed from data
non-specific to the slot.

Other methods for automatic word clustering
have also been proposed
\cite{Hindle90,Pereira92,McKeown93,Grefenstette94,Stolcke94,Abe95,McMahon96,Ushioda96,Hogenhout97}.

\section{Case Dependency Learning}

There has been no method proposed to date, however, that learns
dependencies between case slots. In past research, methods of
resolving ambiguities have been based, for example, on the assumption
that case slots are mutually independent
\cite{Hindle91,Sekine92,Resnik93a,Grishman94,Alshawi94}, or at most
two case slots are dependent \cite{Brill94,Ratnaparkhi94,Collins95}.

\section{Structural Disambiguation}

\subsection{The lexical approach}

There have been many probabilistic methods proposed in the literature
to address the structural disambiguation problem. Some methods tackle
the basic problem of resolving ambiguities in quadruples $(v, n_1, p,
n_2)$ (e.g., (eat, ice-cream, with, spoon)) by mainly using lexical
knowledge. Such methods can be classified into the following three
types: the double approach, the triple approach, and the quadruple
approach.

The first two approaches employ what I call a `generation model' and
the third approach employs what I call a `decision model' (cf.,
Chapter 3).

\subsubsection*{The double approach}
This approach takes doubles of the form $(v,p)$ and $(n_1,p)$, like
those in Table~\ref{tab:doubledata}, as training data to acquire lexical
knowledge and judges the attachment sites of $(p,n_2)$ in quadruples
based on the acquired knowledge.

\begin{table}[htb]
\caption{Example input data as doubles.}
\label{tab:doubledata}
\begin{center}
\begin{tabular}{|l|} \hline
eat in \\
eat with \\
ice-cream with \\
candy with \\ \hline
\end{tabular}
\end{center}
\end{table}

\namecite{Hindle91} propose the use of the so-called `lexical
association' measure calculated based on such doubles: \[ P(p|v), \]
where random variable $v$ represents a verb (in general a head), and
random variable $p$ a slot (preposition).  They further propose
viewing the disambiguation problem as that of hypothesis testing. More
specifically, they calculate the `t-score,' which is a statistic on
the difference between the two estimated probabilities $\hat{P}(p|v)$
and $\hat{P}(p|n_1)$: \[
 t = \frac{\hat{P}(p|v) -
    \hat{P}(p|n_1)}{\sqrt{\frac{\hat{\sigma}^2_v}{N_v}
      + \frac{\hat{\sigma}^2_{n_1}}{N_{n_1}}}},
  \] where $\hat{\sigma}_{v}$ and $\hat{\sigma}_{n_1}$ denote the
  standard deviations of $\hat{P}(p|v)$ and $\hat{P}(p|n_1)$,
  respectively, and $N_v$ and $N_{n_1}$ denote the data sizes used to
  estimate these probabilities. If, for example, $t>1.28$, then
  $(p,n_2)$ is attached to $v$, $t<-1.28$, $(p,n_2)$ is attached to
  $n_1$, and otherwise no decision is made. (See also
  \cite{Hindle93}.)

\subsubsection*{The triple approach}
This approach takes triples $(v,p,n_2)$ and $(n_1,p,n_2)$, i.e., case
slot data, like those in Table~\ref{tab:tripledata}, as training data for
acquiring lexical knowledge, and performs pp-attachment disambiguation
on quadruples.

\begin{table}[htb]
\caption{Example input data as triples.}
\label{tab:tripledata}
\begin{center}
\begin{tabular}{|l|} \hline
eat in park \\
eat with spoon \\
ice-cream with chocolate \\
eat with chopstick  \\
candy with chocolate \\ \hline
\end{tabular}
\end{center}
\end{table}

For example, \namecite{Resnik93a} proposes the use of the selectional
association measure (described in Section 2) calculated on the basis
of such triples. The basic idea of his method is to compare
$A(n_2|v,p)$ and $A(n_2|n_1,p)$ defined in (\ref{eq:defsa}), and make
a disambiguation decision.

\namecite{Sekine92} propose the use of joint probabilities
$P(v,p,n_2)$ and $P(n_1,p,n_2)$ in pp-attachment disambiguation. They
devised a heuristic method for estimating the probabilities.  (See also
\cite{Alshawi94}.)

\subsubsection*{The quadruple approach}
This approach receives quadruples $(v, n_1, p, n_2)$, as well as
labels that indicate which way the pp-attachment goes, such as those
in Table~\ref{tab:quadrupledata}; and it learns disambiguation rules.

\begin{table}[htb]
\caption{Example input data as quadruples and labels.}
\label{tab:quadrupledata}
\begin{center}
\begin{tabular}{|lc|} \hline
eat ice-cream in park & attv \\
eat ice-cream with spoon & attv \\
eat candy with chocolate & attn \\ \hline
\end{tabular}
\end{center}
\end{table}

It in fact employs the conditional
probability distribution -- a `decision model'
\begin{equation}\label{eq:defquad}
P(a|v,n_1,p,n_2),
\end{equation}
where random variable $a$ takes on attv and attn as its values, and
random variables $(v,n_1,p,n_2)$ take on quadruples as their values.
Since the number of parameters in the distribution is very large,
accurate estimation of the distribution would be impossible.

In order to address this problem, \namecite{Collins95} devised a
back-off method. It first calculates the conditional probability
$P(a|v,n_1,p,n_2)$ by using the relative frequency \[
\frac{f(a,v,n_1,p,n_2)}{f(v,n_1,p,n_2)}, \] if the denominator is
larger than 0; otherwise it successively uses lower order frequencies
to heuristically calculate the probability.

\namecite{Ratnaparkhi94} propose to learn the conditional probability
distribution (\ref{eq:defquad}) with Maximum Entropy Estimation. They
adopt the Maximum Entropy Principle (MEP) as the learning strategy,
which advocates selecting the model having the maximum entropy from
among the class of models that satisfies certain constraints (see
Section~\ref{sec:mdl-estimation} for a discussion on the relation
between MDL and MEP). The fact that a model must be one such that the
expected value of a feature with respect to it equals that with
respect to the empirical distribution is usually used as a constraint. 
Ratnaparkhi et al's method defines, for example, a feature as follows
\[ f_i = \left\{ \begin{array}{ll} 1 & \mbox{$(p,n_2)$ is attached to
$n_1$ in (-, ice-cream, with, chocolate)} \\ 0 & \mbox{otherwise}. \\
\end{array} \right. \] It then incrementally selects features, and
efficiently estimates the conditional distribution by using the
Maximum Entropy Estimation technique (see
\cite{Jaynes78,Darroch72,Berger96}).

Another method of the quadruple approach is to employ
`transformation-based error-driven learning' \cite{Brill95}, as
proposed in \cite{Brill94}. This method learns and uses IF-THEN type
rules, where the IF parts represent conditions like ($p$ is `with')
and ($v$ is `see'), and the THEN parts represent transformations from
(attach to $v$) to (attach to $n_1$), and vice-versa.  The first rule
is always a default decision, and all the other rules indicate
transformations (changes of attachment sites) subject to various IF
conditions.

\subsection{The combined approach}

Although the use of lexical knowledge can effectively resolve
ambiguities, it still has limitation. It is preferable, therefore, to
utilize other kind of knowledge in disambiguation, especially when a
decision cannot be made solely on the basis of lexical knowledge.

The following two facts suggest that syntactic knowledge should also
be used for the purposes of disambiguation. First, interpretations are
obtained through syntactic parsing. Second, psycholinguistists observe
that there are certain syntactic principles in human's language
interpretation. For example, in English a phrase on the right tends to
be attached to the nearest phrase on the left, - referred to as the
`right association principle' \cite{Kimball73}. (See also
\cite{Ford82,Frazier79,Hobbs90,Whittemore90}).

We are thus led to the problem of how to define a probability model
which combines the use of both lexical semantic knowledge and
syntactic knowledge. One approach is to introduce probability models
on the basis of syntactic parsing. Another approach is to introduce
probability models on the basis of psycholinguistic principles
\cite{Li96a}.

Many methods belonging to the former approach have been proposed. A
classical method is to employ the PCFG (Probabilistic Context Free
Grammar) model \cite{Fujisaki89,Jelinek90,Lari90}, in
which a CFG rule having the form of \[ A \rightarrow B_1 \cdots B_m \]
is associated with a conditional probability
\begin{equation}\label{eq:cfg} P(B_1,\cdots,B_m |A).  \end{equation}
  In disambiguation the likelihood of an interpretation is defined as
  the product of the conditional probabilities of the rules which are
  applied in the derivation of the interpretation.

The use of PCFG, in fact, resorts more to syntactic knowledge rather
than to lexical knowledge, and its performance seems to be only
moderately good \cite{Chitrao90}. There are also many methods
proposed which more effectively make use of lexical knowledge.

\namecite{Collins97} proposes disambiguation through use of a
generative probability model based on a lexicalized CFG (in fact, a
restricted form of HPSG \cite{Pollard87}). (See also
\cite{Collins96,Schabes92,Hogenhout96,Den96,Charniak97}.) In Collins'
model, each lexicalized CFG rule is defined in the form of \[ A
\rightarrow L_n \cdots L_1 H R_1 \cdots R_m, \] where a capitalized
symbol denotes a category, with $H$ being the head category on the
right hand site. A category is defined in the form of $C(w,t)$,
where $C$ denotes the name of the category, $w$ the head word
associated with the category, and $t$ the part-of-speech tag assigned
to the head word.  Furthermore, each rule is assigned a conditional
probability $P(L_n, \cdots, L_1, H, R_1, \cdots, R_m|A)$
(cf., (\ref{eq:cfg})) that is assumed to satisfy \[ \begin{array}{l}
\\
 P(L_n, \cdots, L_1, H, R_1, \cdots, R_m|A) = P(H|A) \cdot P(L_1, \cdots, L_n
|A,H) \cdot P(R_1, \cdots, R_m |A,H).  \end{array} \] 
In disambiguation,
the likelihood of an interpretation is defined as the product of the
conditional probabilities of the rules which are applied in the
derivation of the interpretation. While Collins has devised several
heuristic methods for estimating the probability model, further
investigation into learning methods for this model still appears
necessary.

\namecite{Magerman95} proposes a new parsing approach based on
probabilistic decision tree models \cite{Quinlan89,Yamanishi92a} to
replace conventional context free parsing. His method uses decision
tree models to construct parse trees in a bottom-up and left-to-right
fashion. A decision might be made, for example, to create a new
parse-tree-node, and conditions for making that decision might be, for
example, the appearances of certain words and certain tags in a node
currently being focussed upon and in its neighbor nodes. Magerman has
also devised an efficient algorithm for finding the parse tree
(interpretation) with the highest likelihood value.  The advantages of
this method are its effective use of contextual information and its
non-use of a hand-made grammar. (See also
\cite{Magerman91,Magerman94,Black93,Ratnaparkhi97,Haruno98})

\namecite{Su88} propose the use of a probabilistic score function for
disambiguation in generalized LR parsing (see also
\cite{Su89,Chang92,Chiang95,Wright90,Kita92,Briscoe93,Inui98}). They
first introduce a conditional probability of a category obtained after
a reduction operation and in the context of the reduced categories and
of the categories immediately left and right of those reduced
categories. The score function, then, is defined as the product of the
conditional probabilities appearing in the derivation of the
interpretation. The advantage of the use of this score function is its
context-sensitivity, which can yield more accurate results in
disambiguation.

\namecite{Alshawi94} propose for disambiguation purposes the use of a
linear combination of various preference functions based on lexical
and syntactic knowledge. They have devised a method for training the
weights of a linear combination. Specifically, they employ the
minimization of a squared-error cost function as a learning strategy
and employ a `hill-climbing' algorithm to iteratively adjust weights
on the basis of training data.

Additionally, some non-probabilistic approaches to structural
disambiguation have also been proposed (e.g.,
\cite{Wilks75,Wermter89,Nagao90,Kurohashi94}).

\section{Word Sense Disambiguation}

Word sense disambiguation is an issue closely related to the
structural disambiguation problem. For example, when analyzing the
sentence ``Time flies like an arrow," we obtain a number of ambiguous
interpretations. Resolving the sense ambiguity of the word `fly'
(i.e., determining whether the word indicates `an insect' or `the
action of moving through the air'), for example, helps resolve the
structural ambiguity, and the converse is true as well.

There have been many methods proposed to address the word sense
disambiguation problem. (A number of tasks in natural language
processing, in fact, fall into the category of word sense
disambiguation \cite{Yarowsky93}. These include homograph
disambiguation in speech synthesis, word selection in machine
translation, and spelling correction in document processing.)

A simple approach to word sense disambiguation is to employ the
conditional distribution -- a `decision model'
\[ P(D|E_1,\cdots,E_n), \]
where random variable $D$ assumes word senses as its values, and
random variables $E_i (i=1,\cdots,n)$ represent pieces of evidence
for disambiguation.  For example, $D$ can be the insect sense or the
action sense of the word `fly,' $E_i$ can be the presence or absence
of the word `time' in the context. Word sense disambiguation, then,
can be realized as the process of finding the sense $d$ whose
conditional probability $P(d|e_1,\cdots,e_n)$ is the largest, where
$e_1,\cdots,e_n$ are the values of the random variables
$E_1,\cdots,E_n$ in the current context.

Since the conditional distribution has a large number of parameters,
however, it is difficult to estimate them. One solution to this
difficulty is to estimate the conditional probabilities by using
Bayes' rule and by assuming that the pieces of evidence for
disambiguation are mutually independent \cite{Yarowsky92}.
Specifically, we select a sense $d$ satisfying: \[ \begin{array}{ll}
 \arg\max_{d \in D} P(d|e_1,\cdots,e_n) & = \arg\max_{d \in D} \{ \frac{P(d)\cdot P(e_1,\cdots,e_n|d)}{P(e_1,\cdots,e_n)} \}, \\
 & = \arg\max_{d \in D} \{ P(d)\cdot P(e_1,\cdots,e_n|d) \}, \\
 & = \arg\max_{d \in D} \{ P(d)\cdot \prod_{i=1}^{n}P(e_i|d) \}, \\
\end{array}
\] 

Another way of estimating the conditional probability distribution is
to represent it in the form of a {\em probabilistic} decision
list\footnote{A probabilistic decision list \cite{Yamanishi92a} is a
kind of conditional distribution and different from a {\em
deterministic} decision list \cite{Rivest87}, which is a kind of
Boolean function.}, as is proposed in \cite{Yarowsky94}. Since a
decision list is a sequence of IF-THEN type rules, the use of it in
disambiguation turns out to utilize only the strongest pieces of
evidence. Yarowsky has also devised a heuristic method for efficient
learning of a probabilistic decision list. The merits of this method
are ease of implementation, efficiency in processing, and clarity.

Another approach to word sense disambiguation is the use of weighted
majority learning \cite{Littlestone88,Littlestone94}. Suppose, for the
sake of simplicity, that the disambiguation decision is binary, i.e.,
it can be represented as 1 or 0.  We can first define a linear
threshold function: \[ \sum_{i=1}^{n} w_i \cdot x_i \] where feature
$x_i (i=1,\cdots,n)$ takes on $1$ and $0$ as its values, representing
the presence and absence of a piece of evidence, respectively, and
$w_i (i=1,\cdots,n)$ denotes a non-negative real-valued weight. In
disambiguation, if the function exceeds a predetermined threshold
$\theta$, we choose $1$, otherwise $0$. We can further employ a
learning algorithm called `winnow' that updates the weights in an
on-line (or incremental) fashion.\footnote{Winnow is similar to the
well-known classical `perceptron' algorithm, but the former uses a
multiplicative weight update scheme while the latter uses an
additive weight update scheme. \namecite{Littlestone88} has shown
that winnow performs much better than perceptron when many attributes
are irrelevant.} This algorithm has the advantage of being able to
handle a large set of features, and at the same time not ordinarily be
affected by features that are irrelevant to the disambiguation
decision. (See \cite{Golding96a}.)

For word sense disambiguation methods, see also
\cite{Black88,Brown91,Guthrie91,Gale92,McRoy92,Leacock93,Yarowsky93,Bruce94,Niwa94,Voorhees95,Yarowsky95,Golding96b,Ng96,Fujii96,Schutze97,Schutze98}.

\section{Introduction to MDL}

The Minimum Description Length principle is a strategy (criterion) for
data compression and statistical estimation, proposed by Rissanen
(1978; 1983; 1984; 1986; 1989; 1996; 1997). Related strategies were
also proposed and studied independently in
\cite{Solomonoff64,Wallace68,Schwarz78}. A number of important
properties of MDL have been demonstrated by \namecite{Barron91} and
\namecite{Yamanishi92a}.

MDL states that, for both data compression and statistical estimation,
the best probability model {\em with respect to given data} is that
which requires the shortest code length in bits for encoding the model
itself and the data observed through it.\footnote{In this thesis, I
  describe MDL as a criterion for both data compression and
  statistical estimation. Strictly speaking, however, it is only referred to as
  the `MDL principle' when used as a criterion for statistical
estimation.}

In this section, we will consider the basic concept of MDL and, in
particular how to calculate description length. Interested readers
are referred to \cite{Quinlan89,Yamanishi92a,Yamanishi92b,Han94} for
an introduction to MDL.

\subsection{Basics of Information Theory}

\subsubsection*{IID process}
Suppose that a data sequence (or a sequence of symbols)
\[
x^{n} = x_1 x_2 \cdots x_n
\] 
is {\em independently} generated according to a discrete probability
distribution
\begin{equation}\label{eq:distribution}
P(X),
\end{equation}
where random variable (information source) $X$ takes on values
from a set of symbols: 
\[
  \{1,2,\cdots,s\}.  \] Such a data generation process is generally
referred to as `i.i.d' (independently and identically distributed).

In order to transmit or compress the data sequence, we need to define
a code for encoding the information source $X$, i.e., to assign to
each value of $X$ a codeword, namely a bit string. In order for the
decoder to be able to decode a codeword as soon as it comes to the end
of that codeword, the code must be one in which no codeword is a
prefix of any other codeword. Such a code is called a `prefix code.'
\begin{theorem}\label{th:kraft} The sufficient and necessary
condition for a code to be a prefix code is as follows, \[
  \sum_{i=1}^{s} 2^{-l(i)} \le 1,
  \]
where $l(i)$
denotes the code length of the codeword assigned to symbol $i$.  \end{theorem}
This is known as Kraft's inequality. 

We define the expected (average) code length of a code for encoding
the information source $X$ as \[ L(X) = \sum_{i=1}^{s} P(i) \cdot
l(i).  \] Moreover, we define the entropy of (the distribution of) $X$
as\footnote{Throughout this thesis, `$\log$' denotes logarithm to the
base 2.}  \[
  H(X) = - \sum_{i=1}^{s} P(i) \cdot \log P(i).
\]
\begin{theorem}\label{th:shannon1} The expected
code length of a prefix code for encoding the information source $X$
is greater than or equal to the entropy of $X$, namely \[ L(X) \ge
H(X).  \] \end{theorem}

We can define a prefix code in which symbol $i$ is assigned a
codeword with code length
\[
l(i) = - \log P(i) \ \ \ \ (i = 1,\cdots,s),
\]
according to Theorem~\ref{th:kraft}, since
\[
\sum_{i=1}^s 2^{\log P(i)} = 1.
\] 
Such a code is on average the most efficient prefix code, according to
Theorem~\ref{th:shannon1}. Hereafter, we refer to this type of code as a `
non-redundant code.' (In real communication, a code length must be a
truncated integer: $\lceil -\log P(i) \rceil$,\footnote{$\lceil x
  \rceil$ denotes the least integer not less than $x$.} but we use
here $-\log P(i)$ for ease of mathematical manipulation.  This is not
harmful and on average the error due to it is negligible.)  When the
distribution $P(X)$ is a uniform distribution, i.e.,
\[
P(i) = \frac{1}{s} \ \ \ \ (i = 1,\cdots,s),
\]
the code length for encoding each symbol $i$ turns out to be
\[
l(i) = -\log P(i) = -\log \frac{1}{s} = \log s \ \ \ \ (i = 1,\cdots,s).
\]

\subsubsection*{General case}
We next consider a more general case. We assume that the data sequence
\[
x^n = x_1x_2\cdots x_n
\]
is generated according to a probability distribution $P(X^n)$ where
random variable $X_i (i=1,\cdots,n)$ takes on values from
$\{1,2,\cdots,s\}$.  The data generation process needs neither be
i.i.d. nor even stationary (for the definition of a stationary
process, see, for example, \cite{Cover91}).  Again, our goal is to
transmit or compress the data sequence.

We define the expected code length for encoding a sequence of $n$
symbols as \[ L(X^n) = \sum_{x^n} P(x^n) \cdot l(x^n), \] where
$P(x^n)$ represents the probability of observing the data sequence
$x^n$ and $l(x^n)$ the code length for encoding $x^n$. We further
define the entropy of (the distribution of) $X^n$ as \[ H(X^n) = -
\sum_{x^n} P(x^n) \log P(x^n). \]
We have the following theorem, widely known as Shannon's first theorem
(cf., \cite{Cover91}).  \begin{theorem}\label{th:shannon2} The
expected code length of a prefix code for encoding a sequence of $n$
symbols $X^n$ is greater than or equal to the entropy of $X^n$, namely
\[ L(X^n) \ge H(X^n). \]
\end{theorem}

As in the i.i.d. case, we can define a non-redundant code in which the
code length for encoding the data sequence $x^n$ is
\begin{equation}\label{eq:datalen} l(x^n) = - \log
  P(x^n). \end{equation} The expected code length of the code for
  encoding a sequence of $n$ symbols then becomes
\begin{equation}\label{eq:avgdatalen} L(X^n) = H(X^n). \end{equation}

Here we assume that we know in advance the distribution $P(X)$ (in
general $P(X^n)$). In practice, however, we usually do not know what
kind of distribution it is. We have to estimate it by using the same
data sequence $x^n$ and transmit first the estimated model and then
the data sequence, which leads us to the notion of two-stage coding.

\subsection{Two-stage code and MDL}

In two-stage coding, we first introduce a class of models which
includes all of the possible models which can give rise to the
data. We then choose a prefix code and encode each model in the class.
The decoder is informed in advance as to which class has been
introduced and which code has been chosen, and thus no matter which
model is transmitted, the decoder will be able to identify it. We next
calculate the total code length for encoding each model and the data
through the model, and select the model with the shortest total code
length. In actual transmission, we transmit first the selected model
and then the data through the model. The decoder then can restore the
data perfectly.

\subsubsection*{Model class} We first introduce a class of models, of
which each consists of a discrete model (an expression) and a
parameter vector (a number of parameters). When a discrete model is
specified, the number of parameters is also determined.

For example, the tree cut models within a thesaurus tree, to be
defined in Chapter 4, form a model class. A discrete model in this
case corresponds to a cut in the thesaurus tree. The number of free
parameters equals the number of nodes in the cut minus one.

The class of `linear regression models' is also an example model
class. A discrete model is \[ a_0 + a_1 \cdot x_1 + \cdots + a_k \cdot x_k +
\epsilon, \] where $x_i (i=1,\cdots,k)$ denotes a
random variable, $a_i (i=0,1,\cdots,k)$ a parameter, and $\epsilon$ a
random variable based on the standard normal distribution
$N(0,1)$. The number of parameters in this model equals $(k+1)$.

A class of models can be denoted as
\[
{\cal M} = \{P_{\theta}(X): \theta \in \Theta(m), m \in M \},
\] where $m$ stands for a discrete model, $M$ a set of
discrete models, $\theta$ a parameter vector, and $\Theta(m)$ a
parameter space associated with $m$.

Usually we assume that the model class we introduced contains the
`true' model which has given rise to the data, but it does not matter
if it does not. In such case, the best model selected from the class
can be considered an approximation of the true model. The model class
we introduce reflects our prior knowledge on the problem.

\subsubsection*{Total description length} We next consider how to
calculate total description length.

Total description length equals the sum total of the code length for
encoding a discrete model (model description length $l(m)$), the code
length for encoding parameters given the discrete model (parameter
description length $l(\theta|m)$), and the code length for encoding
the data given the discrete model and the parameters (data description
length $l(x^n|m,\theta)$). Note that we also sometimes refer to the
model description length as $l(m)+l(\theta|m)$.

Our goal is to find the minimum description length of the data (in
number of bits) with respect to the model class, namely, \[ L_{min}
(x^n: {\cal M}) = \min_{m \in M} \min_{\theta \in \Theta(m)} \left(
\begin{array}{l}l(m) + l(\theta|m) + l(x^n|m,\theta)\end{array}
\right).  \]

\subsubsection*{Model description length} Let us first consider how to
calculate model description length $l(m)$. The choice of a code for
encoding discrete models is subjective; it depends on our prior
knowledge on the model class.

If the set of discrete models $M$ is finite and the probability
distribution over it is a uniform distribution, i.e., \[ P(m) =
\frac{1}{|M|}, m \in M, \] then we need \[ l(m) = \log |M| \] to
encode each discrete model $m$ using a non-redundant code.

If $M$ is a countable set, i.e., each of its members can be assigned a
positive integer, then the `Elias code,' which is usually used for
encoding integers, can be employed \cite{Rissanen89}. Letting $i$ be
the integer assigned to a discrete model $m$, we need \[ l(m) = \log c
+ \log i + \log \log i + \cdots \] to encode $m$. Here the sum
includes all the positive iterates and $c$ denotes a constant of about
2.865.

\subsubsection*{Parameter description length and data description length}
When a discrete model $m$ is fixed, a parameter space will be uniquely
determined. The model class turns out to be
\[
{\cal M}_m = \{P_{\theta}(X): \theta \in \Theta \},
\] where $\theta$ denotes a parameter vector, and $\Theta$ the
parameter space. Suppose that the dimension of the parameter space is
$k$, then $\theta$ is a vector with $k$ real-valued components:
\[
\theta = (\theta_1,\cdots,\theta_k)^T,
\]
where $X^T$ denotes a transpose of $X$.

We next consider a way of calculating the sum of the parameter
description length and the data description length through its
minimization: \[ \min_{\theta \in \Theta}
(l(\theta|m)+l(x^n|m,\theta)).  \]

Since the parameter space $\Theta$ is usually a subspace of the
$k$-dimensional Euclidean space and has an infinite number of points
(parameter vectors), straightforwardly encoding each point in the
space takes the code length to infinity, and thus is
intractable. (Recall the fact that before transmitting an element in a
set, we need encode each element in the set.) One possible way to deal
with this difficulty is to discretize the parameter space; the process
can be defined as a mapping from the parameter space to a discretized
space, depending on the data size $n$: \[ \Delta_n : \Theta
\rightarrow \Theta_n.  \] A discretized parameter space consists of a
finite number of elements (cells). We can designate one point in each
cell as its representative and use only the representatives for
encoding parameters. The minimization then turns out to be \[
\min_{\bar{\theta} \in \Theta_n} (l(\bar{\theta}|m) +
l(x^n|m,\bar{\theta})),
  \] where $l(\bar{\theta}|m)$ denotes the code length for encoding
  a representative $\bar{\theta}$ and $l(x^n|m,\bar{\theta})$ denotes
  the code length for encoding the data $x^n$ through that representative.

  A simple way of conducting discretization is to define a cell as a micro
  $k$-dimensional rectangular solid having length $\delta_i$ on the
  axis of $\theta_i$. If the volume of
  the parameter space is $V$, then we have
  $V/(\delta_1\cdots\delta_k)$ number of cells. If the distribution
  over the cells is uniform, then we need
\[
l(\bar{\theta}|m) = \log \frac{V}{\delta_1\cdots\delta_k} 
\] to encode each representative $\bar{\theta}$ using a non-redundant code.

On the other hand, since the number of parameters is fixed and the
data is given, we can estimate the parameters by employing Maximum
Likelihood Estimation (MLE), obtaining \[ \hat{\theta} =
(\hat{\theta}_1,\cdots,\hat{\theta}_k)^T.  \] We may expect that the
representative of the cell into which the maximum likelihood estimate
falls is the nearest to the true parameter vector among all
representatives. And thus, instead of conducting minimization over all
representatives, we need only consider minimization with respect to
the representative of the cell which the maximum likelihood estimate
belongs to.  This representative is denoted here as
$\tilde{\theta}$. We approximate the difference between $\hat{\theta}$
and $\tilde{\theta}$ as
  \[ \tilde{\theta} - \hat{\theta} \approx \delta \ \ \ \ \delta =
  (\delta_1,\cdots,\delta_k)^T.
  \] 
Data description length using
$\tilde{\theta}$ then becomes \[
  l(x^n|m,\tilde{\theta}) = - \log P_{\tilde{\theta}}(x^n).
\]

Now, we need consider only
\[
\begin{array}{l}
  \min_{\delta} (l(x^n|m,\tilde{\theta})+l(\tilde{\theta}|m))
=
  \min_{\delta} 
\left( \log \frac{V}{\delta_1\cdots\delta_k} - \log
  P_{\tilde{\theta}}(x^n) \right). \\ \end{array} \] There
is a trade-off relationship between the first term and the second
term. If $\delta$ is large, then the first term will be small, while
on the other hand the second term will be large, and vice-versa. That
means that if we discretize the parameter space loosely, we will need
less code length for encoding the parameters, but more code length
for encoding the data. On the other hand, if we discretize the
parameter space precisely, then we need less code length for
encoding the data, but more code length for encoding the
parameters.

In this way of calculation (see Appendix~\ref{append:mdl} for a
derivation), we have
\begin{equation}\label{eq:mdl}
  l(\theta|m)+l(x^n|m,\theta) = -
  \log P_{\hat{\theta}}(x^n) + \frac{k}{2} \cdot \log n
  + O(1),
  \end{equation} 
  where $O(1)$ indicates $\lim_{n \rightarrow \infty} O(1) = c$, a
  constant. The first term corresponds to the data description length
  and has the same form as that in (\ref{eq:datalen}). The second term
  corresponds to the parameter description length. An intuitive
  explanation of it is that the standard deviation of the maximum
  likelihood estimator of one of the parameters is of order
  $O(\frac{1}{\sqrt{n}})$,\footnote{It is well known that under
    certain suitable conditions, when the data size increases, the distribution
    of the maximum likelihood estimator $\hat{\theta}$ will
    asymptotically become the normal distribution
    $N(\theta^{*},\frac{1}{n\cdot I})$ where $\theta^{*}$ denotes the
    true parameter vector, $I$ the Fisher information matrix, and $n$
    the data size \cite{Fisher56}.} and hence encoding the parameters
using more than $k \cdot (-\log \frac{1}{\sqrt{n}})$ $=\frac{k}{2}
  \cdot \log n$ bits would be wasteful for the given data size.

In this way, the sum of the two kinds of description length
$l(\theta|m)+l(x^n|m,\theta)$ is obtained for a fixed discrete model
$m$ (and a fixed dimension $k$). For a different $m$, the sum can also
be calculated.

\subsubsection*{Selecting a model with minimum total description length}
Finally, the minimum total description length becomes, for example,
\[
  L_{min} (x^n: {\cal M}) = \min_{m} \left(
- \log P_{\hat{\theta}}(x^n)
+ \frac{k}{2} \cdot \log n + \log |M|
\right).
\]
We select the model with the minimum total description length for
transmission (data compression).

\subsection{MDL as data compression criterion}
Rissanen has proved that MDL is an optimal criterion for data
compression.

\begin{theorem}{\rm \cite{Rissanen84}}\label{th:rissanen1}
  Under certain suitable conditions, the expected code length of the two-stage
  code described above (with code length (\ref{eq:mdl}) for encoding
the data sequence $x^n$) satisfies \[ L(X^n) = H(X^n) +
\frac{k}{2}\cdot \log n + O(1), \] where $H(X^n)$ denotes the entropy
of $X^n$. \end{theorem}
 This theorem indicates that
when we do not know the true distribution $P(X^n)$ in communication, we
have to waste on average about $\frac{k}{2}\log n$ bits of code
length (cf., (\ref{eq:avgdatalen})).

\begin{theorem}{\rm \cite{Rissanen84}}\label{th:rissanen2}
  Under certain suitable conditions, for {\em any prefix code}, for some
$\epsilon_n > 0$ such that
  $\lim_{n \rightarrow \infty}\epsilon_n = 0$ and $\Omega_n
  \subset \Theta$ such that the volume of it ${\rm
    vol}(\Omega_n)$ satisfies $\lim_{n \rightarrow \infty} {\rm
    vol}(\Omega_n) = 0$, for any model with parameter
  $\theta \in (\Theta - \Omega_n)$, the expected code
  length $L(X^n)$ is bounded from below by \[ L(X^n) \ge H(X^n) +
\left(\frac{k}{2}-\epsilon_n\right)\cdot \log n. \] \end{theorem} This
theorem indicates that in general, i.e., excluding some special cases,
we cannot make the average code length of a prefix code more efficient
than the quantity $H(X^n) + \frac{k}{2} \cdot \log n$.  The
introduction of $\Omega_n$ eliminates the case in which we happen to
select the true model and achieve on average a very short code length:
$H(X^n)$. Theorem~\ref{th:rissanen2} can be considered as an extension
of Shannon's Theorem (Theorem~\ref{th:shannon2}).

Theorems \ref{th:rissanen1} and \ref{th:rissanen2} suggest that using
the two-stage code above is nearly optimal in terms of {\em expected}
code length. We can, therefore, say that encoding a data sequence
$x^n$ in the way described in (\ref{eq:mdl}) is the most efficient
approach not only to encoding the data sequence, but also, on average,
to encoding a sequence of $n$ symbols.

\subsection{MDL as estimation criterion}\label{sec:mdl-estimation}
The MDL principle stipulates that selecting a model having the minimum
description length is also optimal for conducting statistical
estimation that includes model selection.

\subsubsection*{Definition of MDL} The MDL principle can be described
more formally as follows \cite{Rissanen89,Barron91}. For a data
sequence $x^n$ and for a model class ${\cal M}=\{P_{\theta}(X): \theta
\in \Theta(m), m \in M \}$, the minimum description length of the data
with respect to the class is defined as
\begin{equation}\label{eq:formalmdl}
  L_{min} (x^n : {\cal M}) = \min_{m \in M} \inf_{\Delta_n}
\min_{\bar{\theta} \in \Theta_n}
  \left\{- \log P_{\bar{\theta}}(x^n) + l (\bar{\theta}|m) +
l(m)\right\}, \end{equation} where $\Delta_n: \Theta(m) \rightarrow
\Theta_n$ denotes a discretization of $\Theta(m)$ and where
$l(\bar{\theta}|m)$ is the code length for encoding $\bar{\theta} \in
\Theta_n$, satisfying \[ \sum_{\bar{\theta} \in \Theta_n} 2
^{-l(\bar{\theta}|m)} \le 1.  \] Note that `$\inf_{\Delta_n}$' stead
of `$\min_{\Delta_n}$' is used here because there are an infinite
number of points which can serve as a representative for a
cell. Furthermore, $l(m)$ is the code length for encoding $m \in M$,
satisfying \[
 \sum_{m \in M} 2 ^{-l(m)} \le 1. \]
For both data compression and statistical estimation, the best
probability model {\em with respect to the given data} is that which
achieves the minimum description length given in (\ref{eq:formalmdl}).

The minimum description length defined in (\ref{eq:formalmdl}) is also
referred to as the `stochastic complexity' of the data relative to the
model class.

\subsubsection*{Advantages}
MDL offers many advantages as a criterion for statistical
estimation, the most important perhaps being its optimal convergency
rate.

\subsubsection*{Consistency}
The models estimated by MDL converge {\em with probability one} to the
true model when data size increases -- a property referred to as
`consistency' \cite{Barron91}. That means that not only the parameters
themselves but also the number of them converge to those of the true
model.

\subsubsection*{Rate of convergence}
Consistency, however, is a characteristic to be considered only when
data size is large; in practice, when data size can generally be
expected to be small, rate of convergence is a more important guide to
the performance of an estimator.

\namecite{Barron91} have verified that MDL as an estimation strategy
is near optimal in terms of the rate of convergence of its estimated
models to the {\em true} model as the data size increases. {\em When
the true model is included in the class of models considered}, the
models selected by MDL converge in probability to the true model at
the rate of $O(\frac{k^{*} \cdot \log n}{2\cdot n})$, where $k^{*}$ is
the number of parameters in the true model, and $n$ the data
size. This is nearly optimal.

\namecite{Yamanishi92a} has derived an upper bound on the data size
necessary for learning {\em probably approximately correctly} (PAC) a
model from among a class of conditional distributions, which he calls
stochastic rules with finite partitioning. This upper bound is of
order $O(\frac{k^{*}}{\epsilon} \log \frac{k^{*}}{\epsilon} +
\frac{l(m)}{\epsilon})$, where $k^{*}$ denotes the number of
parameters of the true model, and $\epsilon (0 < \epsilon < 1 )$ the
accuracy parameter for the stochastic PAC learning. For MLE, the
corresponding upper bound is of order $O(\frac{k_{max}}{\epsilon} \log
\frac{k_{max}}{\epsilon}+\frac{l(m)}{\epsilon})$, where $k_{max}$
denotes the maximum of the number of parameters of a model in the
model class.  These upper bounds indicate that MDL requires less data
than MLE to achieve the same accuracy in statistical learning,
provided that $k_{max} > k^{*}$ (note that, in general, $k_{max} \ge
k^{*}$).

\subsubsection*{MDL and MLE} When the number of parameters in a
probability model is fixed, and the estimation problem involves only
the estimation of parameters, MLE is known to be satisfactory
\cite{Fisher56}. Furthermore, for such a fixed model, it is known that
MLE is equivalent to MDL: given the data $x^n = x_1 \cdots x_n$, the
maximum likelihood estimator $\hat{\theta}$ is defined as one that
maximizes likelihood with respect to the data, that is,
\begin{equation}\label{eq:mle} \hat{\theta} = \arg \max_{\theta}
P(x^n).  \end{equation} It is easy to see that $\hat{\theta}$ also
satisfies \[ \hat{\theta} = \arg \min_{\theta} - \log P(x^n).  \] This
is, in fact, no more than the MDL estimator in this case, since $ -
\log P_{\hat{\theta}}(x^n)$ is the {\em data description length}.

When the estimation problem involves model selection, MDL's behavior
significantly deviates from that of MLE.  This is because MDL insists
on minimizing the sum total of the data description length {\em and}
the model description length, while MLE is still equivalent to
minimizing the data description length alone. We can, therefore, say
that MLE is a special case of MDL.

Note that in (\ref{eq:mdl}), the first term is of order $O(n)$ and the
second term is of order $O(\log n)$, and thus the first term will
dominate that formula when the data size increases. That means that
when data size is sufficiently large, the MDL estimate will turn out
to be the MLE estimate; otherwise the MDL estimate will be different
from the MLE estimate.

\subsubsection*{MDL and Bayesian Estimation} In an interpretation of
MDL from the viewpoint of Bayesian Estimation, MDL is essentially
equivalent to the `MAP estimation' in Bayesian terminology. Given data
$D$ and a number of models, the Bayesian (MAP) estimator $\hat{M}$ is
defined as one that maximizes posterior probability, i.e.,
\begin{equation}\label{eq:bayes} \begin{array}{lll}
  \hat{M} & = & \arg\max_{M}( P(M|D) ) \\
   & = & \arg\max_{M}( \frac{P(M) \cdot P(D|M)}{P(D)} ) \\
  & = & \arg\max_{M}( P(M) \cdot P(D|M)), \\
\end{array}
 \end{equation} where $P(M)$ denotes the prior probability of model
$M$ and $P(D|M)$ the probability of observing data $D$ through $M$. In
the same way, $\hat{M}$ satisfies \[ \hat{M} = \arg\min_{M}( - \log
P(M) - \log P(D|M)). \] This is equivalent to the MDL estimator if we
take $- \log P(M)$ to be the model description length.  Interpreting
$- \log P(M)$ as the model description length translates, in Bayesian
Estimation, to assigning larger prior probabilities to simpler models,
since it is equivalent to assuming that $P(M) = (\frac{1}{2})^{l(M)}$,
where $l(M)$ is the code length of model $M$.  (Note that if we assign
uniform prior probability to all models, then (\ref{eq:bayes}) becomes
equivalent to (\ref{eq:mle}), giving the MLE estimator.)
 
\subsubsection*{MDL and MEP} The use of the Maximum Entropy Principle
(MEP) has been proposed in statistical language processing
\cite{Ratnaparkhi94,Ratnaparkhi94,Ratnaparkhi97,Berger96,Rosenfeld96}). Like
MDL, MEP is also a learning criterion, one which stipulates that from
among the class of models that satisfies certain constraints, the
model which has the maximum entropy should be selected. Selecting a
model with maximum entropy is, in fact, equivalent to selecting a
model with minimum description length \cite{Rissanen83}. Thus, MDL
provides an information-theoretic justification of MEP.

\subsubsection*{MDL and stochastic complexity} 
The sum of parameter description length and data
description length given in (\ref{eq:mdl}) is still a loose
approximation. Recently, Rissanen has derived this more precise formula:
\begin{equation}\label{eq:formalmdl2}
  l(\theta|m)+l(x^n|m,\theta) =
  - \log P_{\hat{\theta}}(x^n) + \frac{k}{2} \cdot \log \frac{n}{2 \pi} +
\log \int \sqrt{|I(\theta)|} d\theta + o(1), \end{equation} where
$I(\theta)$ denotes the Fisher information matrix, $|{\bf A}|$ the
determinant of matrix ${\bf A}$, and $\pi$ the circular constant, and
$o(1)$ indicates $\lim_{n \rightarrow \infty} o(1) = 0$. It is thus
preferable to use this formula in practice.

This formula can be obtained not only on the basis of the `complete
two-stage code,' but also on that of `quantized maximum likelihood
code,' and has been proposed as the new definition of stochastic
complexity \cite{Rissanen96}. (See also \cite{Clarke90}.)

When the data generation process is i.i.d. and the distribution is a
discrete probability distribution like that in
(\ref{eq:distribution}), the sum of parameter description length and
data description length turns out to be \cite{Rissanen97} \begin{equation}
\label{eq:newmdl} \begin{array}{rl}
  l(\theta)+l(x^n|m,\theta) & = - \sum_{i=1}^{n} \log
  P_{\hat{\theta}}(x_i) + \frac{k}{2} \cdot \log
  \frac{n}{2\cdot\pi} + \log
  \frac{\pi^{(k+1)/2}}{\Gamma(\frac{(k+1)}{2})} + o(1), \\ \end{array}
\end{equation} where $\Gamma$ denotes the Gamma
function\footnote{Euler's Gamma function is defined as
  $\Gamma(x)=\int_0^{\infty}t^{x-1}\cdot e^{-t}dt$.}. This is because
in this case, the determinant of the Fisher information matrix becomes
$\frac{1}{\prod_{i=1}^{k+1}P(i)}$, and the integral of its square root
can be calculated by the Dirichlet's integral as
$\frac{\pi^{(k+1)/2}}{\Gamma(\frac{(k+1)}{2})}$.

\subsection{Employing MDL in NLP}

Recently MDL and related techniques have become popular in natural
language processing and related fields; a number of learning methods
based on MDL have been proposed for various applications
\cite{Ellison91,Ellison92,Cartwright94,Stolcke94,Brent95,Ristad95,Brent96,Grunwald96}.

\subsubsection*{Coping with the data sparseness problem} MDL is a
powerful tool for coping with the data sparseness problem, an inherent
difficulty in statistical language processing.  In general, a
complicated model might be suitable for representing a problem, but it
might be difficult to learn due to the sparseness of training data. On
the other hand, a simple model might be easy to learn, but it might be
not rich enough for representing the problem. One possible way to cope
with this difficulty is to introduce a class of models with various
complexities and to employ MDL to select the model having the most
appropriate level of complexity.

An especially desirable property of MDL is that it takes data size
into consideration. Classical statistics actually assume implicitly
that the data for estimation are always sufficient.  This, however, is
patently untrue in natural language. Thus, the use of MDL might yield
more reliable results in many NLP applications. 

\subsubsection*{Employing efficient algorithms} In practice, the
process of finding the optimal model in terms of MDL is very likely to
be intractable because a model class usually contains too many models
to calculate a description length for each of them. Thus, when we have
modelized a natural language acquisition problem on the basis of a
class of probability models and want to employ MDL to select the best
model, what is necessary to consider next is how to perform the task
efficiently, in other words, how to develop an efficient algorithm.

When the model class under consideration is restricted to one related
to a tree structure, for instance, the dynamic programming technique
is often applicable and the optimal model can be efficiently
found. \namecite{Rissanen97}, for example, has devised such an
algorithm for learning a decision tree.

Another approach is to calculate approximately the description lengths
for the probability models, by using a computational-statistic
technique, e.g., the Markov chain Monte-Carlo method, as is proposed
in \cite{Yamanishi96}.

In this thesis, I take the approach of restricting a model class to a
simpler one (i.e., reducing the number of models to consider) when
doing so is still reasonable for tackling the problem at hand. 

\chapter{Models for Lexical Knowledge Acquisition}

\begin{tabular}{p{5.5cm}r}
 &
\begin{minipage}{10cm}
\begin{tabular}{p{9cm}}
{\em The world as we know it is our interpretation of the observable
  facts in the light of theories that we ourselves invent.}\\
\multicolumn{1}{r}{- Immanuel Kant (paraphrase)} \\
\end{tabular}
\end{minipage}
\end{tabular}
\vspace{0.5cm}

In this chapter, I define probability models for each subproblem of
the lexical semantic knowledge acquisition problem: (1) the hard case
slot model and the soft case slot model; (2) the word-based case frame
model, the class-based case frame model, and the slot-based case frame
model; and (3) the hard co-occurrence model and the soft co-occurrence
model. These are respectively the probability models for (1) case slot
generalization, (2) case dependency learning, and (3) word
clustering.

\section{Case Slot Model}

\subsubsection*{Hard case slot model} We can assume that case slot
data for a case slot for a verb like that shown in
Table~\ref{tab:csdata1} are generated according to a conditional
probability distribution, which specifies the conditional probability
of a noun given the verb and the case slot.  I call such a
distribution a `case slot model.'

When the conditional probability of a noun is defined as that of the
noun class to which the noun belongs, divided by the size of the noun
class, I call the case slot model a `hard-clustering-based case slot
model,' or simply a `hard case slot model.'

Suppose that ${\cal N}$ is the set of nouns, ${\cal V}$ is the set of
verbs, and ${\cal R}$ is the set of slot names. A partition $\Pi$ of
${\cal N}$ is defined as a set satisfying $\Pi \subseteq 2^{\cal
  N}$,\footnote{$2^A$ denotes the power set of a set $A$; if, for
  example, $A=\{a,b\}$, then $2^A=\{\{\},\{a\},\{b\},\{a,b\}\}$.}
$\cup_{C \in \Pi} C = {\cal N}$ and $\forall C_i,C_j \in \Pi, C_i \cap
C_j = \emptyset, (i \not= j)$. An element $C$ in $\Pi$ is referred to as
a `class.' A hard case slot model with respect to a partition $\Pi$ is
defined as a conditional probability distribution:
\begin{equation}\label{eq:hard-csmodel} P(n|v,r) = \frac{1}{|C|} \cdot
P(C|v,r) \ \ \ \ n \in C, \end{equation} where random variable $n$
assumes a value from ${\cal N}$, random variable $v$ from ${\cal V}$,
and random variable $r$ from ${\cal R}$, and where $C \in \Gamma$ is
satisfied.
\footnote{Rigorously, a hard case slot model with respect to a noun
  partition $\Pi$ should be represented as
\[
\begin{array}{l}
  P_{\Pi}(n|v,r) = \sum_{C \in \Pi} P(n|C)\cdot P(C|v,r) \\ P(n|C) =
  \left\{ \begin{array}{ll} \frac{1}{|C|} & n \in C \\ 0 &
    \mbox{otherwise}. \\ 
\end{array} \right. \\
\end{array}
\]}

We can formalize the case slot generalization problem as that of
estimating a hard case slot model. The problem, then, turns out to be
that of selecting a model, from a class of hard case slot models,
which is most likely to have given rise to the case slot data. 

This formalization of case slot generalization will make it possible
to deal with the data sparseness problem, an inherent difficulty in a
statistical approach to natural language processing. Since many words
in natural language are synonymous, it is natural to classify them
into the same word class and employ class-based probability models. A
class-based model usually has far fewer parameters than a word-based
model, and thus the use of it can help handle the data sparseness
problem. An important characteristic of the approach taken here is
that it automatically conducts the optimization of word clustering by
means of statistical model selection. That is to say, neither the
number of word classes nor the way of word classification are
determined in advance, but are determined automatically on the basis
of the input data.

The uniform distribution assumption in the hard case slot model seems
to be necessary for dealing with the data sparseness problem. If we
were to assume that the distribution of words (nouns) within a class
is a word-based distribution, then the number of parameters would not
be reduced and the data sparseness problem would still prevail.

Under the uniform distribution assumption, generalization turns out to
be the process of finding the best configuration of classes such that
the words in each class are equally likely to be the value of the slot
in question. (Words belonging to a single word class should be similar
in terms of likelihood; they do not necessarily have to be synonyms.)
Conversely, if we take the generalization to be such a process, then
viewing it as statistical estimation of a hard case slot model seems
to be quite appropriate, because the class of hard case slot models
contains all of the possible models for the purposes of
generalization. The word-based case slot model (i.e., one in which
each word forms its own word class) is a (discrete) hard case slot
model, and any grouping of words (nouns) leads to one (discrete) hard
case slot model.

\subsubsection*{Soft case slot model} Note that in the hard case slot
model a word (noun) is assumed to belong to a single class.  In
practice, however, many words have sense ambiguities and a word can
belong to several different classes, e.g., `bird' is a member of both
$\langle$animal$\rangle$ and $\langle$meat$\rangle$. It is also
possible to extend the hard case slot model so that each word
probabilistically belongs to several different classes, which would
allow us to resolve both syntactic and word sense ambiguities at the
same time. Such a model can be defined in the form of a `finite
mixture model,' which is a linear combination of the word probability
distributions within individual word (noun) classes. I call such a
model a `soft-clustering-based case slot model,' or simply a `soft
case slot model.'

First, a covering $\Gamma$ of the noun set ${\cal N}$ is defined as a
set satisfying $\Gamma \subseteq 2^{\cal N}$, $\cup_{C \in \Gamma} C =
{\cal N}$. An element $C$ in $\Gamma$ is referred to as a `class.' A
soft case slot model with respect to a covering $\Gamma$ is defined as
a conditional probability distribution:
\begin{equation}\label{eq:soft-csmodel} P(n|v,r) = \sum_{C \in \Gamma}
P(C|v,r) \cdot P(n|C) \end{equation} where random variable $n$ denotes
a noun, random variable $v$ a verb, and random variable $r$ a slot
name. We can also formalize the case slot generalization problem as
that of estimating a soft case slot model.

If we assume, in a soft case slot model, that a word can only belong
to a single class alone and that the distribution within a class is a
uniform distribution, then the soft case slot model will become a hard
case slot model.

\subsubsection*{Numbers of parameters} Table~\ref{tab:cspara} shows
the numbers of parameters in a word-based case slot model
(\ref{eq:word-csmodel}), a hard case slot model
(\ref{eq:hard-csmodel}), and a soft case slot model
(\ref{eq:soft-csmodel}). Here $N$ denotes the size of the set of
nouns, $\Pi$ the partition in the hard case slot model, and
$\Gamma$ the covering in the soft case slot model.

\begin{table}[htb] 
\caption{Numbers of parameters in case slot models.}
\label{tab:cspara} 
\begin{center} 
\begin{tabular}{|lc|} \hline 
word-based model & $O(N)$ \\ 
hard case slot model & $O(|\Pi|)$ \\ 
soft case slot model & $O(|\Gamma| + \sum_{C \in \Gamma} |C|)$ \\ \hline 
\end{tabular}
\end{center}
\end{table}

The number of parameters in a hard case slot models is generally
smaller than that in a soft case slot model. Furthermore, the number
of parameters in a soft case slot model is generally smaller than that
in a word-based case slot model (note that the parameters $P(n|C)$ is
common to each soft case slot model). As a result, hard case slot
models require less data for parameter estimation than soft case slot
models, and soft case slot models less data than word-based case slot
models. That is to say, hard and soft case slot models are more useful
than word-based models, given the fact that usually the size of data
for training is small.

Unfortunately, currently available data sizes are still insufficient
for the accurate estimating of a soft case slot model. 
(Appendix~\ref{append:fmm} shows a method for learning a soft case
slot model.) (See \cite{Li97b} for a method of using a finite mixture
model in document classification, for which more data are generally
available.)

In this thesis, I address only the issue of estimating a hard case
slot model. With regard to the word-sense ambiguity problem, one can
employ an existing word-sense disambiguation technique (cf., Chapter2)
in pre-processing, and use the disambiguated word senses as virtual
words in the subsequent learning process.

\section{Case Frame Model}

We can assume that case frame data like that in
Table~\ref{tab:cfdata1} are generated according to a multi-dimensional
discrete joint probability distribution in which random variables
represent case slots. I call such a distribution a `case frame model.'
We can formalize the case dependency learning problem as that of
estimating a case frame model. The dependencies between case slots are
represented as {\em probabilistic} dependencies between random
variables. (Recall that random variables $X_1,\cdots, X_n$ are {\em
mutually} independent, if for any $k\le n$, and any $1 \le i_1 <
\cdots < i_k \le n$, $P(X_{i_1},\cdots,X_{i_k}) = P(X_{i_1})\cdots
P(X_{i_k})$; otherwise, they are {\em mutually} dependent.)

The case frame model is the joint probability
distribution of type,
\[
P_{Y}(X_1,X_2,\cdots,X_n),
\]
where index $Y$ stands for the verb, and each of the random variables
$X_i, i=1,2,\cdots,n,$ represents a case slot. 

In this thesis, `case slots' refers to {\em surface} case slots, but
they can also be {\em deep} case slots. Furthermore, obligatory cases
and optional cases are uniformly treated.  The possible case slots can
vary from verb to verb.  They can also be a predetermined set for all
of the verbs, with most of the slots corresponding to (English)
prepositions.

\begin{table}[htb]
\caption{Example case frame data generated by a word-based model.}
\label{tab:cfdata1b}
\begin{center}
\begin{tabular}{|lc|} \hline
Case frame & Frequency \\ \hline
(fly (arg1 girl)(arg2 jet)) & 2 \\
(fly (arg1 boy)(arg2 helicopter)) & 1 \\
(fly (arg1 company)(arg2 jet)) & 2 \\
(fly (arg1 girl)(arg2 company)) & 1 \\ 
(fly (arg1 boy)(to Tokyo)) & 1 \\ 
(fly (arg1 girl)(from Tokyo) (to New York)) & 1 \\ 
(fly (arg1 JAL)(from Tokyo) (to Bejing)) & 1 \\ 
\hline
\end{tabular}
\end{center}
\end{table}

\begin{table}[htb]
\caption{Example case frame data generated by a class-based model.}
\label{tab:cfdata2}
\begin{center}
\begin{tabular}{|lc|} \hline
Case frame & Frequency \\ \hline
(fly (arg1 $\langle$person$\rangle$)(arg2 $\langle$airplane$\rangle$)) & 3 \\
(fly (arg1 $\langle$company$\rangle$)(arg2 $\langle$airplane$\rangle$)) & 2 \\
(fly (arg1 $\langle$person$\rangle$)(arg2 $\langle$company$\rangle$)) & 1 \\
(fly (arg1 $\langle$person$\rangle$)(to $\langle$place$\rangle$)) & 1 \\
(fly (arg1 $\langle$person$\rangle$)(from $\langle$place$\rangle$)(to $\langle$place$\rangle$)) & 1 \\
(fly (arg1 $\langle$company$\rangle$)(from $\langle$place$\rangle$)(to $\langle$place$\rangle$)) & 1 \\ \hline
\end{tabular}
\end{center}
\end{table}

\begin{table}[htb]
\caption{Example case frame data generated by a slot-based model.}
\label{tab:cfdata3}
\begin{center}
\begin{tabular}{|lc|} \hline
Case frame & Frequency \\ \hline
(fly (arg1 1)(arg2 1)) & 6 \\
(fly (arg1 1)(to 1)) & 1 \\
(fly (arg1 1)(from 1)(to 1)) & 2 \\ \hline
\end{tabular}
\end{center}
\end{table}

The case frame model can be further classified into three types of
probability models according to the type of value each random variable
$X_i$ assumes. When $X_i$ assumes a word or a special symbol $0$ as
its value, the corresponding model is referred to as a `word-based
case frame model.' Here $0$ indicates the absence of the case slot in
question. When $X_i$ assumes a word-class (such as
$\langle$person$\rangle$ or $\langle$company$\rangle$) or $0$ as its
value, the corresponding model is referred to as a `class-based case
frame model.' When $X_i$ takes on $1$ or $0$ as its value, the model
is called a `slot-based case frame model.' Here $1$ indicates the
presence of the case slot in question, and $0$ the absence of it. For
example, the data in Table~\ref{tab:cfdata1b} could have been
generated by a word-based model, the data in Table~\ref{tab:cfdata2}
by a class-based model, where $\langle \cdots \rangle$ denotes a word
class, and the data in Table~\ref{tab:cfdata3} by a slot-based
model. Suppose, for simplicity, that there are only 4 possible case
slots corresponding, respectively, to subject, direct object, `from'
phrase, and `to' phrase. Then, \[ \begin{array}{l}
  P_{\rm fly}(X_{\rm arg1}={\rm girl},X_{\rm arg2}={\rm jet}, X_{\rm from}=0,X_{\rm to}=0)
\end{array}
\]
is specified by a word-based case frame model. In
contrast,
\[
\begin{array}{l}
  P_{\rm fly}(X_{\rm arg1}=\langle{\rm person}\rangle,X_{\rm arg2}=\langle{\rm
    airplane} \rangle,X_{\rm from}=0,X_{\rm to}=0)
\end{array}
\]
is specified by a class-based case frame model, where
$\langle{\rm person}\rangle$ and $\langle{\rm airplane}\rangle$ denote
word classes. Finally,
\[
\begin{array}{l}
  P_{\rm fly}(X_{\rm arg1}=1,X_{\rm arg2}=1,X_{\rm from}=0,X_{\rm to}=0)
\end{array}
\]
is specified by a slot-based case frame model. One can also
define a combined model in which, for example, some random variables
assume word classes and 0 as their values while others assume 1 and 0.

Note that since in general
\[
\begin{array}{l}
  P_{\rm fly}(X_{\rm arg1}=1,X_{\rm arg2}=1,X_{\rm from}=0,X_{\rm to}=0) \\
  \not= P_{\rm fly}(X_{\rm arg1}=1,X_{\rm arg2}=1),
  \end{array} \] one should not use here the joint probability
$P_{\rm fly}(X_{\rm arg1}=1,X_{\rm arg2}=1)$ as the probability of the case frame
`(fly (arg1 1)(arg2 1)).'

In learning and using of the case frame models, it is also assumed
that word sense ambiguities have been resolved in pre-processing.

One may argue that when the ambiguities of a verb are resolved, there
would not exist case dependencies at all (cf., `fly' in sentences of
(\ref{eq:fly})). Sense ambiguities, however, are generally difficult
to define precisely. I think that it is preferable not to resolve them
until doing so is necessary in a particular application. That is to
say, I think that, in general, case dependencies do exist and the
development of a method for learning them is needed.

\subsubsection*{Numbers of parameters}
Table~\ref{tab:cfpara} shows the numbers of parameters in a word-based
case frame model, a class-based case frame model, and a slot-based
case frame model, where $n$ denotes the number of random variables,
$N$ the size of the set of nouns, and $k_{max}$ the maximum number of
classes in any slot.

\begin{table}[htb] 
\caption{Numbers of parameters in case frame models.}
\label{tab:cfpara} 
\begin{center} 
\begin{tabular}{|lc|} \hline 
word-based case frame model & $O(N^n)$ \\ 
class-based case frame model & $O(k_{max}^n )$ \\ 
slot-based case frame model & $O(2^n)$ \\ \hline 
\end{tabular}
\end{center} 
\end{table}

\section{Co-occurrence Model}

\subsubsection*{Hard co-occurrence model} We can assume that
co-occurrence data over a set of nouns and a set of verbs like that in
Figure~\ref{fig:co-occurrence} are generated according to a joint
probability distribution that specifies the co-occurrence
probabilities of noun verb pairs. I call such a distribution a
`co-occurrence model.'

I call the co-occurrence model a `hard-clustering-based co-occurrence
model,' or simply a `hard co-occurrence model,' when the joint
probability of a noun verb pair can be defined as the product of the
joint probability of the noun class and the verb class to which
the noun and the verb respectively belong, the conditional probability
of the noun given its noun class, and the conditional probability of
the verb given its verb class.

\begin{figure}[htb]
\begin{center}
\epsfxsize8cm\epsfysize8cm\epsfbox{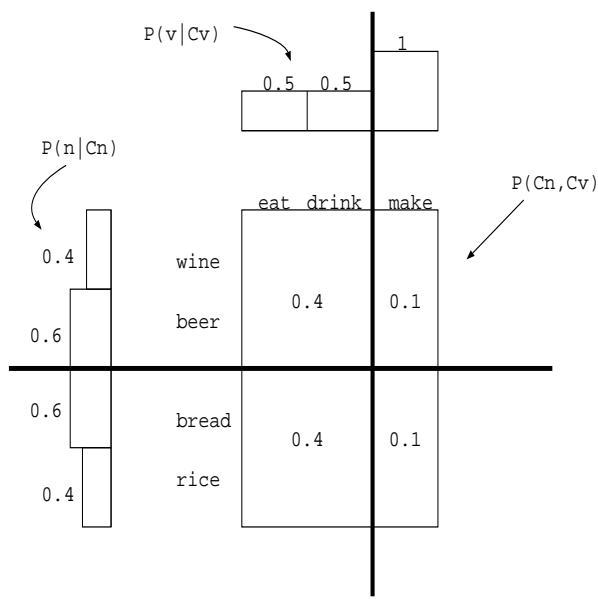}
\end{center}
\caption{An example hard co-occurrence model.}
\label{fig:hardmodel} 
\end{figure}

Suppose that ${\cal N}$ is the set of nouns, and ${\cal V}$ is the set
of verbs. A partition $\Pi_n$ of ${\cal N}$ is defined as a set which
satisfies $\Pi_n \subseteq 2^{\cal N}$, $\cup_{C_n \in \Pi_n} C_n =
{\cal N}$ and $\forall C_i, C_j \in \Pi_n, C_i\cap C_j =\emptyset, (i
\not= j)$. A partition $\Pi_v$ of ${\cal V}$ is defined as a set which
satisfies $\Pi_v \subseteq 2^{\cal V}$, $\cup_{C_v \in \Pi_v} C_v =
{\cal V}$ and $\forall C_i, C_j \in \Pi_v, C_i\cap C_j =\emptyset, (i
\not= j)$. Each element in a partition forms a `class' of words. I
define a hard co-occurrence model with respect to a noun partition
$\Pi_n$ and a verb partition $\Pi_v$ as a joint probability
distribution of type:
\begin{equation}\label{eq:hardmodel} P(n,v) = P(C_n,C_v) \cdot
  P(n|C_n) \cdot P(v|C_v) \ \ \ \ n \in C_n, v \in C_v,
\end{equation} where random variable $n$ denotes a noun and random
variable $v$ a verb and where $C_n \in \Pi_n$ and $C_v \in \Pi_v$ are
satisfied. 
\footnote{
Rigorously, a hard co-occurrence model with respect to a noun
partition $\Pi_n$ and a verb partition $\Pi_v$ should be represented as
\[
\begin{array}{l}
P_{\Pi_n\Pi_v}(n,v) = \sum_{C_n \in \Pi_n, C_v \in \Pi_v} P(C_n,C_v) \cdot
P(n|C_n) \cdot P(v|C_v) \\
P(x|C_x) = \left\{ \begin{array}{ll}
Q(x|C_x) & x \in C_x \\
0 & \mbox{otherwise} \\
\end{array} \right. (x=n,v) \\
\forall C_x, \sum_{x \in C_x} Q(x|C_x) = 1, (x=n,v).
\end{array}
\]
}
Figure~\ref{fig:hardmodel} shows a hard co-occurrence
model, one that can give rise to the co-occurrence data in
Figure~\ref{fig:co-occurrence}.

Estimating a hard co-occurrence model means selecting, from the class
of such models, one that is most likely to have given rise to the
co-occurrence data. The selected model will contain a hard clustering
of words. We can therefore formalize the problem of word clustering as
that of estimating a hard co-occurrence model.

We can restrict the hard co-occurrence model by assuming that words
within a same class are generated with an equal probability
\cite{Li96b,Li97a}, obtaining \[ P(n,v) = P(C_n,C_v) \cdot
\frac{1}{|C_n|} \cdot \frac{1}{|C_v|} \ \ \ \ n \in C_n, v \in C_v. \]
Employing such a restricted model in word clustering,
however, has an undesirable tendency to result in classifying into
different classes those words that have similar co-occurrence
patterns but have different absolute frequencies.

The hard co-occurrence model in (\ref{eq:hardmodel}) can also be
considered an extension of that proposed in \cite{Brown92}. First,
dividing the equation by $P(v)$, we obtain \[ \frac{P(n,v)}{P(v)} =
P(C_n|C_v) \cdot P(n|C_n) \cdot \left(\frac{P(C_v) \cdot
P(v|C_v)}{P(v)}\right) \ \ \ \ n \in C_n, v \in C_v. \] Since
$\frac{P(C_v) \cdot P(v|C_v)}{P(v)}=1$ holds, we have \[ P(n|v) =
P(C_n|C_v) \cdot P(n|C_n) \ \ \ \ n \in C_n, v \in C_v. \] We can
rewrite the model for word sequence predication as
\begin{equation}\label{eq:class-bigram} P(w_2|w_1) = P(C_2|C_1) \cdot
P(w_2|C_2) \ \ \ \ w_1 \in C_1, w_2 \in C_2, \end{equation} where
random variables $w_1$ and $w_2$ take on words as their values.  In
this way, the hard co-occurrence model turns out to be a bigram class
model and is similar to that proposed in \cite{Brown92} (cf., Chapter
2).\footnote{Strictly speaking, the bigram class model proposed
by \cite{Brown92} and the hard case slot model defined here are
different types of probability models; the former is a conditional
distribution, while the latter is a joint distribution.} The difference
is that the model in (\ref{eq:class-bigram}) assumes that the
configuration of word groups for $C_2$ and the configuration of word
groups for $C_1$ can be different, while Brown et al's model assumes
that the configurations for the two are always the same.

\subsubsection*{Soft co-occurrence model} The co-occurrence model can
also be defined as a double mixture model, which is a double linear
combination of the word probability distributions within individual
noun classes and those within individual verb classes. I call such a
model a `soft-clustering-based co-occurrence model,' or simply `soft
co-occurrence model.'

First, a covering $\Gamma_n$ of the noun set ${\cal N}$ is defined as
a set which satisfies $\Gamma_n \subseteq 2^{\cal
  N}$, $\cup_{C_n \in \Gamma_n} C_n = {\cal N}$. A covering $\Gamma_v$
of the verb set ${\cal V}$ is defined as a set which satisfies
$\Gamma_v \subseteq 2^{\cal V}$, $\cup_{C_v \in \Gamma_v} C_v = {\cal
V}$. Each element in a covering is referred to as a `class.' I
define a soft co-occurrence model with respect to a noun covering
$\Gamma_n$ and a verb covering $\Gamma_v$ as a joint probability
distribution of type: \[ P(n,v) = \sum_{C_n \in \Gamma_n} \sum_{C_v
\in \Gamma_v} P(C_n,C_v) \cdot P(n|C_n) \cdot P(v|C_v), \] where
random variable $n$ denotes a noun and random variable $v$ a
verb. Obviously, the soft co-occurrence model includes the hard
co-occurrence model as a special case.

If we assume that a verb class consists of a single verb alone,
i.e., $\Gamma_v = \{\{v\}| v \in {\cal V} \}$, then the soft
co-occurrence model turns out to be \[ P(n,v) = \sum_{C_n \in \Gamma_n}
P(C_n,v) \cdot P(n|C_n), \] which is equivalent to that proposed in
\cite{Pereira93}.

Estimating a soft co-occurrence model, thus, means selecting, from the
class of such models, one that is most likely to have given rise to
the co-occurrence data. The selected model will contain a soft
clustering of words. We can formalize the word clustering problem as
that of estimating a soft co-occurrence model.

\subsubsection*{Numbers of parameters} Table~\ref{tab:clupara} shows
the numbers of parameters in a hard co-occurrence model and in a soft
co-occurrence model. Here $N$ denotes the size of the set of nouns,
$V$ the size of the set of verbs, $\Pi_n$ and $\Pi_v$ are the
partitions in the hard co-occurrence model, and $\Gamma_n$ and
$\Gamma_v$ are the coverings in the soft co-occurrence model.

\begin{table}[htb] 
 \caption{Numbers of parameters in co-occurrence models.}
\label{tab:clupara} 
\begin{center}
\begin{tabular}{|lc|} \hline 
hard co-occurrence model & $O(|\Pi_n| \cdot |\Pi_v| + V + N)$ \\ 
soft co-occurrence model & $O(|\Gamma_n| \cdot |\Gamma_v| + \sum_{C_n \in \Gamma_n} |C_n| + \sum_{C_v \in \Gamma_v} |C_v|)$ \\ \hline 
\end{tabular}
\end{center} 
\end{table}

In this thesis, I address only the issue of estimating a hard
co-occurrence model.

\section{Relations between Models}

Table~\ref{tab:formalization} summarizes the formalization I have made
above.

\begin{table}[htb] 
 \caption{Summary of the formalization.}
\label{tab:formalization} 
\begin{center}
\begin{tabular}{|lll|} \hline 
Input & Output & Side effect \\ \hline
case slot data & hard/soft case slot model & case slot generalization \\
case frame data & word/class/slot-based case frame model & case dependency learning \\
co-occurrence data & hard/soft co-occurrence model & word clustering \\ \hline
\end{tabular}
\end{center} 
\end{table}

The models described above are closely related. The soft case slot
model includes the hard case slot model, and the soft co-occurrence model
includes the hard co-occurrence model. The slot-based case frame model
will become the class-based case frame model when we granulate
slot-based case slot values into class-based slot values. The
class-based case frame model will become the word-based case frame
model when we perform further granulation. The relation between the
hard case slot model and the case frame models, that between the hard
case slot model and the hard co-occurrence model, and that between the
soft case slot model and the soft co-occurrence model are described
below.

\begin{figure}[htb]
\begin{center}
\epsfxsize8cm\epsfysize8cm\epsfbox{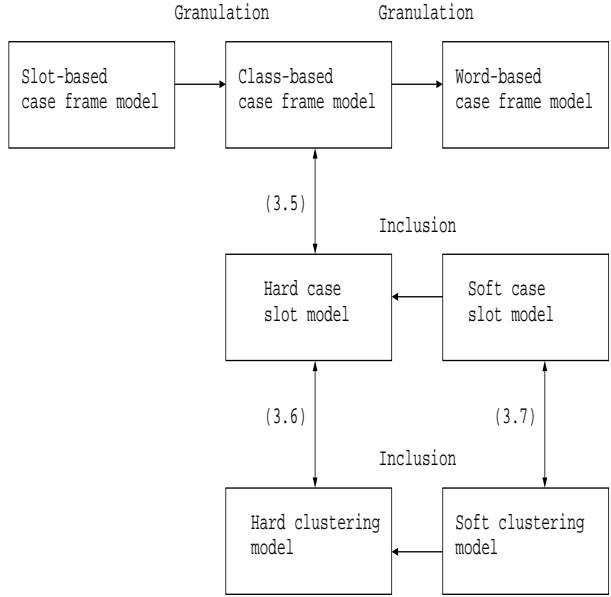}
\end{center}
\caption{Relations between models.}
\label{fig:modelrel} 
\end{figure}

\subsubsection*{Hard case slot model and case frame models} The
relationship between the hard case slot model and the case frame
models may be expressed by transforming the notation of the
conditional probability $P(C|v,r)$ in the hard case slot model to
\begin{equation}\label{eq:hardcscmrel}
P(C|v,r)=P_{v}(X_{r}=C|X_{r}=1)=\frac{P_v(X_r=C)}{P_v(X_r=1)},
\end{equation} which is the ratio between a marginal probability in
the class-based case frame model and a marginal probability in the
slot-based case frame model.

This relation (\ref{eq:hardcscmrel}) implies that we can generalize
case slots by using the hard case slot model and then acquire
class-based case frame patterns by using the class-based case frame
model.

\subsubsection*{Hard case slot model and hard co-occurrence model}
If we assume that the verb set consists of a single verb alone, then the
hard co-occurrence model {\em with respect to slot $r$} becomes \[
P_{r}(n,v) = P_{r}(C_n,v) \cdot P_{r}(n|C_n) \ \ \ \ n \in C_n.  \]
If we further assume that nouns within a same noun class have an equal
probability, then we have
\begin{equation}\label{eq:hardcsclurel}
\frac{P_{r}(n,v)}{P_r(v)} = P_r(C_n|v) \cdot \frac{1}{|C_n|} \ \ \ \
n \in C_n.  
\end{equation} 
This is no more than the hard case slot model, which has a
different notation.

\subsubsection*{Soft case slot model and soft co-occurrence model} If we
assume that the verb set consists of a single verb alone, then the
soft co-occurrence model {\em with respect to slot $r$} becomes 
\[
P_{r}(n,v) = \sum_{C_n \in \Gamma_n} P_{r}(C_n,v) \cdot P_{r}(n|C_n).
\] 
Suppose that $P_{r}(n|C_n)$ is common to each slot $r$, then we can denote
it as $P(n|C_n)$ and have
\begin{equation}\label{eq:softcsclurel}
\frac{P_{r}(n,v)}{P_r(v)} = \sum_{C_n \in \Gamma_n} P_{r}(C_n|v) \cdot
P(n|C_n).
\end{equation}
This is equivalent to the soft case slot model.

\section{Discussions}

\subsubsection*{Generation models v.s. decision models}
The models defined above are what I call `generation models.' A
case frame generation model is a probability distribution that gives
rise to a case frame with a certain probability. In disambiguation, a
generation model predicts the {\em likelihood of the occurrence of
each case frame}.

Alternatively, we can define what I call `decision models' to
perform the disambiguation task. A decision model is a
conditional distribution which represents the conditional
probabilities of disambiguation (or parsing) decisions. For instance,
the decision tree model and the decision list model are example
decision models (cf., Chapter 2).  In disambiguation, a
decision model predicts the {\em
  likelihood of the correctness of each decision}.

A generation model can generally be represented as a joint
distribution $P({\bf X})$ (or a conditional distribution $P({\bf
X}|.)$), where random variables ${\bf X}$ denote linguistic
(syntactical and/or lexical) features. A decision model can
generally be represented by a conditional distribution $P(Y|{\bf X})$
where random variables ${\bf X}$ denote linguistic features and random
variable $Y$ denotes usually a {\em small} number of decisions.

Estimating a generation model requires merely positive examples. On
the other hand, estimating a decision model requires both positive and
negative examples.

A case frame generation model can be used for purposes other than
structural disambiguation. A decision model, on the other hand,
is defined specifically for the purpose of disambiguation.

In this thesis, I investigate generation models because of their
important generality.

The case slot models are, in fact, `one-dimensional lexical generation
models,' the co-occurrence models are `two-dimensional lexical
generation models,' and the case frame models are `multi-dimensional
lexical generation models.' Note that the case frame models are not
simply straightforward extensions of the case slot models and the
co-occurrence models; one can easily define different
multi-dimensional models as extensions of the case slot models and the
co-occurrence models (from one or two dimensions to multi-dimensions).

\subsubsection*{Linguistic models}
The models I have so far defined can also be considered to
be linguistic models in the sense that they straightforwardly
represent case frame patterns (or selectional patterns,
subcategorization patterns) proposed in the linguistic theories of
\cite{Fillmore68,Katz63,Pollard87}. In other words, they are generally
intelligible to humans, because they contain descriptions of language
usage.

\subsubsection*{Probability distributions v.s. probabilistic measures}
An alternative to defining probability distributions for lexical
knowledge acquisition, and consequently for disambiguation, is to
define probabilistic measures (e.g., the association ratio, the
selectional association measure). Calculating these measures in a
theoretically sound way can be difficult, however, and needs further
investigation.

The methods commonly employed to calculate the association ratio
measure (cf., Chapter 2) are based on heuristics. For example, it is
calculated as \[ \hat{S}(n|v,r) = \log
\frac{\hat{P}(n|v,r)}{\hat{P}(n)}, \] where $\hat{P}(n|v,r)$ and
$\hat{P}(n)$ denote, respectively, the Laplace estimates of the
probabilities $P(n|v,r)$ and $P(n)$. Here, each of the two estimates
can only be calculated with a certain degree of precision which
depends on the size of training data. Any small inaccuracies in the
two may be greatly magnified when they are calculated as a ratio, and
this will lead to an extremely unreliable estimate of $S(n|v,r)$ (note
that association ratio is an unbounded measure). Since training data
is always insufficient, this phenomenon may occur very
frequently. Unfortunately, a theoretically sound method of calculation
has yet to be developed.

Similarly, a theoretically sound method for calculating the
selectional association measure also has yet to be developed. (See
\cite{Abe96} for a heuristic method for learning a similar measure on
the basis of the MDL principle.) In this thesis I employ probability
distributions rather than probabilistic measures.

\section{Disambiguation Methods}

The models proposed above can be independently used for disambiguation
purposes, they can also be combined into a single natural language
analysis system. In this section, I first describe how they can be
independently used and then how they can be combined.

\subsubsection*{Using case frame models} Suppose for example that in
the analysis of the sentence \[ \mbox{The girl will fly a jet from
Tokyo,} \] the following alternative interpretations are obtained.  
\[ \mbox{(fly (arg1 {\rm girl}) (arg2 ({\rm jet})) (from {\rm Tokyo}))}
\] \[ \mbox{(fly (arg1 {\rm girl}) (arg2 ({\rm jet} (from {\rm
Tokyo}))))}.  \] We wish to select the more appropriate of the two
interpretations.  Suppose for simplicity that there are four possible
case slots for the verb `fly,' and there is only one possible case
slot for the noun `jet.' A disambiguation method based on word-based
case frame models would calculate the following likelihood values and
select the interpretation with higher likelihood value: \[ P_{\rm
fly}(X_{\rm arg1}={\rm girl},X_{\rm arg2}={\rm jet},X_{\rm from}={\rm
  Tokyo},X_{\rm to}=0)\cdot
P_{{\rm jet}}(X_{\rm from}=0)
\]
and
\[
P_{\rm fly}(X_{\rm arg1}={\rm girl},X_{\rm arg2}={\rm jet},X_{\rm from}=0,X_{\rm to}=0)\cdot P_{{\rm jet}}(X_{\rm from}={\rm Tokyo}).
\]
If the former is larger than the latter, we select the former
interpretation, otherwise we select the latter interpretation.

If we assume here that case slots are independent, then we need only
compare 
\[
P_{\rm fly}(X_{\rm from}={\rm Tokyo})\cdot P_{{\rm jet}}(X_{\rm from}=0)
\]
and
\[
P_{\rm fly}(X_{\rm from}=0)\cdot P_{{\rm jet}}(X_{\rm from}={\rm Tokyo}).
\]

Similarly, when the models are slot-based and the case slots are
assumed to be independent, we need only compare
\[
\begin{array}{l}
P_{\rm fly}(X_{\rm from}=1)\cdot P_{{\rm jet}}(X_{\rm from}=0) \\
= P_{\rm fly}(X_{\rm from}=1)\cdot \biggl(1-P_{{\rm jet}}(X_{\rm from}=1)\biggr)
\end{array}
\]
and
\[
\begin{array}{l}
P_{\rm fly}(X_{\rm from}=0)\cdot P_{{\rm jet}}(X_{\rm from}=1) \\
= \biggl(1-P_{\rm fly}(X_{\rm from}=1)\biggr) \cdot P_{{\rm jet}}(X_{\rm from}=1). \\
\end{array}
\]
That is to say, we need noly compare
\[
P_{\rm fly}(X_{\rm from}=1)
\]
and
\[
P_{{\rm jet}}(X_{\rm from}=1).
\]
The method proposed by \namecite{Hindle91} in fact compares the same
probabilities; they do it by means of statistical hypothesis
testing.

\subsubsection*{Using hard case slot models}
Another way of conducting disambiguation under the assumption that
case slots are independent is to employ the hard case slot model.
Specifically we compare
\[
P({\rm Tokyo}|{\rm fly},{\rm from})
\]
and
\[
P({\rm Tokyo}|{\rm jet},{\rm from}).
\]
If the former is larger than the latter, we select the former
interpretation, otherwise we select the latter interpretation.

\subsubsection*{Using hard co-occurrence models}
We can also use the hard co-occurrence model to perform the
disambiguation task, under the assumption that case slots are
independent. Specifically, we compare
\[
P_{\rm from}({\rm Tokyo}|{\rm fly}) = \frac{P_{\rm from}({\rm Tokyo},{\rm fly})}{\sum_{n \in {\cal N}} P_{\rm from}(n,{\rm fly})}
\]
and
\[
P_{\rm from}({\rm Tokyo}|{\rm jet}) = \frac{P_{\rm from}({\rm Tokyo},{\rm jet})}{\sum_{n \in {\cal N}} P_{\rm from}(n,{\rm jet})}.
\]
Here, $P_{\rm from}({\rm Tokyo},{\rm fly})$ is calculated on the basis
of a hard co-occurrence model over the set of nouns and the set of verbs
with respect to the `from' slot, and $P_{\rm from}({\rm Tokyo},{\rm
jet})$ on the basis of a hard co-occurrence model over the set of nouns
with respect to the `from' slot.

Since the joint probabilities above are all estimated on the basis of
class-based models, the conditional probabilities are in fact
calculated on the basis of not only the co-occurrences of the related
words but also of those of similar words. That means that this
disambiguation method is similar to the similarity-based approach
(cf., Chapter 2). The difference is that the method described here is
based on a probability model, while the similarity-based approach
usually is based on heuristics.

\subsubsection*{A combined method}
Let us next consider a method based on combination of the above
models.

We first employ the hard co-occurrence model to construct a thesaurus
for each case slot (we can, however, construct only thesauruses for
which there are enough co-occurrence data with respect to the
corresponding case slots). We next employ the hard case slot model to
generalize values of case slots into word classes (word classes used
in a hard case slot model can be either from a hand-made thesaurus or
from an automatically constructed thesaurus; cf., Chapter 4). Finally,
we employ the class-based case frame model to learn class-based case
frame patterns.

In disambiguation, we refer to the case frame patterns, calculate
likelihood values for the ambiguous case frames, and select the most
likely case frame as output.

With regard to the above example, we can calculate and compare the
following likelihood values: \[ \begin{array}{l} L(1) = P_{\rm
fly}(X_{\rm arg1}=\langle {\rm person} \rangle, X_{\rm arg2}=\langle
{\rm airplane} \rangle, X_{\rm from}= \langle {\rm place} \rangle )
\cdot
  P_{\rm jet}(X_{\rm from}=0) \\
\end{array}
\]
and
\[
\begin{array}{l}
L(2) = P_{\rm fly}(X_{\rm arg1}=\langle {\rm person} \rangle, X_{\rm arg2}=\langle {\rm airplane} \rangle, 
X_{\rm from}=0)\cdot 
  P_{\rm jet}(X_{\rm from}=\langle {\rm place} \rangle), \end{array}
\] assuming that there are only three case slots: arg1, arg2 and
`from' for the verb `fly,' and there is one case slot: `from' for the
noun `jet.'  Here $\langle \cdots \rangle$ denotes a word class. We
make the pp-attachment decision as follows: if $L(1) > L(2) $, we
attach the phrase `from Tokyo' to `fly;' if $L(1) < L(2)$, we attach
it to `jet;' otherwise we make no decision.

Unfortunately, it is still difficult to attain high performance with
this method at the current stage of statistical language processing,
since the corpus data currently available is far less than that
necessary to estimate accurately the class-based case frame models.

\section{Summary}

I have proposed the soft/hard case slot model for case slot
generalization, the word-based/class-based/slot-based case frame model
for case dependency learning, and the soft/hard co-occurrence model
for word clustering. In Chapter 4, I will describe a method
for learning the hard case slot model, i.e., generalizing case slots;
in Chapter 5, a method for learning the case frame model, i.e.,
learning case dependencies; and in Chapter 6, a method for learning
the hard co-occurrence model, i.e., conducting word clustering. In
Chapter 7, I will describe a disambiguation method, which is based on
the learning methods proposed in Chapters 4 and 6. (See
Figure~\ref{fig:thesisorg}.)

\chapter{Case Slot Generalization}

\begin{tabular}{p{5.5cm}r}
 &
\begin{minipage}{10cm}
\begin{tabular}{p{9cm}}
{\em Make everything as simple as possible - but not simpler.}\\
\multicolumn{1}{r}{- Albert Einstein} \\
\end{tabular}
\end{minipage}
\end{tabular}
\vspace{0.5cm}

In this chapter, I describe one method for learning the hard case slot
model, i.e., generalizing case slots.

\section{Tree Cut Model}\label{sec:treecut}

As described in Chapter 3, we can formalize the case slot
generalization problem into that of estimating a conditional
probability distribution referred to as a `hard case slot model.' The
problem thus turns to be that of selecting the best model from among
all possible hard case slot models. Since the number of partitions for
a set of nouns is very large, the number of such models is very large,
too. The problem of estimating a hard case slot model, therefore, is
most likely intractable. (The number of partitions for a set of nouns
is $\sum_{i=1}^{N} \sum_{j=1}^{i} \frac{(-1)^{i-j} \cdot j^N}{(i-j)!
  \cdot j!}$, where $N$ is the size of the set of nouns (cf.,
\cite{Knuth73}), and is roughly of order $O(N^{N})$.)

\setlength{\unitlength}{0.0125in}%
\begin{figure}[htb]
\begin{center}
\epsfxsize7.5cm\epsfysize2.5cm\epsfbox{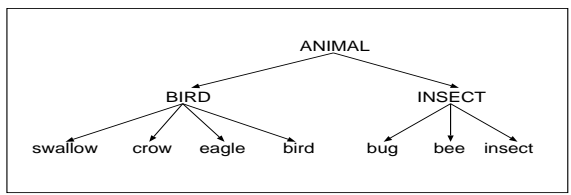}
\end{center}
\caption{An example thesaurus.}
\label{fig:thesau}
\end{figure}

To deal with this difficulty, I take the approach of restricting the
class of case slot models. I reduce the number of partitions necessary
for consideration by using a thesaurus, following a similar proposal
given in \cite{Resnik92}. Specifically, I restrict attention to those
partitions that exist within the thesaurus in the form of a cut. Here
by `thesaurus' is meant a rooted tree in which each leaf node stands
for a noun, while each internal node represents a noun class, and a
directed link represents set inclusion (cf., Figure~\ref{fig:thesau}).
A `cut' in a tree is any set of nodes in the tree that can represent a
partition of the given set of nouns. For example, in the thesaurus of
Figure~\ref{fig:thesau}, there are five cuts: [ANIMAL],[BIRD, INSECT],
[BIRD, bug, bee, insect], [swallow, crow, eagle, bird, INSECT], and
[swallow, crow, eagle, bird, bug, bee, insect].

The class of `tree cut models' {\em with respect to a fixed thesaurus
  tree} is then obtained by restricting the partitions in the
definition of a hard case slot model to be those that are present as a cut
in that thesaurus tree. The number of models, then, is drastically
reduced, and is of order $\Theta(2^{\frac{N}{b}})$ when the thesaurus
tree is a complete $b$-ary tree, because the number of cuts in a
complete $b$-ary tree is of that order (see
Appendix~\ref{append:cut}). Here, $N$ denotes the number of leaf
nodes, i.e., the size of the set of nouns.

A tree cut model $M$ can be represented by a pair consisting of a tree
cut $\Gamma$ (i.e., a discrete model), and a probability parameter
vector $\theta$ of the same length, that is, \[ M=(\Gamma,\theta), \]
where $\Gamma$ and $\theta$ are \[
 \Gamma = [C_1,C_2,\cdots,C_{k+1}], \theta=
 [P(C_1),P(C_2),\cdots,P(C_{k+1})], \] where $C_1,C_2,\cdots,C_{k+1}$
forms a cut in the thesaurus tree and where $\sum_{i=1}^{k+1} P(C_i) =
1$ is satisfied. Hereafter, for simplicity I sometimes write $P(C_i)$
for $P(C_i|v,r)$, where $i=1,\cdots,(k+1)$.

\begin{figure}[htb]
\begin{center}
\epsfxsize7.5cm\epsfysize4cm\epsfbox{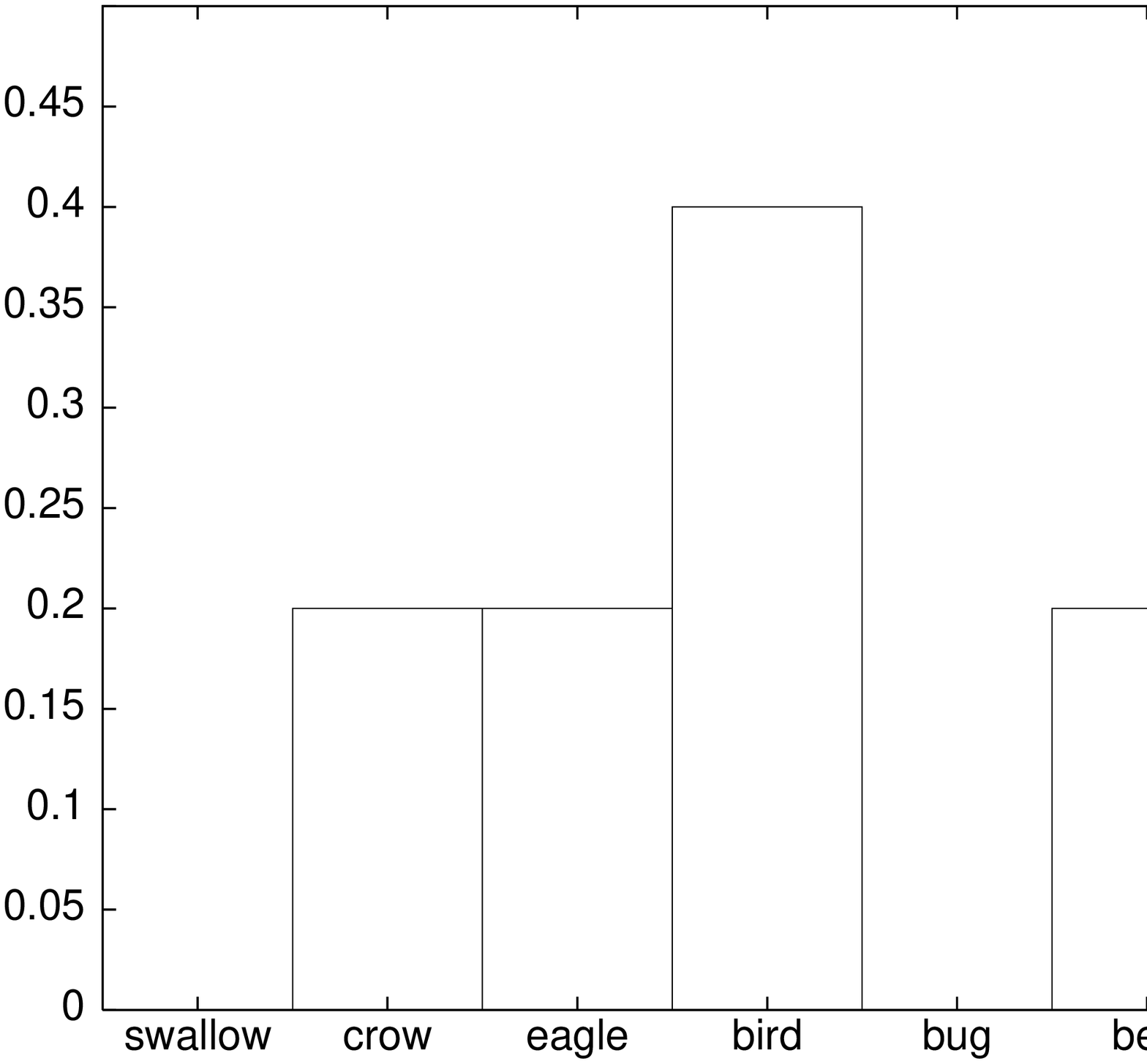}
\end{center}
\caption{A tree cut model with [swallow, crow, eagle, bird, bug, bee, insect].}
\label{fig:tcmodel1}
\end{figure}

\begin{figure}[htb]
\begin{center}
\epsfxsize7.5cm\epsfysize4cm\epsfbox{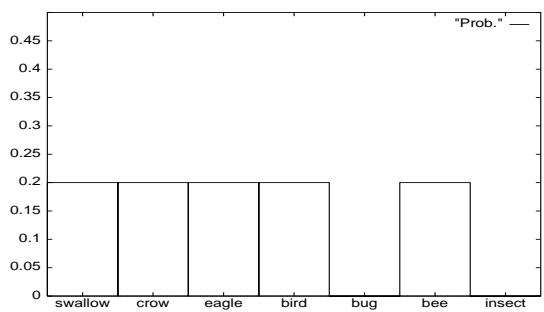}
\end{center}
\caption{A tree cut model with [BIRD, bug, bee, insect].}
\label{fig:tcmodel2}
\end{figure}

\begin{figure}[htb]
\begin{center}
\epsfxsize7.5cm\epsfysize4cm\epsfbox{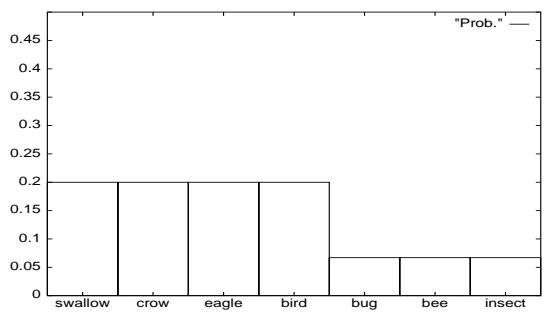}
\end{center}
\caption{A tree cut model with [BIRD, INSECT].}
\label{fig:tcmodel3}
\end{figure}

If we employ MLE for parameter estimation, we can obtain five tree cut
models from the case slot data in Figure~\ref{fig:csfreq};
Figures~\ref{fig:tcmodel1}-\ref{fig:tcmodel3} show three of these. For
example, $\hat{M} =$ $({\rm [BIRD, bug, bee, insect]}$, $[0.8, 0, 0.2,
0])$ shown in Figure~\ref{fig:tcmodel2} is one such tree cut model.
Recall that $\hat{M}$ defines a conditional probability distribution
$P_{\hat{M}}(n|v,r)$ in the following way: for any noun that is in the
tree cut, such as `bee,' the probability is given as explicitly
specified by the model, i.e., $P_{\hat{M}}({\rm bee}|{\rm fly},{\rm
  arg1}) = 0.2$; for any class in the tree cut, the probability is
distributed uniformly to all nouns included in it. For example, since
there are four nouns that fall under the class BIRD, and `swallow' is
one of them, the probability of `swallow' is thus given by
$P_{\hat{M}}({\rm swallow}|{\rm fly},{\rm arg1}) = 0.8/4 = 0.2$. Note
that the probabilities assigned to the nouns under BIRD are {\em
  smoothed}, even if the nouns have different observed frequencies.

In this way, the problem of generalizing the values of a case slot has
been formalized into that of estimating a model from the class of tree
cut models for some fixed thesaurus tree.

\section{MDL as Strategy}\label{sec:mdl}

The question now becomes what strategy (criterion) we should employ to
select the best tree cut model. I propose to adopt the MDL
principle.

\begin{table}[htb]
\caption{Number of parameters and KL divergence
for the five tree cut models.}
\label{tab:kl}
\begin{center}
\begin{tabular}{|lcc|} \hline
 {$\Gamma$} & Number of parameters & KL divergence \\ \hline
 {[ANIMAL]} & $0$ & $1.4$ \\
 {[BIRD, INSECT]} & $1$ & $0.72$ \\ 
 {[BIRD, bug, bee, insect]} & $3$ & $0.4$ \\
 {[swallow, crow, eagle, bird, INSECT]} & $4$ & $0.32$ \\
 {[swallow, crow, eagle, bird, bug, bee, insect]} & $6$ & $0$ \\ \hline
\end{tabular}
\end{center}
\end{table}

In our current problem, a model nearer the root of the thesaurus tree,
such as that of Figure~\ref{fig:tcmodel3}, generally tends to be
simpler (in terms of the number of parameters), but also tends to have
a poorer fit to the data. By way of contrast, a model nearer the
leaves of the thesaurus tree, such as that in
Figure~\ref{fig:tcmodel1}, tends to be more complex, but also tends to
have a better fit to the data. Table~\ref{tab:kl} shows the number of
free parameters and the `KL divergence' between the empirical
distribution (namely, the word-based distribution estimated by MLE) of
the data shown in Figure~\ref{fig:csmodel} and each of the five tree
cut models.\footnote{The KL divergence (also known as
  `relative entropy') is a measure of the `distance' between two
probability distributions, and is defined as $D(P||Q) = \sum_i p_i
\cdot \log \frac{p_i}{q_i}$ where $p_i$ and $q_i$ represent,
respectively, probabilities in discrete distributions $P$ and $Q$
\cite{Cover91}.}  In the table, we can see that there is a
trade-off between the simplicity of a model and the goodness of its
fit to the data. The use of MDL can balance the trade-off
relationship.

Let us consider how to calculate description length for the current
problem, where the notations are slightly different from those in
Chapter 2. Suppose that $S$ denotes a sample (or data), which is a
{\em multi-set} of examples, each of which is an {\em occurrence} of a
noun at a slot $r$ for a verb $v$ (i.e., duplication is allowed).
Further suppose that $|S|$ denotes the size of $S$, and $n \in S$
indicates the inclusion of $n$ in $S$.  For example, the column
labeled `slot value' in Table~\ref{tab:csdata2} represents a sample
$S$ for the arg1 slot for `fly,' and in this case $|S| = 10$.

Given a sample $S$ and a tree cut $\Gamma$, we can employ MLE to
estimate the parameters of the corresponding tree cut model $\hat{M} =
(\Gamma,\hat{\theta})$, where $\hat{\theta}$ denotes the estimated
parameters.

The total description length $l(\hat{M},S)$ of the tree cut model
$\hat{M}$ and the data $S$ observed through $\hat{M}$ may be computed
as the sum of model description length $l(\Gamma)$, parameter
description length $l(\hat{\theta}|\Gamma)$, and data description
length $l(S|\Gamma,\hat{\theta})$, i.e., \[
  l(\hat{M},S) = l((\Gamma,\hat{\theta}),S) = l(\Gamma) +
  l(\hat{\theta}|\Gamma) + l(S|\Gamma,\hat{\theta}).
\]

Model description length $l(\Gamma)$, here, may be calculated
as\footnote{Throughout this thesis, `$\log$' denotes the logarithm to
base 2.}  \[
  l(\Gamma) = \log |{\cal G}|, \] where ${\cal G}$ denotes
  the set of all cuts in the thesaurus tree $T$. From the viewpoint
of Bayesian Estimation, this corresponds to
  assuming that each tree cut model to be equally likely {\em a priori}.

Parameter description length $l(\hat{\theta}|\Gamma)$ 
may be calculated by
 \[
l(\hat{\theta}|\Gamma) = \frac{k}{2} \cdot \log |S|,
\]
where $|S|$ denotes the sample size and $k$ denotes the number of free
parameters in the tree cut model, i.e., $k$ equals the number of nodes
in $\Gamma$ minus one.

Finally, data description length 
$l(S|\Gamma,\hat{\theta})$ may be calculated as
\[
  l(S|\Gamma,\hat{\theta}) = - \sum_{n \in S} \log \hat{P}(n),
\] where for simplicity I write $\hat{P}(n)$ for
$P_{\hat{M}}(n|v,r)$. Recall that $\hat{P}(n)$ is obtained by 
MLE, i.e., 
\[ 
\hat{P}(n) = \frac{1}{|C|} \cdot \hat{P}(C)
\] 
for each $n \in C$, where for each $C \in \Gamma$ 
\[
\hat{P}(C) =  \frac{f(C)}{|S|},
\]
where $f(C)$ denotes the frequency of nouns in class $C$ in data
$S$.

With the description length defined in the above manner, we wish to
select a model with the minimum description length, and then output it
as the result of generalization. Since every tree cut has an equal
$l(\Gamma)$, technically we need only calculate and compare
$L'(\hat{M},S)= l(\hat{\theta}|\Gamma) + l(S|\Gamma,\hat{\theta})$. In
the discussion which follows, I sometimes use $L'(\Gamma)$ for
$L'(\hat{M},S)$, where $\Gamma$ is the tree cut of $\hat{M}$, for the
sake of simplicity.

\begin{table}[htb]
  \caption{Calculating description length.}
\label{tab:prob}
\begin{center}
\begin{tabular}{|lcccc|} \hline
$C$ & BIRD & bug & bee & insect \\ \hline
$f(C)$ & $8$ & $0$ & $2$ & $0$ \\
$|C|$ & $4$ & $1$ & $1$ & $1$ \\
$\hat{P}(C)$ & $0.8$ & $0.0$ & $0.2$ & $0.0$ \\
$\hat{P}(n)$ & $0.2$ & $0.0$ & $0.2$ & $0.0$ \\ \hline
$\Gamma$ & \multicolumn{4}{|c|}{[BIRD, bug, bee, insect]} \\ 
\hline
$l(\hat{\theta}|\Gamma)$ & \multicolumn{4}{|c|}{$\frac{(4-1)}{2}\times \log10=4.98$} \\
$l(S|\Gamma,\hat{\theta})$ & \multicolumn{4}{|c|}{$-(2+4+2+2) \times \log0.2=23.22$} \\ \hline
\end{tabular}
\end{center}
\end{table}

\begin{table}[htb]
\caption{Description lengths for the five tree cut models.}
\label{tab:length}
\begin{center}
\begin{tabular}{|lccc|} \hline
 {$\Gamma$} & $l(\hat{\theta}|\Gamma)$ & $l(S|\Gamma,\hat{\theta})$ & $L'(\Gamma)$  \\ \hline
 {[ANIMAL]} & $0$ & $28.07$ & $28.07$ \\
 {[BIRD, INSECT]} & $1.66$ & $26.39$ & \underline{$28.05$} \\ 
 {[BIRD, bug, bee, insect]} & $4.98$ & $23.22$ & $28.20$ \\
 {[swallow, crow, eagle, bird, INSECT]} & $6.64$ & $22.39$ & $29.03$ \\
 {[swallow, crow, eagle, bird, bug, bee, insect]} & $9.97$ & $19.22$ &
 $29.19$ \\ \hline
\end{tabular}
\end{center}
\end{table}

The description lengths of the data in Figure~\ref{fig:csfreq} for the
tree cut models with respect to the thesaurus tree in
Figure~\ref{fig:thesau} are shown in Table~\ref{tab:length}.
(Table~\ref{tab:prob} shows how the description length is calculated
for the model with tree cut [BIRD, bug, bee, insect].)  These figures
indicate that according to MDL, the model in Figure~\ref{fig:tcmodel3}
is the best model. Thus, given the data in Table~\ref{tab:csdata2} as
input, we are able to obtain the generalization result shown in
Table~\ref{tab:gen}.

\begin{table}[htb]
  \caption{Generalization result.}
\label{tab:gen}
\begin{center}
\begin{tabular}{|lccc|} \hline
Verb & Slot name & Slot value & Probability \\ \hline
fly & arg1 & BIRD & 0.8 \\
fly & arg1 & INSECT & 0.2 \\ \hline
\end{tabular}
\end{center}
\end{table}

Let us next consider some justifications for calculating description
lengths in the above ways.

For the model description length $l(\Gamma)$, I assumed the length to
be equal for all the {\em discrete} tree cut models. We could,
alternatively, have assigned larger code lengths to models nearer the
root node and smaller code lengths to models nearer the leaf nodes.  I
chose not to do so for the following reasons: (1) in general, when we
have no information about a class of models, it is optimal to assume,
on the basis of the `minmax strategy' in Bayesian Estimation, that
each model has equal prior probability (i.e., to assume `equal
ignorance'); (2) when the data size is large enough, the model
description length, which is only of order $O(1)$, will be negligible
compared to the parameter description length, which is of order
$O(\log |S|)$; (3) this way of calculating the model description
length is compatible with the dynamic-programming-based learning
algorithm described below.

With regard to the calculation of parameter description length
$l(\hat{\theta}|\Gamma)$, we should note that the use of the looser
form (\ref{eq:mdl}) rather than the more precise form
(\ref{eq:newmdl}) is done out of similar consideration of
compatibility with the dynamic programming technique.

\section{Algorithm}

In generalizing the values of a case slot using MDL, if computation
time were of no concern, one could in principle calculate the
description length for every possible tree cut model and output a
model with the minimum description length as a generalization result,
But since the number of cuts in a thesaurus tree is usually
exponential (cf., Appendix~\ref{append:cut}), it is impractical to do
so. Nonetheless, we were able to devise a simple and efficient
algorithm, based on dynamic programming, which is guaranteed to find a
model with the minimum description length.

The algorithm, which we call `Find-MDL,' recursively finds the optimal
submodel for each child subtree of a given (sub)tree and follows one
of two possible courses of action: (1) it either combines these
optimal submodels and returns this combination as output, or (2) it
collapses all these optimal submodels into the (sub)model containing
the root node of the given (sub)tree. Find-MDL simply chooses the
course of action which will result in the shorter description length
(cf., Figure~\ref{fig:tcalgo}). Note that for simplicity I describe
Find-MDL as outputting a tree cut, rather than a tree cut model.

\begin{figure}[htb]
\begin{tabbing}
Let $t$ denote a thesaurus (sub)tree, 
while ${\rm root}(t)$ denotes the root of $t$. \\ 
Let $c$ denote a tree cut in $t$. 
Initially $t$ is set to the entire tree. \\ 
{\bf algorithm} Find-MDL($t$):= $c$ \\ 
1. \tab {\bf if} \\ 
2. \tab \tab $t$ is a leaf node \\ 
3. \tab {\bf then} \\ 
4. \tab \tab return($[t]$); \\ 
5. \tab {\bf else} \\
6. \tab \tab For each child subtree $t_i$ of $t$ $c_i$ $:=$Find-MDL($t_i$); \\ 
7. \tab \tab $c$$:=$ append($c_i$); \\ 
8. \tab \tab {\bf if} \\ 
9. \tab \tab \tab $L'([{\rm root}(t)]) < L'(c)$ \\ 
10. \tab \tab {\bf then} \\ 
11. \tab \tab \tab return($[{\rm root}(t)]$); \\ 
12. \tab \tab {\bf else} \\ 
13. \tab \tab \tab return($c$).
\end{tabbing}
\caption{The Find-MDL algorithm.}
\label{fig:tcalgo}
\end{figure}

Note in the above algorithm that the parameter description length is
calculated as $\frac{k+1}{2} \cdot \log|S|$, where $k+1$ is the number
of nodes in the current cut, both when $t$ is the entire tree and when
it is a proper subtree.  This contrasts with the fact that the number
of {\em free} parameters is $k$ for the former, while it is $k+1$ for
the latter. For the purpose of finding a tree cut model with the
minimum description length, however, this distinction can be ignored
(cf., Appendix~\ref{append:prop1}).

\begin{figure}[htb]
\begin{center}
\epsfxsize13.5cm\epsfysize5cm\epsfbox{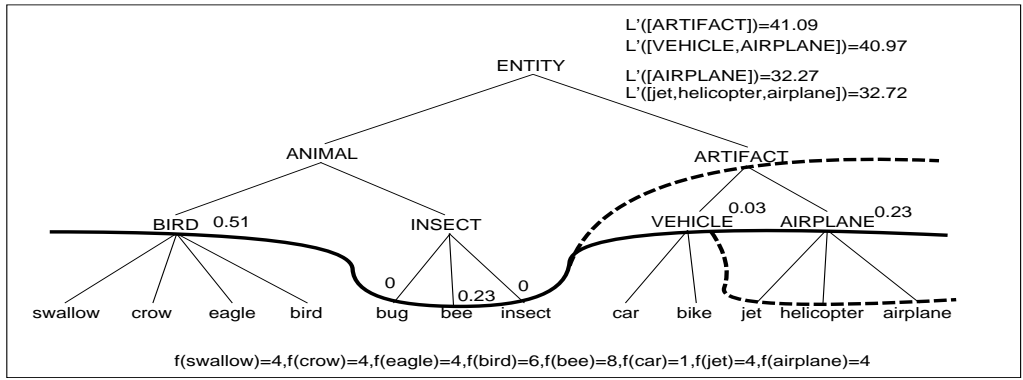}
\end{center}
\caption{An example application of Find-MDL.}
\label{fig:exfindmdl}
\end{figure}

Figure~\ref{fig:exfindmdl} illustrates how the algorithm works. In the
recursive application of Find-MDL on the subtree rooted at AIRPLANE,
the if-clause on line 9 is true since $L'([{\rm AIRPLANE}]) =32.20$,
$L'([{\rm jet, helicopter, airplane}])=32.72$, and hence $[{\rm
AIRPLANE}]$ is returned. Similarly, in the application of Find-MDL on
the subtree rooted at ARTIFACT, the same if-clause is false since
$L'([{\rm VEHICLE, AIRPLANE}])=40.83$, $L'([{\rm
  ARTIFACT}])=40.95$, and hence $[{\rm VEHICLE, AIRPLANE}]$ is
returned.

Concerning the above algorithm, the following proposition holds:
\begin{proposition}\label{prop1} The algorithm {\rm Find-MDL}
terminates in time $O(N)$, where $N$ denotes the number of leaf nodes
in the thesaurus tree $T$, and it outputs a tree cut model of $T$ with
the minimum description length (with respect to the coding scheme
described in Section~\ref{sec:mdl}).  \end{proposition} 
See Appendix~\ref{append:prop1} for a proof of the proposition.

\section{Advantages}

\subsubsection*{Coping with the data sparseness problem} Using the
MDL-based method described above, we can generalize the values of a
case slot. The probability of a noun being the value of a slot can
then be represented as a conditional probability estimated (smoothed)
from a class-based model on the basis of the MDL principle.

The advantage of this method over the word-based method described in
Chapter 2 lies in its ability to cope with the data sparseness
problem.  Formalizing this problem as a statistical estimation problem
that includes model selection enables us to select models with various
complexities, while employing MDL enables us to select, on the basis
of training data, a model with the most appropriate level of
complexity.

\subsubsection*{Generalization} The case slot generalization problem
can also be restricted to that of generalizing individual nouns
present in case slot data into classes of nouns present in a given
thesaurus. For example, given the thesaurus in Figure~\ref{fig:thesau}
and frequency data in Figure~\ref{fig:csfreq}, we would like our
system to judge that the class `BIRD' and the noun `bee' can be the
value of the arg1 slot for the verb `fly.' The problem of deciding
whether to stop generalizing at `BIRD' and `bee' or to continue
generalizing further to `ANIMAL' has been addressed by a number of
researchers (cf., \cite{Webster89,Velardi91,Nomiyama92}). The
MDL-based method described above provides a disciplined way to realize
this on the basis of data compression and statistical estimation.

The MDL-based method, in fact, conducts generalization in the
following way.  When the differences between the frequencies of the
words in a class are not large enough (relative to the entire data
size and the number of the words), it generalizes them into the class.
When the differences are especially noticeable (relative to the entire
data size and the number of the words), on the other hand, it stops
generalization at that level.

As described in Chapter 3, the class of hard case slot models contains
all of the possible models for generalization, if we view the
generalization process as that of finding the best configuration of
words such that the words in each class are equally likely to the
value of a case slot. And thus if we could estimate the best model
from the class of hard case slot models on the basis of MDL, we would
be able to obtain the most appropriate generalization result. When we
make use of a thesaurus (hand-made or automatically constructed) to
restrict the model class, the generalization result will inevitablely
be affected by the thesaurus used, and the tree cut model selected may
be a loose approximation of the best model. Because MDL achieves a
balanced trade-off between model simplicity and data fit, we may
expect that the model it selects will represent a reasonable
compromise.

\subsubsection*{Coping with extraction noise} Avoiding the influence
of noise in case slot data is another problem that needs consideration
in case slot generalization.  For example, suppose that the case slot
data on the noun `car' in Figure~\ref{fig:exfindmdl} is noise. In such
case, the MDL-based method tends to generalize a noun to a class at
quite high a level, since the differences between the frequency of the
noun and those of its neighbors are not high (e.g., $f({\rm car})=1$
and $f({\rm bike})=0$). The probabilities of the generalized classes
will, however, be small. If we discard those classes in the obtained
tree cut that have small probabilities, we will still acquire reliable
generalization results.  That is to say, the proposed method is robust
against noise.

\section{Experimental Results}

\subsection{Experiment 1: qualitative evaluation}

I have applied the MDL-based generalization method to a data corpus
and inspected the obtained tree cut models to see if they agree with
human intuition.  In the experiments, I used existing techniques
(cf., \cite{Manning92,Smadja93}) to extract case slot data from the
{\em tagged} texts of the Wall Street Journal corpus (ACL/DCI CD-ROM1)
consisting of 126,084 sentences. I then applied the method to
generalize the slot values. 

Table~\ref{tab:inputeat} shows some example case slot data for the
arg2 slot for the verb `eat.' There were some extraction errors
present in the data, but I chose not to remove them because extraction
errors are such a generally common occurrence that a realistic
evaluation should include them.

\begin{table}[htb]
  \caption{Example input data (for the arg2 slot for `eat').}
\label{tab:inputeat}
\begin{center}
\begin{tabular}{|lc|lc|lc|} \hline
eat arg2 food & 3 & eat arg2 lobster & 1 & eat arg2 seed & 1 \\
eat arg2 heart & 2 & eat arg2 liver & 1 & eat arg2 plant & 1 \\
eat arg2 sandwich & 2 & eat arg2 crab & 1 & eat arg2 elephant & 1 \\
eat arg2 meal & 2 & eat arg2 rope & 1 & eat arg2 seafood & 1 \\
eat arg2 amount & 2 & eat arg2 horse & 1 & eat arg2 mushroom & 1 \\
eat arg2 night & 2 & eat arg2 bug & 1 & eat arg2 ketchup & 1 \\
eat arg2 lunch & 2 & eat arg2 bowl & 1 & eat arg2 sawdust & 1 \\
eat arg2 snack & 2 & eat arg2 month & 1 & eat arg2 egg & 1 \\
eat arg2 jam & 2 & eat arg2 effect & 1 & eat arg2 sprout & 1 \\
eat arg2 diet & 1 & eat arg2 debt & 1 & eat arg2 nail & 1 \\
eat arg2 pizza & 1 & eat arg2 oyster & 1 & & \\ \hline
\end{tabular}
\end{center}
\end{table}

When generalizing, I used the noun taxonomy of WordNet (version1.4)
\cite{Miller95} as the thesaurus. The noun taxonomy of WordNet is
structured as a directed acyclic graph (DAG), and each of its nodes
stands for a word sense (a concept), often containing several words
having the same word sense. WordNet thus deviates from the notion of
a thesaurus as defined in Section~\ref{sec:treecut} -- a tree in which
each leaf node stands for a noun, and each internal node stands for
a class of nouns; we need to take a few measures to deal with this.

First, each subgraph having multiple parents is copied so that the
WordNet is transformed into a tree structure \footnote{In fact, there
  are only few nodes in WordNet, which have multiple parent nodes,
  i.e., the structure of WordNet approximates that of a tree.} and the
algorithm Find-MDL can be applied. Next, the issue of word sense
ambiguity is heuristically addressed by equally dividing the observed
frequency of a noun between all the nodes containing that noun. 
Finally, the highest nodes actually containing the values of the slot
are used to form the `staring cut' from which to begin generalization
and the frequencies of all the nodes below to a node in the starting
cut are added to that node. Since word senses of nouns that occur in
natural language tend to concentrate in the middle of a
taxonomy,\footnote{Cognitive scientists have observed that concepts in
  the middle of a taxonomy tend to be more important with respect to
  learning, recognition, and memory, and their linguistic expressions
  occur more frequently in natural language -- a phenomenon known as
  `basic level primacy.' (cf., \cite{Lakoff87}) } a starting cut given
by this method usually falls around the middle of the thesaurus.

\begin{figure}[htb]
\begin{center}
\epsfxsize13.5cm\epsfysize6cm\epsfbox{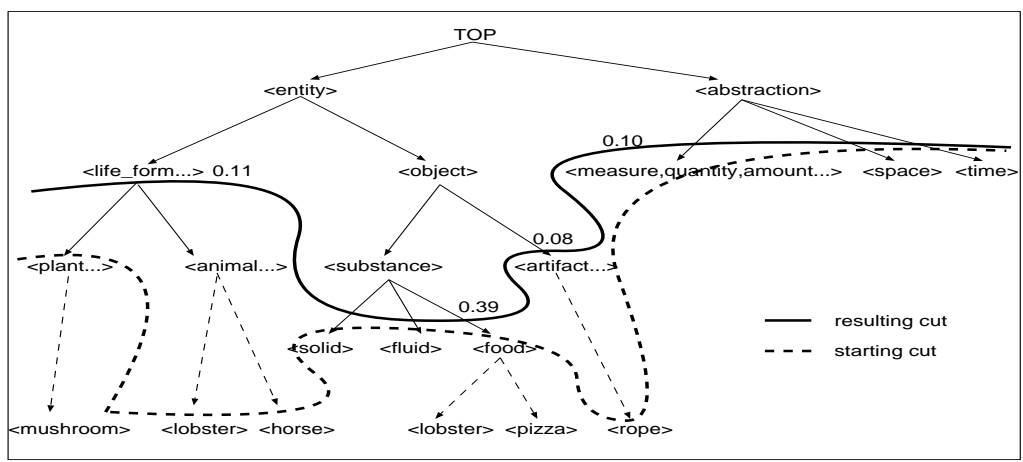}
\end{center}
\caption{Example generalization result 
(for the arg2 slot for `eat').}
\label{fig:wordnet}
\end{figure}

Figure~\ref{fig:wordnet} indicates the starting cut and the resulting
cut in WordNet for the arg2 slot for `eat' with respect to the data in
Table~\ref{tab:inputeat}, where $\langle \cdots \rangle$ denotes a
node in WordNet. The starting cut consists of those nodes $\langle
{\rm plant}, \cdots \rangle$,$\langle {\rm food} \rangle$,etc. which
are the highest nodes containing the values of the arg2 slot for
`eat.'  Since $\langle {\rm food} \rangle$ has significantly more
frequencies than its neighbors $\langle {\rm solid} \rangle$ and
$\langle {\rm fluid} \rangle$, MDL has the generalization stop
there. By way of contrast, because the nodes under $\langle {\rm
life\_form}, \cdots \rangle$ have relatively small differences in
their frequencies, they are generalized to the node $\langle {\rm
life\_form}, \cdots\rangle$. The same is true of the nodes under
$\langle {\rm artifact}, \cdots \rangle$. Since $\langle \cdots,{\rm
amount}, \cdots \rangle$ has a much higher frequency than its
neighbors $\langle {\rm time}\rangle$ and $\langle {\rm space}
\rangle$, generalization does not proceed any higher.  All of these
results seem to agree with human intuition, indicating that the method
results in an appropriate level of generalization.

Table~\ref{tab:result1} shows generalization results for the arg2 slot
for `eat' and three other arbitrarily selected verbs, where classes
are sorted in descending order with respect to probability
values. (Classes with probabilities less than $0.05$ have been
discarded due to space limitations.) Despite the fact that the
employed extraction method is not noise-free, and word sense
ambiguities remain after extraction, the generalization results seem
to agree with intuition to a satisfactory degree. (With regard to
noise, at least, this is not too surprising since the noisy portion
usually has a small probability and thus tends to be discarded.)

\begin{table}[htb]
\caption{Examples of generalization results.}
\label{tab:result1}
\begin{center}
\begin{tabular}{|lcl|} \hline
Class & Probability & Example words \\ \hline
\multicolumn{3}{|c|}{arg2 slot of `eat'} \\ \hline
$\langle$food, nutrient$\rangle$ & $0.39$ & pizza, egg \\
$\langle$life\_form, organism, $\cdots$$\rangle$ & $0.11$ & lobster, horse \\
$\langle$measure, quantity, $\cdots$$\rangle$ & $0.10$ & amount {\em of} \\
$\langle$artifact, article, $\cdots$$\rangle$ & $0.08$ & {\em as if eat} rope \\ \hline
\multicolumn{3}{|c|}{arg2 slot of `buy'} \\ \hline
$\langle$object, $\cdots$$\rangle$ & $0.30$ & computer, painting \\
$\langle$asset$\rangle$ & $0.10$ & stock, share \\
$\langle$group, grouping$\rangle$ & $0.07$ & company, bank \\
$\langle$legal\_document, $\cdots$ $\rangle$ &
$0.05$ & security, ticket \\ \hline
\multicolumn{3}{|c|}{arg2 slot of `fly'} \\ \hline
$\langle$entity$\rangle$ & $0.35$ & airplane, flag, executive\\
$\langle$linear\_measure, $\cdots$ $\rangle$ & $0.28$ & mile \\
$\langle$group, grouping$\rangle$ & $0.08$ & delegation \\ \hline
\multicolumn{3}{|c|}{arg2 slot of `operate'} \\ \hline
$\langle$group, grouping$\rangle$ & $0.13$ & company, fleet \\
$\langle$act, human\_action, $\cdots$ $\rangle$ & $0.13$ & flight, operation \\
$\langle$structure, $\cdots$ $\rangle$ & $0.12$ & center  \\
$\langle$abstraction$\rangle$ & $0.11$ & service, unit \\ 
$\langle$possession$\rangle$ & $0.06$ & profit, earnings \\ \hline
\end{tabular}
\end{center}
\end{table}

\begin{table}[htb]
  \caption{Required computation time and number of generalized levels.}  
\label{tab:eva}
\begin{center}
\begin{tabular}{|lcc|} \hline
Verb & CPU time (second) & Average number of generalized levels \\ \hline
eat & $1.00$ & $5.2$ \\
buy & $0.66$ & $4.6$ \\
fly & $1.11$ & $6.0$ \\
operate & $0.90$ & $5.0$ \\ \hline
Average & $0.92$ & $5.2$ \\ \hline
\end{tabular}
\end{center}
\end{table}

Table~\ref{tab:eva} shows the computation time required (on a SPARC
`Ultra 1' work station, not including that for loading WordNet) to
obtain the results shown in Table~\ref{tab:result1}. Even though the
noun taxonomy of WordNet is a large thesaurus containing approximately
50,000 nodes, the MDL-based method still manages to generalize case
slots efficiently with it. The table also shows the average number of
levels generalized for each slot, i.e., the average number of links
between a node in the starting cut and its ancestor node in the
resulting cut.  (For example, the number of levels generalized for
$\langle {\rm plant}, \cdots \rangle$ is one in
Figure~\ref{fig:wordnet}.)  One can see that a significant amount of
generalization is performed by the method -- the resulting tree cut is
on average about 5 levels higher than the starting cut.

\subsection{Experiment 2: pp-attachment disambiguation}

Case slot patterns obtained by the method can be used in various tasks
in natural language processing. Here, I test the effectiveness of the
use of the patterns in pp-attachment disambiguation.

In the experiments described below, I compare the performance of the
proposed method, referred to as `MDL,' against the methods proposed by
\cite{Hindle91}, \cite{Resnik93b}, and \cite{Brill94}, referred to
respectively as `LA,' `SA,' and `TEL.'

\subsubsection*{Data set} As a data set, I used the bracketed data of
the Wall Street Journal corpus (Penn Tree Bank 1)
\cite{Marcus93}. First I randomly selected one of the 26 directories
of the WSJ files as test data and what remained as training data. I
repeated this process ten times and obtained ten sets of data
consisting of different training and test data. I used these ten data
sets to conduct {\em cross validation}, as described below.

From the {\em test} data in each data set, I extracted $(v,n_1,p,n_2)$
quadruples using the extraction tool provided by the Penn Tree Bank
called `tgrep.' At the same time, I obtained the {\em answer} for the
pp-attachment for each quadruple. I did not double-check to confirm
whether or not the answers were actually correct. From the {\em
training} data of each data set, I then extracted $(v,p)$ and
$(n_1,p)$ doubles, and $(v,p,n_2)$ and $(n_1,p,n_2)$ triples using
tools I developed. I also extracted quadruples from the training data
as before. I then applied 12 heuristic rules to further preprocess the
data; this processing included (1) changing the inflected form of a
word to its stem form, (2) replacing numerals with the word `number,'
(3) replacing integers between $1900$ and $2999$ with the word `year,'
(4) replacing `co.,' `ltd.' with the word `company,' (5) etc. After
preprocessing some minor errors still remained, but I did not attempt
to remove them because of lacking a good method to do so
automatically.  Table~\ref{tab:datanum} shows the number of different
types of data obtained in the above process.

\begin{table}[htb]
\caption{Number of data items.}
\label{tab:datanum}
\begin{center}
\begin{tabular}{|lc|} \hline
\multicolumn{2}{|c|}{Training data} \\ \hline
average number of doubles per data set & $91218.1$ \\
average number of triples per data set & $91218.1$ \\ 
average number of quadruples per data set & $21656.6$ \\ \hline
\multicolumn{2}{|c|}{Test data} \\ \hline
average number of quadruples per data set & $820.4$ \\
\hline
\end{tabular}
\end{center}
\end{table}

\subsubsection*{Experimental procedure} I first compared the accuracy
and coverage for MDL, SA and LA.

For MDL, $n_2$ is generalized on the basis of two sets of triples
$(v,p,n_2)$ and $(n_1,p,n_2)$ that are given as training data for each
data set, with WordNet being used as the thesaurus in the same manner
as it was in Experiment 1. When disambiguating, rather than comparing
$\hat{P}(n_2|v,p)$ and $\hat{P}(n_2|n_1,p)$ I compare $\hat{P}(C_1|v,p)$ and
$\hat{P}(C_2|n_1,p)$, where $C_1$ and $C_2$ are classes in the output tree
cut models dominating $n_2$\footnote{Recall that a node in WordNet
  represents a word sense and not a word, $n_2$ can belong to several
  different classes in the thesaurus. In fact, I compared $\max_{C_i
    \ni n_2} (\hat{P}(C_i|v,p))$ and $\max_{C_j \ni n_2}
(\hat{P}(C_j|n_1,p))$.}; because I empirically found that to do so
gives a slightly better result. For SA, I employ a basic application
(also using WordNet) in which $n_2$ is generalized given $(v,p,n_2)$
and $(n_1,p,n_2)$ triples. For disambiguation I compare
$\hat{A}(n_2|v,p)$ and $\hat{A}(n_2|n_1,p)$ (defined in
(\ref{eq:assocratio}) in Chapter 2)). For LA, I estimate
$\hat{P}(p|v)$ and $\hat{P}(p|n_1)$ from the training data of each
data set and compare them for disambiguation. 

I then evaluated the results achieved by the three methods in terms of
accuracy and coverage. Here `coverage' refers to the percentage of
test data by which a disambiguation method can reach a decision, and
`accuracy' refers to the proportion of correct decisions among all
decisions made.

\begin{figure}[htb]
\begin{center}
\epsfxsize12cm\epsfysize7cm\epsfbox{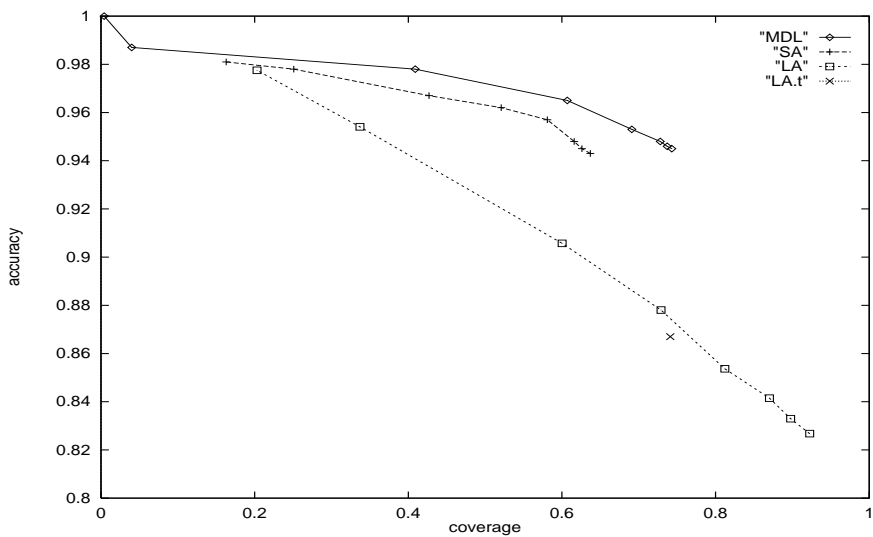}
\end{center}
\caption{Accuracy-coverage plots for MDL, SA, and LA.}
\label{fig:disamcurve}
\end{figure}

Figure~\ref{fig:disamcurve} shows the accuracy-coverage curves for the
three methods.  In plotting these curves, I first compare the
respective values for the two possible attachments. If the difference
between the two values exceeds a certain threshold, I make the
decision to attach at the higher-value site. The threshold here was
set successively to $0$,$0.01$,$0.02$,$0.05$,$0.1$,$0.2$,$0.5$,and
$0.75$ for each of the three methods. When the difference between the
two values is less than the threshold, no decision is made.  These
curves were obtained by averaging over the ten data sets.
Figure~\ref{fig:disamcurve} shows that, with respect to
accuracy-coverage curves, MDL outperforms both SA and LA throughout,
while SA is better than LA.

I also implemented the method proposed by \cite{Hindle91} which makes
disambiguation judgements using t-scores (cf., Chapter 2).
Figure~\ref{fig:disamcurve} shows the result as `LA.t,' where the
threshold for the t-score is $1.28$ (at a significance level of 90
percent.)

Next, I tested the method of applying a default rule after applying
each method. That is, attaching $(p,n_2)$ to $v$ for the part of the
test data for which no decision was made by the method in
question. (Interestingly, over the data set as a whole it is
  more favorable to attach $(p,n_2)$ to $n_1$, but for what remains
  after applying LA, SA, and MDL, it turns out to be more favorable to
  attach $(p,n_2)$ to $v$.) I refer to these combined methods as
MDL+Default, SA+Default, LA+Default, and LA.t+Default.
Table~\ref{tab:result2} shows the results, again averaged over the ten
data sets.

Finally, I used transformation-based error-driven learning (TEL) to
acquire transformation rules for each data set and applied the
obtained rules to disambiguate the test data (cf., Chapter 2).  The
average number of obtained rules for a data set was
$2752.3$. Table~\ref{tab:result2} shows disambiguation results
averaged over the ten data sets. 

From Table~\ref{tab:result2}, we see that TEL performs the best,
edging out the second place MDL+Default by a tiny margin, and followed
by LA+Default, and SA+Default. I discuss these results below.

\begin{table}[htb]
\caption{PP-attachment disambiguation results.}
\label{tab:result2}
\begin{center}
\begin{tabular}{|lcc|} \hline
Method & Coverage($\%$) & Accuracy($\%$) \\ \hline
Default & $100$ & $56.2$ \\
MDL + Default & $100$ & $82.2$ \\ 
SA + Default & $100$ & $76.7$ \\
LA + Default & $100$ & $80.7$ \\
LA.t + Default & $100$ & $78.1$ \\
TEL & $100$ & $82.4$ \\
\hline
\end{tabular}
\end{center}
\end{table}

\subsubsection*{MDL and SA}
Experimental results show that the accuracy and coverage of MDL appear
to be somewhat better than those of SA. Table~\ref{tabprotect} shows
example generalization results for MDL (with classes with probability
less than $0.05$ discarded) and SA.  Note that MDL tends to select a
tree cut model closer to the root of the thesaurus. This is probably
the key reason that MDL has a wider coverage than SA for the same
degree of accuracy.  One may be concerned that MDL may be
`over-generalizing' here, but as shown in Figure~\ref{fig:disamcurve},
this does not seem to degrade its disambiguation accuracy.

\begin{table}[htb]
\caption{Example generalization results for SA and MDL.}
\label{tabprotect}
\begin{center}
\begin{tabular}{|lccc|} \hline
\multicolumn{4}{|c|}{Input} \\ \hline
Verb & Preposition & Noun & Frequency \\ \hline
protect & against & accusation & $1$ \\
protect & against & damage & $1$ \\
protect & against & decline & $1$ \\
protect & against & drop & $1$ \\
protect & against & loss & $1$ \\
protect & against & resistance & $1$ \\
protect & against & squall & $1$ \\
protect & against & vagary & $1$ \\ \hline
\multicolumn{4}{|c|}{Generalization result of MDL} \\ \hline
Verb & Preposition & Noun class & Probability \\ \hline
protect & against & $\langle$act, human\_action, human\_activity$\rangle$ & $0.212$ \\
protect & against & $\langle$phenomenon$\rangle$ & $0.170$ \\
protect & against & $\langle$psychological\_feature$\rangle$ & $0.099$ \\
protect & against & $\langle$event$\rangle$ & $0.097$ \\
protect & against & $\langle$abstraction$\rangle$ & $0.093$ \\ \hline
\multicolumn{4}{|c|}{Generalization result of SA} \\ \hline
Verb & Preposition & Noun class & SA \\ \hline
protect & against & $\langle$caprice, impulse, vagary, whim$\rangle$ & $1.528$ \\
protect & against & $\langle$phenomenon$\rangle$ & $0.899$ \\
protect & against & $\langle$happening, occurrence, natural\_event$\rangle$ & $0.339$ \\
protect & against & $\langle$deterioration, worsening, decline, declination$\rangle$ & $0.285$ \\
protect & against & $\langle$act, human\_action, human\_activity$\rangle$ & $0.260$ \\
protect & against & $\langle$drop, bead, pearl$\rangle$ & $0.202$ \\
protect & against & $\langle$drop$\rangle$ & $0.202$ \\
protect & against & $\langle$descent, declivity, fall, decline, downslope$\rangle$ & $0.188$ \\
protect & against & $\langle$resistor, resistance$\rangle$ & $0.130$ \\
protect & against & $\langle$underground, resistance$\rangle$ & $0.130$ \\
protect & against & $\langle$immunity, resistance$\rangle$ & $0.124$ \\
protect & against & $\langle$resistance, opposition$\rangle$ & $0.111$ \\
protect & against & $\langle$loss, deprivation$\rangle$ & $0.105$ \\
protect & against & $\langle$loss$\rangle$ & $0.096$ \\
protect & against & $\langle$cost, price, terms, damage$\rangle$ & $0.052$ \\ \hline
\end{tabular}
\end{center}
\end{table}

Another problem which must be dealt with concerning SA is how to
increase the reliability of estimation. Since SA actually uses the
ratio between two probability estimates, namely
$\frac{\hat{P}(C|v,r)}{\hat{P}(C)}$, when one of the estimates is
unreliably estimated, the ratio may be lead astray. For instance, the
high estimated value shown in Table~\ref{tabprotect} for
$\langle$drop,bead,pearl$\rangle$ at `protect against' is rather odd,
and arises because the estimate of $\hat{P}(C)$ is unreliable (very
small). This problem apparently costs SA a non-negligible drop in the
disambiguation accuracy.

\subsubsection*{MDL and LA} LA makes its disambiguation decision
completely ignoring $n_2$.  As \cite{Resnik93b} pointed out, if we
hope to improve disambiguation performance with increasing training
data, we need a richer model, such as those used in MDL and SA.  I
found that $8.8\%$ of the quadruples in the entire test data were such
that they shared the same $(v,p,n_1)$ but had different $n_2$, and
their pp-attachment sites went both ways in the same data, i.e., both
to $v$ and to $n_1$. Clearly, for these examples, the pp-attachment
site cannot be reliably determined without knowing
$n_2$. Table~\ref{tabhard} shows some of these examples.  (I have
adopted the attachment sites given in the Penn Tree Bank, without
correcting apparently wrong judgements.)

\begin{table}[htb]
  \caption{Some hard examples for LA.}
\label{tabhard}
\begin{center}
\begin{tabular}{|ll|} \hline
Attached to $v$ & Attached to $n_1$ \\ \hline
acquire interest in year & acquire interest in firm \\
buy stock in trade & buy stock in index \\
ease restriction on export & ease restriction on type \\
forecast sale for year & forecast sale for venture \\
make payment on million & make payment on debt \\
meet standard for resistance & meet standard for car \\
reach agreement in august & reach agreement in principle \\
show interest in session & show interest in stock \\
win verdict in winter & win verdict in case \\ \hline
\end{tabular}
\end{center}
\end{table}

\subsubsection*{MDL and TEL} TEL seems to perform slightly better than
MDL. We can, however, develop a more sophisticated MDL method which
outperforms TEL, as may be seen in Chapter 7.

\section{Summary}

I have proposed a method for generalizing case slots. The method has
the following merits: (1) it is theoretically sound; (2) it is
computationally efficient; (3) it is robust against noise. One of the
disadvantages of the method is that its performance depends on the
structure of the particular thesaurus used.  This, however, is a
problem commonly shared by any generalization method which uses a
thesaurus as prior knowledge.

The approach of applying MDL to estimate a tree cut model in an
existing thesaurus is not limited to just the problem of generalizing
values of a case slot. It is potentially useful in other natural
language processing tasks, such as estimating n-gram models (cf.,
\cite{Brown92,Stolcke94b,Pereira95,Rosenfeld96,Ristad95,Saul97}) or semantic tagging (cf.,
\cite{Cucchiarelli97}).

\chapter{Case Dependency Learning}

\begin{tabular}{p{5.5cm}r}
 &
\begin{minipage}{10cm}
\begin{tabular}{p{9cm}}
  {\em The concept of the mutual independence of events 
    is the most essential sprout in the development of probability theory.}\\
  \multicolumn{1}{r}{- Andrei Kolmogorov} \\ 
\end{tabular}
\end{minipage}
\end{tabular}
\vspace{0.5cm}

In this chapter, I describe one method for learning the case frame
model, i.e., learning dependencies between case frame slots.

\section{Dependency Forest Model}

As described in Chapter 3, we can view the problem of learning
dependencies between case slots for a given verb as that of learning a
multi-dimensional discrete joint probability distribution referred to
as a `case frame model.' The number of parameters in a joint
distribution will be exponential, however, if we allow
interdependencies among all of the variables (even the slot-based case
frame model has $O(2^{n})$ parameters, where $n$ is the number of
random variables ), and thus their accurate estimation may not be
feasible in practice.  It is often assumed implicitly in natural
language processing that case slots (random variables) are {\em
mutually independent}.

Although assuming that random variables are mutually independent would
drastically reduce the number of parameters (e.g., under the
independence assumption, the number of parameters in a slot-based
model becomes $O(n)$). As illustrated in (\ref{eq:fly}) in Chapter 1,
this assumption is not necessarily valid in practice.

What seems to be true in practice is that some case slots are in fact
dependent on one another, but that the overwhelming majority of them
are mutually independent, due partly to the fact that usually only a
few case slots are obligatory; the others are optional. (Optional case
slots are not necessarily independent, but if two optional case slots
are randomly selected, it is very likely that they are independent of
one another.) Thus the target joint distribution is likely to be
approximatable as the product of lower order component distributions,
and thus has in fact a reasonably small number of parameters. We are
thus lead to the approach of approximating the target joint
distribution by a simplified distribution based on corpus data.

In general, any n-dimensional discrete joint distribution can be
written as
\[
P(X_1,X_2,\cdots,X_n) = \prod_{i=1}^{n} P(X_{m_i}\vert X_{m_1},\cdots, X_{m_{i-1}})
\]
for a permutation ($m_1,m_2,\cdots,m_n$) of $(1,2,\cdots,n)$, letting
$P(X_{m_1}\vert X_{m_0})$ denote $P(X_{m_1})$.

A plausible assumption regarding the dependencies between random
variables is that each variable {\em directly} depends on at most one
other variable. This is one of the simplest assumptions that can be
made to relax the independence assumption. For example, if the joint
distribution $P(X_1,X_2,X_3)$ over 3 random variables $X_1,X_2,X_3$
can be written (approximated) as follows, it (approximately) satisfies
such an assumption:
\begin{equation}\label{eq:exdend}
  P(X_1,X_2,X_3) =(\approx) P(X_1)\cdot P(X_2\vert X_1)\cdot
  P(X_3\vert X_2).
\end{equation}
I call such a distribution a `dependency forest model.'

A dependency forest model can be represented by a dependency forest
(i.e., a set of dependency trees), whose nodes represent random
variables (each labeled with a number of parameters), and whose
directed links represent the dependencies that exist between these
random variables.  A dependency forest model is thus a restricted form
of a Bayesian network \cite{Pearl88}. Graph (5) in
Figure~\ref{fig:graph} represents the dependency forest model defined
in (\ref{eq:exdend}). Table~\ref{tab:exdend} shows the parameters
associated with each node in the graph, assuming that the dependency
forest model is slot-based. When a distribution can be represented by
a single dependency tree, I call it a `dependency tree model.'

\begin{table}[htb]
\caption{Parameters labeled with each node.}
\label{tab:exdend}
\begin{center}
\begin{tabular}{|ll|} \hline 
 Node & Parameters \\ \hline
 $X_1$ & $P(X_1=1)$, $P(X_1=0)$ \\
 $X_2$ & $P(X_2=1\vert X_1=1)$, $P(X_2=0\vert X_1=1)$, $P(X_2=1\vert
 X_1=0)$, $P(X_2=0\vert X_1=0)$ \\
 $X_3$ & $P(X_3=1\vert X_2=1)$, $P(X_3=0\vert X_2=1)$, $P(X_3=1\vert
 X_2=0)$, $P(X_3=0\vert X_2=0)$ \\ \hline
\end{tabular}
\end{center}
\end{table}

It is not difficult to see that disregarding the actual values of the
probability parameters, we will have 16 and only 16 dependency forest
models (i.e., 16 dependency forests) as approximations of the joint
distribution $P(X_1,X_2,X_3)$, Since some of them are equivalent with
each other, they can be further reduced into 7 equivalent classes of
dependency forest models. Figure~\ref{fig:graph} shows the 7
equivalent classes and their members. (It is easy to verify that the
dependency tree models based on a `labeled free tree' are equivalent
to one another (cf., Appendix~\ref{append:dend}). Here a `labeled free
tree' refers to a tree in which each node is uniquely associated with
a label and in which any node can be the root \cite{Knuth73}.)

\section{Algorithm}

Now we turn to the problem of how to select the best dependency forest
model from among all possible ones to approximate a target joint
distribution based on the data. This problem has already been
investigated in the area of machine learning and related fields. One
classical method is Chow \& Liu's algorithm for estimating a
multi-dimensional discrete joint distribution as a dependency tree
model, in a way which is both efficient and theoretically sound
\cite{Chow68}.\footnote{In general, learning a Baysian network is an
intractable
  task \cite{Cooper92}.} More recently, Suzuki has extended their
algorithm, on the basis of the MDL principle, so that it estimates the
target joint distribution as a dependency forest model
\cite{Suzuki93}, and Suzuki's is the algorithm I employ here.

\vspace{0.5cm}
\setlength{\unitlength}{1.1mm}
\begin{figure}[htb]
\begin{center}
{\small
\begin{picture}(150,140)(0,0)
\put(20,45){$X_1$}
\put(10,25){$X_2$}
\put(30,25){$X_3$}
\put(5,20){$P(X_1,X_2,X_3)$}
\put(1,15){$=P(X_1)P(X_2\vert X_1)P(X_3\vert X_2)$}
\put(1,10){$=P(X_2)P(X_1\vert X_2)P(X_3\vert X_2)$}
\put(1,5){$=P(X_3)P(X_2\vert X_3)P(X_1\vert X_2)$}
\put(15,26){\vector(1,0){15}}
\put(22,44){\vector(-1,-2){8}}
\put(15,0){(5)}

\put(70,45){$X_1$}
\put(60,25){$X_2$}
\put(80,25){$X_3$}
\put(55,20){$P(X_1,X_2,X_3)$}
\put(51,10){$=P(X_1)P(X_3\vert X_1)P(X_2\vert X_1)$}
\put(51,15){$=P(X_2)P(X_1\vert X_2)P(X_3\vert X_1)$}
\put(51,5){$=P(X_3)P(X_1\vert X_3)P(X_2\vert X_1)$}
\put(72,44){\vector(1,-2){8}}
\put(64,28){\vector(1,2){8}}
\put(65,0){(6)}

\put(120,45){$X_1$}
\put(110,25){$X_2$}
\put(130,25){$X_3$}
\put(105,20){$P(X_1,X_2,X_3)$}
\put(101,15){$=P(X_1)P(X_3\vert X_1)P(X_2\vert X_3)$}
\put(101,10){$=P(X_3)P(X_1\vert X_3)P(X_2\vert X_3)$}
\put(101,5){$=P(X_2)P(X_3\vert X_2)P(X_1\vert X_3)$}
\put(130,26){\vector(-1,0){15}}
\put(122,44){\vector(1,-2){8}}
\put(115,0){(7)}

\put(20,100){$X_1$}
\put(10,80){$X_2$}
\put(30,80){$X_3$}
\put(5,75){$P(X_1,X_2,X_3)$}
\put(1,70){$=P(X_1)P(X_2)P(X_3\vert X_2)$}
\put(1,65){$=P(X_1)P(X_3)P(X_2\vert X_3)$}
\put(15,81){\vector(1,0){15}}
\put(15,60){(2)}

\put(70,100){$X_1$}
\put(60,80){$X_2$}
\put(80,80){$X_3$}
\put(55,75){$P(X_1,X_2,X_3)$}
\put(51,70){$=P(X_1)P(X_2\vert X_1)P(X_3)$}
\put(51,65){$=P(X_2)P(X_1\vert X_2)P(X_3)$}
\put(72,99){\vector(-1,-2){8}}
\put(65,60){(3)}

\put(120,100){$X_1$}
\put(110,80){$X_2$}
\put(130,80){$X_3$}
\put(105,75){$P(X_1,X_2,X_3)$}
\put(101,70){$=P(X_1)P(X_3\vert X_1)P(X_2)$}
\put(101,65){$=P(X_3)P(X_1\vert X_3)P(X_2)$}
\put(122,99){\vector(1,-2){8}}
\put(115,60){(4)}

\put(20,145){$X_1$}
\put(10,125){$X_2$}
\put(30,125){$X_3$}
\put(5,120){$P(X_1,X_2,X_3)$}
\put(1,115){$=P(X_1)P(X_2)P(X_3)$}
\put(15,110){(1)}
\end{picture}
}
\end{center}
\caption{Example dependency forests.}
\label{fig:graph}
\end{figure}
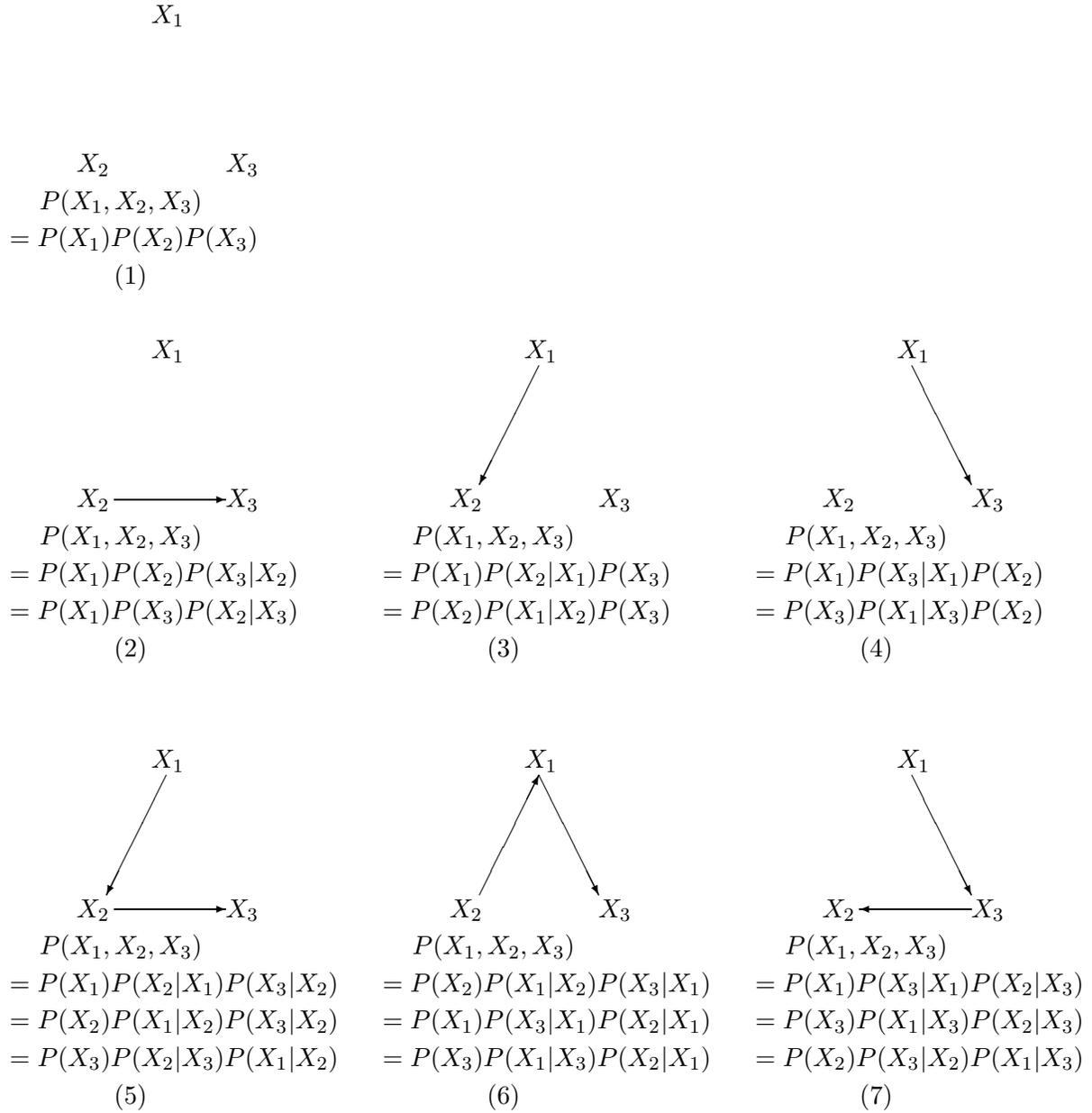

Suzuki's algorithm first calculates the statistic $\theta$ between
all node pairs. The statistic $\theta(X_i,X_j)$ between node $X_i$ and
$X_j$ is defined as
\[
\theta(X_i,X_j) = \hat{I}(X_i,X_j) - \frac{(k_i-1)\cdot (k_j-1)}{2}\cdot \log N,
\]
where $\hat{I}(X_i,X_j)$ denotes the empirical mutual information
between random variables $X_i$ and $X_j$; $k_i$ and $k_j$ denote,
respectively, the number of possible values assumed by $X_i$ and
$X_j$; and $N$ the input data size. The empirical mutual information
between random variables $X_i$ and $X_j$ is defined as
\[ 
\begin{array}{l}
\hat{I}(X_i,X_j)=\hat{H}(X_i)-\hat{H}(X_i|X_j) \\
\hat{H}(X_i) = - \sum_{x_i \in X_i} \hat{P}(x_i) \cdot \log \hat{P}(x_i) \\
\hat{H}(X_i|X_j) = - \sum_{x_i \in X_i}\sum_{x_j \in X_j} \hat{P}(x_i,x_j) \cdot \log \hat{P}(x_i|x_j), \\
\end{array}
\]
where $\hat{P}(.)$ denotes the maximum likelihood estimate of
probability $P(.)$. Furthermore, $0 \times \log 0 = 0$ is assumed to
be satisfied.

\begin{figure}[htb]
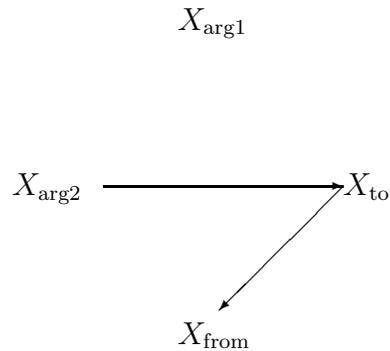

\begin{tabbing}
{\bf Algorithm:} \\
1. Let $T:= \emptyset$; \\
2. Let $V$ = $\left\{ \{ X_i\}, i=1,2,\cdots,n\right \}$; \\
3. Calculate $\theta(X_i,X_j)$ for all node pairs $(X_i,X_j)$
; \\
4. Sort the node pairs in descending order of $\theta$, and store them into queue $Q
$; \\
5. {\bf while} \\
6. \tab $\max_{(X_i,X_j)\in Q} \theta(X_i,X_j) > 0$ \\
7. {\bf do} \\
8. \tab Remove $\arg\max_{(X_i,X_j)\in Q}\theta(X_i,X_j)$ from $Q$; \\
9. \tab {\bf if} \\
10. \tab \tab $X_i$ and $X_j$ belong to different sets $W_1$,$W_2$ in $V$ \\
11. \tab {\bf then} \\
12. \tab \tab Replace $W_1$ and $W_2$ in $V$ with $W_1 \cup W_2$, and add
edge $(X_i,X_j)$ to $T$; \\
13. Output $T$ as the set of edges of the dependency forest. \\
\end{tabbing}
\caption{The learning algorithm.}
\label{fig:algorithm}
\end{figure}

The algorithm then sorts the node pairs in descending order with
respect to $\theta$. It then puts a link between the node pair with
the largest $\theta$ value, provided that this value is larger than
zero. It repeats this process until no node pair is left unprocessed,
provided that adding that link will not create a loop in the current
dependency graph. Figure~\ref{fig:algorithm} shows the algorithm.
Note that the dependency forest that is output by the algorithm may
not be uniquely determined.

Concerning the above algorithm, the following proposition holds:
\begin{proposition}\label{prop2} The algorithm 
outputs a dependency forest model with
the minimum description length.  
\end{proposition} 
See Appendix~\ref{append:prop2} for a proof of the proposition.

It is easy to see that the number of parameters in a dependency forest
model is of the order $O(n \cdot k^2)$, where $k$ is the maximum of
all $k_i$, and $n$ is the number of random variables. If we employ the
`quick sort algorithm' to perform line 4, average case time complexity
of the algorithm will be only of the order $O(n^2 \cdot (k^2+\log
n))$, and worst case time complexity will be only of the order
$O(n^2 \cdot (k^2+n^2))$.

Let us now consider an example of how the algorithm works. Suppose
that the input data is as given in Table~\ref{tab:cfdata2} and there
are 4 nodes (random variables) $X_{\rm arg1}$, $X_{\rm arg2}$, $X_{\rm
  from}$, and $X_{\rm to}$. Table~\ref{tab:mutual} shows the
statistic $\theta$ for all node pairs. The dependency forest shown in
Figure~\ref{fig:example} has been constructed on the basis of the
values given in Table~\ref{tab:mutual}. The dependency forest
indicates that there is dependency between the `to' slot and the arg2
slot, and between the `to' slot and the `from' slot.

\begin{table}[htb]
\caption{The statistic $\theta$ for node pairs.}
\label{tab:mutual}
\begin{center}
\begin{tabular}{|lcccc|} \hline
$\theta$ & $X_{\rm arg1}$ & $X_{\rm arg2}$ & $X_{\rm from}$ & $X_{\rm to}$ \\ \hline
$X_{\rm arg1}$ & & $-0.28$ & $-0.16$ & $-0.18$ \\
$X_{\rm arg2}$ & & $$ & \underline{$0.11$} & \underline{$0.57$} \\
$X_{\rm from}$ & & $$ & $$ & \underline{$0.28$} \\
$X_{\rm to}$ & & $$ & $$ & $$ \\ \hline
\end{tabular}
\end{center}
\end{table}

\begin{figure}[htb]
\begin{center}
\begin{picture}(80,60)(0,0)
\put(30,60){$X_{\rm arg1}$}
\put(10,40){$X_{\rm arg2}$}
\put(30,22){$X_{\rm from}$}
\put(50,40){$X_{\rm to}$}
\put(21,41){\vector(1,0){29}}
\put(50,41){\vector(-1,-1){15}}

\put(5,10){$P(X_{\rm arg1},X_{\rm arg2},X_{from},X_{\rm to})$}
\put(1,5){$=P(X_{\rm arg1})P(X_{\rm arg2})P(X_{\rm to}\vert X_{\rm arg2})P(X_{\rm from}\vert
  X_{\rm to})$}
\end{picture}
\end{center}
\caption{A dependency forest as case frame patterns.}
\label{fig:example}
\end{figure}

As previously noted, the algorithm is based on the MDL principle. In
the current problem, a simple model means a model with fewer
dependencies, and thus MDL provides a theoretically sound way to learn
only those dependencies that are statistically significant in the
given data. As mentioned in Chapter 2, an especially interesting
feature of MDL is that it incorporates the input data size in its
model selection criterion.  This is reflected, in this case, in the
derivation of the threshold $\theta$. Note that when we do not have
enough data (i.e., $N$ is too small), the thresholds will be large and
few nodes will be linked, resulting in a simpler model in which most
of the random variables are judged to be mutually independent. This is
reasonable since with a small data size most random variables cannot
be determined to be dependent with any significance.

Since the number of dependency forest models for a fixed number of
random variables $n$ is of order $O(2^{n-1}\cdot n^{n-2})$ (the number
of dependency tree models is of order $\Theta(n^{n-2})$
\cite{Knuth73}), it would be impossible to calculate description
length straightforwardly for all of them. Suzuki's algorithm
effectively utilizes the {\em tree
structures} of the models and efficiently calculates description
lengths by doing it locally (as does Chow \& Liu's algorithm).

\section{Experimental Results}

I have experimentally tested the performance of the proposed method of
learning dependencies between case slots. Most specifically, I have
tested to see how effective the dependencies acquired by the proposed
method are when used in disambiguation experiments. In this section, I
describe the procedures and the results of those experiments.

\subsection{Experiment 1: slot-based model}

In the first experiment, I tried to learn slot-based dependencies. As
training data, I used the entire bracketed data of the Wall Street
Journal corpus (Penn Tree Bank). I extracted case frame data from the
corpus using heuristic rules. There were $354$ verbs for which more
than $50$ case frame instances were extracted from the
corpus. Table~\ref{tab:verb} shows the most frequent verbs and the
corresponding numbers of case frames.  In the experiment, I only
considered the 12 most frequently occurring case slots (shown in
Table~\ref{tab:slotname}) and ignored others.

\begin{table}[htb]
\caption{Verbs appearing most frequently.}
\label{tab:verb}
\begin{center}
\begin{tabular}{|lc|} \hline
Verb & Number of case frames \\ \hline
be & 17713 \\
say & 9840 \\
have & 4030 \\
make & 1770 \\
take & 1245 \\
expect & 1201 \\
sell & 1147 \\
rise & 1125 \\
get & 1070 \\
go & 1042 \\
do & 982 \\
buy & 965 \\
fall & 862 \\
add & 740 \\
come & 733 \\
include & 707 \\
give & 703 \\
pay & 700 \\
see & 680 \\
report & 674 \\ \hline
\end{tabular}
\end{center}
\end{table}

\begin{table}[htb]
\caption{Case slots considered.}
\label{tab:slotname}
\begin{center}
\begin{tabular}{|llllll|} \hline
arg1 & arg2 & on & in & for & at \\
by & from & to & as & with &
against \\ \hline
\end{tabular}
\end{center}
\end{table}

\subsubsection*{Example case frame patterns}
I acquired slot-based case frame patterns for the 354 verbs. There
were on average $484/354=1.4$ dependency links acquired for each of
these 354 verbs. As an example, Figure~\ref{fig:buy} shows the case
frame patterns (dependency forest model) obtained for the verb `buy.'
There are four dependencies in this model; one indicates
that, for example, the arg2 slot is dependent on the arg1 slot.

\begin{figure}[htb]
{\small
\begin{verbatim}
buy:
[arg1]: [P(arg1=0)=0.004 P(arg1=1)=0.996]
[arg2]: [P(arg2=0|arg1=0)=0.100,P(arg2=1|arg1=0)=0.900,
         P(arg2=0|arg1=1)=0.136,P(arg2=1|arg1=1)=0.864]
[for]: [P(for=0|arg1=0)=0.300,P(for=1|arg1=0)=0.700,
        P(for=0|arg1=1)=0.885,P(for=1|arg1=1)=0.115]
[at]: [P(at=0|for=0)=0.911,P(at=1|for=0)=0.089,
         P(at=0|for=1)=0.979,P(at=1|for=1)=0.021]
[in]: [P(in=0|at=0)=0.927,P(in=0|at=0)=0.073,
         P(in=0|at=1)=0.994,P(in=1|at=1)=0.006]
[on]: [P(on=0)=0.975,P(on=1)=0.025]
[from]: [P(from=0)=0.937,P(from=1)=0.063]
[to]: [P(to=0)=0.997,P(on=1)=0.003]
[by]: [P(by=0)=0.995,P(by=1)=0.005]
[with]: [P(with=0)=0.993,P(with=1)=0.007]
[as]: [P(as=0)=0.991,P(as=1)=0.009]
[against]: [P(against=0)=0.999,P(against=1)=0.001]
\end{verbatim}
}
\caption{Case frame patterns (dependency forest model) for `buy.'}
\label{fig:buy}
\end{figure}

I found that there were some verbs whose arg2 slot is dependent on a
preposition (hereafter, $p$ for short) slot.  Table~\ref{tab:depen1}
shows the $40$ verbs having the largest values of $P(X_{\rm
arg2}=1,X_{p}=1)$, sorted in descending order of these values. The
dependencies found by the method seem to agree with human intuition.

Furthermore, I found that there were some verbs having preposition
slots that depend on each other (I refer to these as $p1$ and $p2$ for
short).  Table~\ref{tab:depen2} shows the $40$ verbs having the
largest values of $P(X_{p1}=1,X_{p2}=1)$, sorted in descending order.
Again, the dependencies found by the method seem to agree with human
intuition.

\begin{table}
\caption{Verbs and their dependent case slots.}
\label{tab:depen1}
\begin{center}
\begin{tabular}{|lll|}\hline
Verb & Dependent slots & Example \\ \hline
base & arg2 on & base pay on education \\
advance & arg2 to & advance 4 to 40 \\
gain & arg2 to & gain 10 to 100 \\
compare & arg2 with & compare profit with estimate \\
invest & arg2 in & invest share in fund \\
acquire & arg2 for & acquire share for billion \\
estimate & arg2 at & estimate price at million \\
convert & arg2 to & convert share to cash \\
add & arg2 to & add 1 to 3\\
engage & arg2 in & enage group in talk \\
file & arg2 against & file suit against company \\
aim & arg2 at & aim it at transaction \\
sell & arg2 to & sell facility to firm \\
lose & arg2 to & lose million to 10\% \\
pay & arg2 for & pay million for service \\
leave & arg2 with & leave himself with share \\
charge & arg2 with & charge them with fraud \\
provide & arg2 for & provide engine for plane \\
withdraw & arg2 from & withdraw application from office \\
prepare & arg2 for & prepare case for trial \\
succeed & arg2 as & succeed Taylor as chairman \\
discover & arg2 in & discover mile in ocean \\
move & arg2 to & move employee to New York \\
concentrate & arg2 on & concentrate business on steel \\
negotiate & arg2 with & negotiate rate with advertiser \\
open & arg2 to & open market to investor \\
protect & arg2 against & protect investor against loss \\
keep & arg2 on & keep eye on indicator \\
describe & arg2 in & describe item in inch \\
see & arg2 as & see shopping as symptom \\
boost & arg2 by & boost value by 2\% \\
pay & arg2 to & pay commission to agent \\
contribute & arg2 to & contribute million to leader \\
bid & arg2 for & bid million for right \\
threaten & arg2 against & threaten sanction against lawyer \\
file & arg2 for & file lawsuit for dismissal \\
know & arg2 as & know him as father  \\
sell & arg2 at & sell stock at time \\
settle & arg2 at & settle session at 99 \\
see & arg2 in & see growth in quarter \\ \hline
\end{tabular}
\end{center}
\end{table}

\begin{table}[htb]
\caption{Verbs and their dependent case slots.}
\label{tab:depen2}
\begin{center}
\begin{tabular}{|lll|} \hline
Head & Dependent slots & Example \\ \hline
range & from to & range from 100 to 200 \\
climb & from to & climb from million to million \\
rise & from to & rise from billion to billion \\
shift & from to & shift from stock to bond \\
soar & from to & soar from 10\% to 20\% \\
plunge & from to & plunge from 20\% to 2\%  \\
fall & from to & fall from million to million \\ 
surge & from to & surge from 100 to 200 \\
increase & from to & increase from million to million \\
jump & from to & jump from yen to yen \\
yield & from to & yield from 1\% to 5\% \\
climb & from in & climb from million in period \\
apply & to for & apply to commission for permission \\
grow & from to & grow from million to million \\
draw & from in & draw from thrift in bonus \\
boost & from to & boost from 1\% to 2\% \\
convert & from to & convert from form to form \\
raise & from to & raise from 5\% to 10\% \\
retire & on as & retire on 2 as officer \\
move & from to & move from New York to Atlanta \\
cut & from to & cut from 700 to 200 \\
sell & to for & sell to bakery for amount \\
open & for at & open for trading at yen \\
lower & from to & lower from 10\% to 2\% \\
rise & to in & rise to 5\% in month \\
trade & for in & trade for use in amount \\
supply & with by & supply with meter by 1990 \\
elect & to in & elect to congress in 1978  \\
point & to as & point to contract as example \\
drive & to in & drive to clinic in car \\
vote & on at & vote on proposal at meeting \\
acquire & from for & acquire from corp. for million \\
end & at on & end at 95 on Friday \\
apply & to in & apply to congress in 1980 \\
gain & to on & gain to 3 on share \\
die & on at & die on Sunday at age \\
bid & on with & bid on project with Mitsubishi \\
file & with in & file with ministry in week \\
slow & from to & slow from pound to pound \\
improve & from to & improve from 10\% to 50\% \\ \hline
\end{tabular}
\end{center}
\end{table}

\subsubsection*{Perplexity reduction} I also evaluated the acquired
case frame patterns (slot-based models) for all of the $354$ verbs in
terms of reduction of the `test data perplexity.'\footnote{The test
data perplexity is a measure of testing how well an estimated
probability model predicts future data, and is defined as
$2^{H(P_T,P_M)}, H(P_T,P_M) = - \sum_{x} P_T(x) \cdot \log P_M(x)$,
where $P_M(x)$ denotes the estimated model, $P_T(x)$ the empirical
distribution of the test data (cf., \cite{Bahl83}). It is roughly the
case that the smaller perplexity a model has, the closer to the true
model it is.}

I conducted the evaluation through a ten-fold cross validation. That
is, to acquire case frame patterns for the verb, I used nine tenths of
the case frames for each verb as training data, saving what remained
for use as test data, and then calculated the test data perplexity. I
repeated this process ten times and calculated average perplexity.  I
also calculated average perplexity for `independent models' which were
acquired based on the assumption that each case slot is independent.

\begin{table}
\caption{Verbs with significant perplexity reduction.}
\label{tab:per1}
\begin{center}
\begin{tabular}{|lcc|} \hline
Verb & Independent & Dependency forest (reduction in percentage) \\ \hline
base & $5.6$ & $3.6(36\%)$ \\
lead & $7.3$ & $4.9(33\%)$ \\
file & $16.4$ & $11.7(29\%)$ \\
result & $3.9$ & $2.8(29\%)$ \\
stem & $4.1$ & $3.0(28\%)$ \\
range & $5.1$ & $3.7(28\%)$ \\
yield & $5.4$ & $3.9(27\%)$ \\
benefit & $5.6$ & $4.2(26\%)$ \\
rate & $3.5$ & $2.6(26\%)$ \\
negotiate & $7.2$ & $5.6(23\%)$ \\ \hline
\end{tabular}
\end{center}
\end{table}

\begin{table}
\caption{Randomly selected verbs and their perplexities.}
\label{tab:per2}
\begin{center}
\begin{tabular}{|lcc|} \hline
Verb & Independent & Dependency forest (reduction in percentage) \\ \hline
add & $4.2$ & $3.7(9\%)$ \\
buy & $1.3$ & $1.3(0\%)$ \\
find & $3.2$ & $3.2(0\%)$ \\
open & $13.7$ & $12.3(10\%)$ \\
protect & $4.5$ & $4.7(-4\%)$ \\
provide & $4.5$ & $4.3(4\%)$ \\
represent & $1.5$ & $1.5(0\%)$ \\
send & $3.8$ & $3.9(-2\%)$ \\
succeed & $3.7$ & $3.6(4\%)$ \\
tell & $1.7$ & $1.7(0\%)$ \\ \hline
\end{tabular}
\end{center}
\end{table}

Experimental results indicate that for some verbs the use of the
dependency forest model results in less perplexity than does use of
the independent model. For $30$ of the $354$ ($8\%$) verbs, perplexity
reduction exceeded $10\%$, while average perplexity reduction overall
was $1\%$. Table~\ref{tab:per1} shows the $10$ verbs having the
largest perplexity reductions. Table~\ref{tab:per2} shows perplexity
reductions for $10$ randomly selected verbs. There were a small number
of verbs showing perplexity increases with the worst case being $5\%$. 
It seems safe to say that the dependency forest model is more suitable
for representing the `true' model of case frames than the independent
model, at least for $8\%$ of the $354$ verbs.

\subsection{Experiment 2: slot-based disambiguation}

To evaluate the effectiveness of the use of dependency knowledge in
natural language processing, I conducted a pp-attachment
disambiguation experiment. Such disambiguation would be, for example,
to determine which word, `fly' or `jet,' the phrase `from Tokyo'
should be attached to in the sentence ``She will fly a jet from Tokyo." 
A straightforward way of disambiguation would be to compare the
following likelihood values, based on slot-based models,
\[ P_{\rm fly}(X_{\rm arg2}=1,X_{\rm from}=1)\cdot P_{{\rm
jet}}(X_{\rm from}=0) \] and \[ P_{\rm fly}(X_{\rm arg2}=1,X_{\rm
from}=0)\cdot P_{{\rm jet}}(X_{\rm from}=1), \] assuming that there are
only two case slots: arg2 and `from' for the verb `fly,' and there is
one case slot: `from' for the noun `jet.'  In fact, we need only
compare \[ P_{\rm fly}(X_{\rm from}=1|X_{\rm arg2}=1)\cdot (1-P_{{\rm
jet}}(X_{\rm from}=1)) \] and \[ (1-P_{\rm fly}(X_{\rm from}=1|X_{\rm
arg2}=1))\cdot P_{{\rm jet}}(X_{\rm from}=1), \] or equivalently, \[
P_{\rm fly}(X_{\rm from}=1|X_{\rm arg2}=1) \] and \[ P_{{\rm
jet}}(X_{\rm from}=1).  \]

Obviously, if we assume that the case slots are independent, then we need
only compare $P_{\rm fly}(X_{\rm from}=1)$ and $P_{\rm jet}(X_{\rm
from}=1)$. This is equivalent to the method proposed by
 \cite{Hindle91}. Their method actually compares the two probabilities
by means of hypothesis testing.

It is here that we first employ the proposed dependency learning
method to judge if slots $X_{\rm arg2}$ and $X_{\rm from}$ with
respect to verb `fly' are mutually dependent; if they are dependent,
we make a disambiguation decision based on the t-score between $P_{\rm
  fly}(X_{\rm from}=1|X_{\rm from}=1)$ and $P_{\rm jet}(X_{\rm
  from}=1)$; otherwise, we consider the two slots independent and make
a decision based on the t-score between $P_{\rm fly}(X_{\rm from}=1)$
and $P_{\rm jet}(X_{\rm from}=1)$. I refer to this method as
`DepenLA.'

In the experiment, I first randomly selected the files under one
directory for a portion of the WSJ corpus, a portion containing
roughly one $26$th of the entire bracketed corpus data, and extracted
$(v,n_1,p,n_2)$ quadruples (e.g., (fly, jet, from, Tokyo)) as test
data.  I then extracted case frames from the remaining bracketed
corpus data as I did in Experiment 1 and used them as training data. I
repeated this process ten times and obtained ten data sets consisting
of different training data and test data. In each training data set,
there were roughly $128,000$ case frames on average for verbs and
roughly $59,000$ case frames for nouns. On average, there were $820$
quadruples in each test data set.

I used these ten data sets to conduct disambiguation through cross
validation. I used the training data to acquire dependency forest
models, which I then used to perform disambiguation on the test data
on the basis of DepenLA. I also tested the method of LA. I set the
threshold for the t-score to $1.28$.  For both LA and DepenLA, there
were still some quadruples remaining whose attachment sites could not
be determined. In such cases, I made a default decision, i.e.,
forcibly attached $(p,n_2)$ to $v$, because I empirically found that,
at least for our data set for what remained after applying LA and
DepenLA, it is more likely for $(p,n_2)$ to go with $v$. 
Tab.~\ref{tab:disam} summarizes the results, which are evaluated in
terms of disambiguation accuracy, averaged over the ten trials.

\begin{table}[htb]
\begin{center}
\caption{PP-attachment disambiguation results.}
\label{tab:disam}
\begin{tabular}{|lc|} \hline
Method & Accuracy($\%$) \\ \hline
Default & $56.2$ \\
LA+Default & $78.1$ \\ 
DepenLA+Default & $78.4$ \\ 
LA+Default($11\%$ of data) & $93.8$ \\ 
DepenLA+Default($11\%$ of data) & $97.5$ \\ 
\hline
\end{tabular}
\end{center}
\end{table}

I found that as a whole DepenLA+Default only slightly improves
LA+Default. I further found, however, that for about $11\%$ of the
data in which the dependencies are strong (i.e., $P(X_p=1|X_{\rm
  arg2}=1)> 0.2$ or $P(X_p=1|X_{\rm arg2}=1)<0.002$), DepenLA+Default
significantly improves LA+Default. That is to say that when
significant dependencies between case slots are found, the
disambiguation results can be improved by using dependency knowledge. 
These results to some extent agree with the perplexity reduction
results obtained in Experiment 1.

\subsection{Experiment 3: class-based model}

I also used the $354$ verbs in Experiment 1 to acquire case frame
patterns as class-based dependency forest models.  Again, I considered
only the 12 slots listed in Table~\ref{tab:slotname}. I generalized
the values of the case slots within these case frames using the method
proposed in Chapter 4 to obtain class-based case frame data like those
presented in Table~\ref{tab:cfdata2}.\footnote{Since a node in WordNet
  represents a word sense and not a word, a word can belong to several
  different classes (nodes) in an output tree cut model. I have
  heuristically replaced a word $n$ with the word class $C$ such that
  $\max_{C \ni n} (P(C|v,r))$ is satisfied.} I used these data as
input to the learning algorithm.

On average, there was only a $64/354=0.2$ dependency link found in the
patterns for a verb. That is, very few case slots were determined to
be dependent in the case frame patterns. This is because the number of
parameters in a class based model was larger than the size of the data
we had available.

The experimental results indicate that it is often valid in practice
to assume that class-based case slots (and also word-based case slots)
are mutually independent, when the data size available is at the level
of what is provided by Penn Tree Bank. For this reason, I did not
conduct disambiguation experiments using the class-based
dependency forest models.

I believe that the proposed method provides a theoretically sound and
effective tool for detecting whether there exists a statistically
significant dependency between case slots in given data; this
decision has up to now been based simply on human intuition.

\subsection{Experiment 4: simulation}

\begin{figure}[htb]
\begin{center}
\epsfxsize9cm\epsfysize6cm\epsfbox{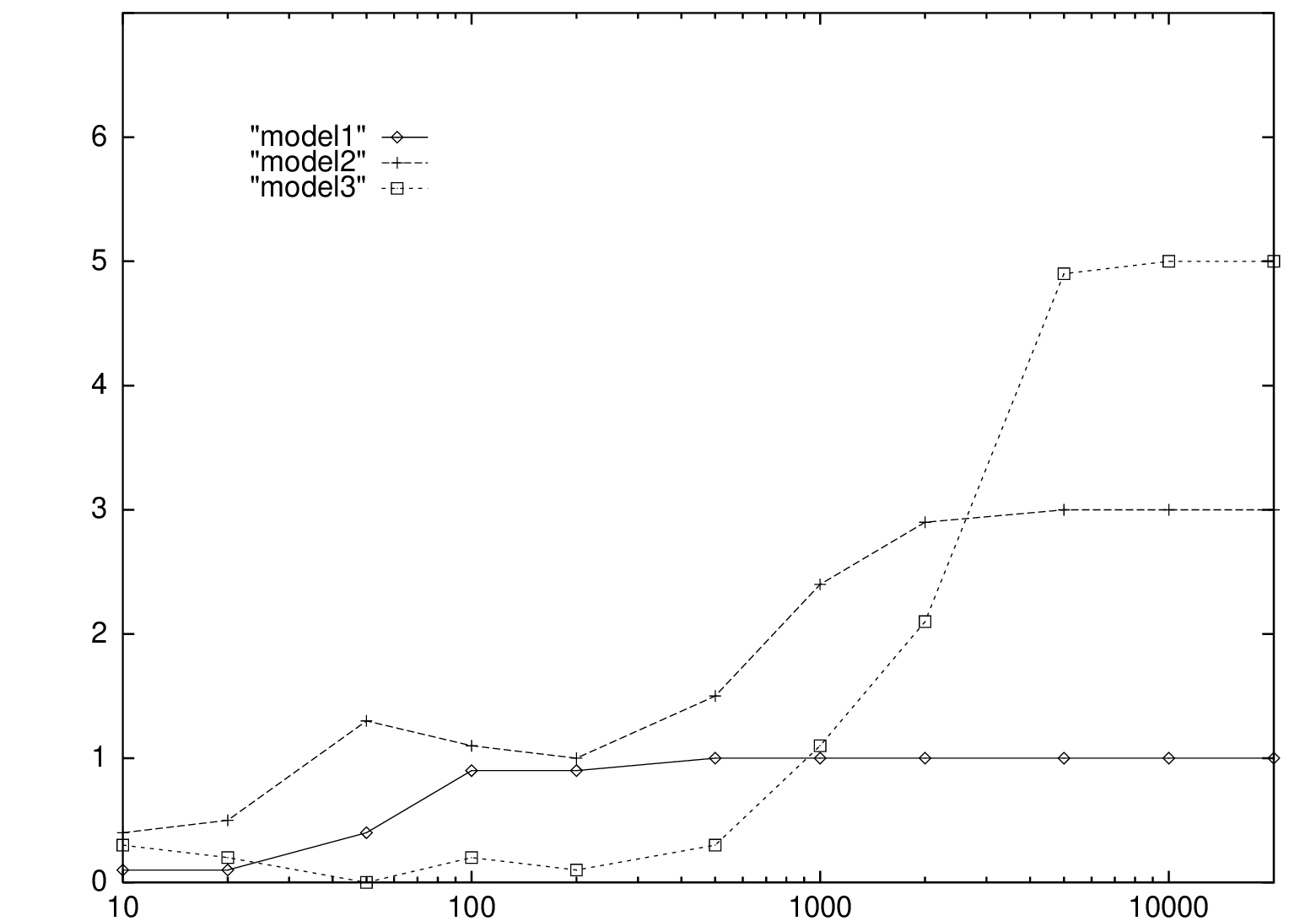}
\end{center}
\caption{Number of links versus data size.}
\label{fig:sima} 
\end{figure}

\begin{figure}[htb]
\begin{center}
\epsfxsize9cm\epsfysize6cm\epsfbox{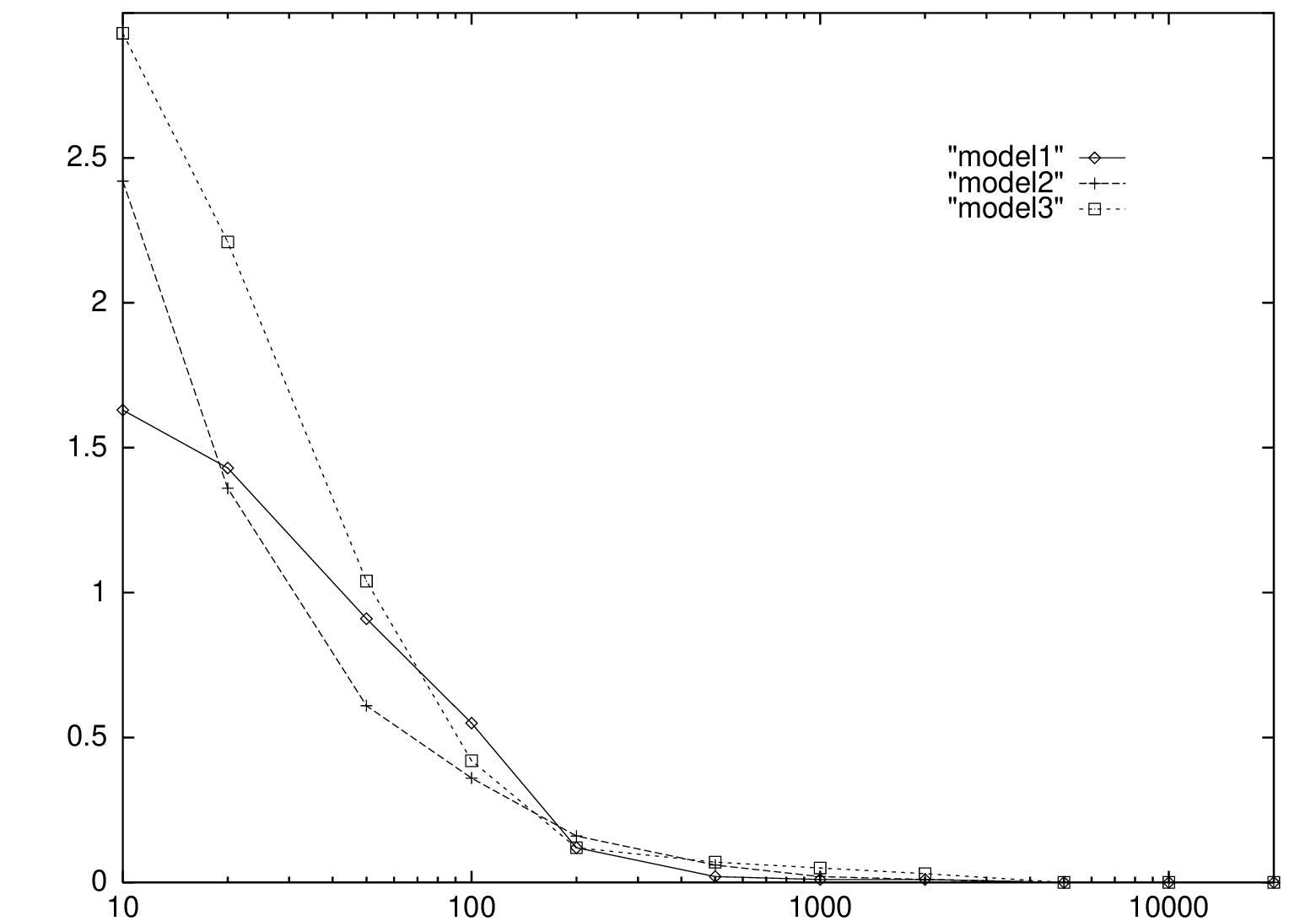}
\end{center}
\caption{KL divergence versus data size.}
\label{fig:simb} 
\end{figure}

In order to test how large a data size is required to estimate a
dependency forest model, I conducted the following experiment. I
defined an artificial model in the form of a dependency forest model
and generated data on the basis of its distribution. I then used the
obtained data to estimate a model, and evaluated the estimated model
by measuring the KL divergence between the estimated model and the
true model. I also checked the number of dependency links in the
obtained model. I repeatedly generated data and observed the `learning
curve,' namely the relationship between the data size used in
estimation and the number of links in the estimated model, and the
relationship between the data size and the KL divergence separating
the estimated and the true model. I defined two other artificial
models and conducted the same experiments. Figures~\ref{fig:sima} and
\ref{fig:simb} show the results of these experiments for the three
artificial models averaged over $10$ trials. The number of parameters
in Model 1, Model 2, and Model 3 are $18$, $30$, and $44$
respectively, and the number of links in them $1$, $3$, and $5$. Note
that the KL divergences between the estimated models and the true
models converge to $0$, as expected. Also note that the numbers of
links in the estimated models converge to the correct value (1, 3, and
5) in each of the three examples.

These simulation results verify the consistency property of MDL (i.e.,
the numbers of parameters in the selected models converge in
probability to that of the true model as the data size increases),
which is crucial for the goal of learning dependencies. Thus we can be
confident that the dependencies between case slots can be accurately
learned when there are enough data, as long as the `true' model exists
as a dependency forest model.

We also see that to estimate a model accurately the data size required
is as large as $5$ to $10$ times the number of parameters. For
example, for the KL divergence to go to below $0.1$, we need more than
$200$ examples, which is roughly $5$ to $10$ times the number of
parameters.

Note that in Experiment 3, I considered 12 slots, and for each slot
there were roughly 10 classes as its values; thus a class-based model
tended to have about $120$ parameters. The corpus data available to us
was insufficient for accurate learning of the dependencies between
case slots for most verbs (cf., Table~\ref{tab:verb}).

\section{Summary}

I conclude this chapter with the following remarks.
\begin{enumerate}
\item The primary contribution of the research reported in this
chapter is the proposed method of learning dependencies between case
slots, which is theoretically sound and efficient.
\item For slot-based models, some case slots are found to be
dependent. Experimental results demonstrate that by using the
knowledge of dependency, when dependency does exist, we can
significantly improve pp-attachment disambiguation results.
\item For class-based models, most case slots are judged
  independent with the data size currently available in the Penn Tree
  Bank. This empirical finding indicates that it is often valid to
  assume that case slots in a class-based model are mutually
  independent.
\end{enumerate}

The method of using a dependency forest model is not limited to just
the problem of learning dependencies between case slots. It is
potentially useful in other natural language processing tasks, such as
word sense disambiguation (cf., (\cite{Bruce94})).

\chapter{Word Clustering}

\begin{tabular}{p{5.5cm}r}
 &
\begin{minipage}{10cm}
\begin{tabular}{p{9cm}}
  {\em We may add that objects can be classified, and can become
    similar or dissimilar, only in this way - by being related to
    needs and interests.} \\ \multicolumn{1}{r}{- Karl Popper} \\ 
\end{tabular}
\end{minipage}
\end{tabular}
\vspace{0.5cm}

In this chapter, I describe one method for learning the hard
co-occurrence model, i.e., clustering of words on the basis
of co-occurrence data. This method is a natural extension of that
proposed by Brown et al (cf., Chapter 2), and it overcomes the
drawbacks of their method while retaining its merits.

\section{Parameter Estimation}

As described in Chapter 3, we can view the problem of clustering words
(constructing a thesaurus) on the basis of co-occurrence data as that
of estimating a hard co-occurrence model.

The fixing of partitions determines a discrete hard co-occurrence
model and the number of parameters. We can estimate the values
of the parameters on the basis of co-occurrence data by
employing Maximum Likelihood Estimation (MLE). For given
co-occurrence data \[ {\cal S} =
\{(n_1,v_1),(n_2,v_2),\cdots,(n_m,v_m)\}, \] where $n_i \
(i=1,\cdots,m)$ denotes a noun, and $v_i \ (i=1,\cdots,m)$ a verb.
The maximum likelihood estimates of the parameters are
defined as the values that maximize the following likelihood
function with respect to the data:
\[ \prod_{i=1}^{m}
P(n_i,v_i) = \prod_{i=1}^{m} (P(n_i|C_{n_i}) \cdot
P(v_i|C_{v_i}) \cdot P(C_{n_i},C_{v_i})).  \]

It is easy to verify that we can estimate the parameters as \[
\hat{P}(C_n,C_v)= \frac{f(C_n,C_v)}{m} \] \[ \hat{P}(n|C_n)=
\frac{f(n)}{f(C_n)} \] \[ \hat{P}(v|C_v)= \frac{f(v)}{f(C_v)}, \] so
as to maximize the likelihood function, under the conditions that the
sum of the joint probabilities over noun classes and verb classes
equals one, and that the sum of the conditional probabilities over
words in each class equals one. Here, $m$ denotes the entire data
size, $f(C_n,C_v)$ the frequency of word pairs in class pair
$(C_n,C_v)$, $f(n)$ the frequency of noun $n$, $f(v)$ that of $v$,
$f(C_n)$ the frequency of words in class $C_n$, and $f(C_v)$ that in
$C_v$.

\section{MDL as Strategy}

I again adopt the MDL principle as a strategy for statistical
estimation.

Data description length may be calculated as
\[
L({\cal S}|M) = - \sum_{(n,v) \in {\cal S}} \log \hat{P}(n,v).
\]

Model description length may be calculated, here, as
\[
L(M) = \frac{k}{2} \cdot \log m, \] where $k$ denotes the number of
{\em free} parameters in the model, and $m$ the data size. We in fact
implicitly assume here that the description length for encoding the
discrete model is equal for all models and view only the description
length for encoding the parameters as the model description length. 
Note that there are alternative ways of calculating the model
description length. Here, for efficiency in clustering, I use the
simplest formulation.

If computation time were of no concern, we could in principle
calculate the total description length for each model and select the
optimal model in terms of MDL. However, since the number of hard
co-occurrence models is of order $O(N^N \cdot V^V)$ (cf., Chapter 4),
where $N$ and $V$ denote the sizes of the set of nouns and the set of
verbs respectively, it would be infeasible to do so. We therefore need
to devise an efficient algorithm that will heuristically perform this
task.

\section{Algorithm}

The algorithm that we have devised, denoted here as `2D-Clustering,'
iteratively selects a suboptimal MDL model from among a class of hard
co-occurrence models. These models include the current model and those
which can be obtained from the current model by merging a noun (or
verb) class pair. The minimum description length criterion can be
reformalized in terms of (empirical) mutual information. The algorithm
can be formulated as one which calculates, in each iteration, the
reduction of mutual information which would result from merging any
noun (or verb) class pair. It would perform the merge having the least
mutual information reduction, {\em provided that the least mutual
information reduction is below a threshold, which will vary
  depending on the data size and the number of classes in the current
  situation}.

\vspace{12pt}
\noindent{2D-Clustering$({\cal S})$}
\\ ${\cal S}$ is input co-occurrence data. $b_n$ and $b_v$ are positive
integers.
\begin{enumerate}
\item Initialize the set of noun classes $\Pi_n$ and the set of
  verb classes $\Pi_v$ as:
\[
\Pi_n = \{ \{ n \} | n \in {\cal N} \}
\]
\[
\Pi_v = \{ \{ v \} | v \in {\cal V} \}
\]
${\cal N}$ and ${\cal V}$ denote the set of nouns and the set of
verbs, respectively.
\item Repeat the following procedure:
\begin{enumerate}
\item execute Merge$({\cal S}, \Pi_n, \Pi_v, b_n)$ to update $\Pi_n$,
\item execute Merge$({\cal S}, \Pi_v, \Pi_n, b_v)$ to update $\Pi_v$,
\item if $\Pi_n$ and $\Pi_v$ are unchanged, go to Step 3.
\end{enumerate}
\item Construct and output a thesaurus of nouns based on the history
of $\Pi_n$, and one for verbs based on the history of $\Pi_v$.
\end{enumerate} \vspace{12pt}

For the sake of simplicity, let us next consider only the procedure
for Merge as it is applied to the set of noun classes while the set
of verb classes is fixed.

\vspace{12pt} \noindent{Merge$({\cal S}, T_n, T_v, b_n)$}
\begin{enumerate} 
\item For each class pair in $T_n$, calculate the
reduction in mutual information which would result from merging
them. (Details of such a calculation are given below.) Discard those
class pairs whose mutual information reduction is not less than the
threshold of
\begin{equation}\label{eq:threshold}
\frac{(k_B - k_A)\cdot\log m}{2 \cdot m},
\end{equation}
where $m$ denotes total data size, $k_B$ the number of free parameters
in the model before the merge, and $k_A$ the number of free parameters
in the model after the merge. Sort the remaining class pairs in
ascending order with respect to mutual information reduction.
\item Merge the first $b_n$ class pairs in the 
sorted list.
\item Output current $T_n$.
\end{enumerate}
\vspace{12pt}

For improved efficiency, the algorithm performs a maximum of $b_n$
merges at step 2, which will result in the output of an at most
$b_n$-ary tree. Note that, strictly speaking, once we perform one
merge, the model will change and there will no longer be any guarantee
that the remaining merges continue to be justifiable from the
viewpoint of MDL.

Next, let us consider why the criterion formalized in terms of
description length can be reformalized in terms of mutual information. 
Let $M_B$ refer to the pre-merge model, $M_A$ to the post-merge model. 
According to MDL, $M_A$ should be that model which has the least
increase in data description length \[ \delta L_{dat} = L({\cal
S}|M_A) - L({\cal S}|M_B) > 0 \] and that at the same time satisfies
\[ \delta L_{dat} < \frac{(k_B-k_A) \cdot \log m}{2}, \] since the
decrease in model description length equals \[ L(M_B) - L(M_A) =
\frac{(k_B-k_A) \cdot \log m}{2} > 0,  \] and the decrease in model
description length is common to each merge.

In addition, suppose that $M_A$ is obtained by merging two noun
classes $C_i$ and $C_j$ in $M_B$ to a noun class $C_{ij}$. We in
fact need only calculate the difference in description lengths with
respect to these classes, i.e., 
\[ \begin{array}{lll}
  \delta L_{dat} & = & - \sum_{C_v \in \Pi_v}\sum_{n \in C_{ij},v \in C_v}\log \hat{P}(n,v) + \sum_{C_v \in \Pi_v}\sum_{n \in C_i,v \in C_v}\log \hat{P}(n,v) \\ 
& & + \sum_{C_v \in \Pi_v}\sum_{n \in C_j, v \in C_v}\log \hat{P}(n,v). \\ 
\end{array}
\]
Since
\[
P(n,v)  =  \frac{P(n)}{P(C_n)}\cdot \frac{P(v)}{P(C_v)}\cdot P(C_n,C_v) =
\frac{P(C_n,C_v)}{P(C_n)P(C_v)} \cdot P(n)\cdot P(v)
\]
holds, we also have
\[
\hat{P}(n) = \frac{f(n)}{m},
\]
\[
\hat{P}(v) = \frac{f(v)}{m},
\]
\[
\hat{P}(C_n) = \frac{f(C_n)}{m},
\]
and
\[
\hat{P}(C_v) = \frac{f(C_v)}{m}.
\]
Hence,
\begin{equation}\label{eq:deltaLdat}
\begin{array}{lll}
  \delta L_{dat} & = & -
  \sum_{C_v \in \Pi_v} f(C_{ij},C_v) \cdot \log\frac{\hat{P}(C_{ij},C_v)}{\hat{P}(C_{ij})\hat{P}(C_v)} + \sum_{C_v \in \Pi_v}f(C_i,C_v)\cdot\log\frac{\hat{P}(C_i,C_v)}{\hat{P}(C_i)\hat{P}(C_v)} \\ 
  & & + \sum_{C_v \in \Pi_v}
  f(C_j,C_v)\cdot\log\frac{\hat{P}(C_j,C_v)}{\hat{P}(C_j)\hat{P}(C_v)}. \\ 
\end{array}
\end{equation}
The quantity $\delta L_{dat}$ is equivalent to the data size times the
empirical mutual information reduction. We can, therefore, say that in
the current context a clustering with the least data description
length increase is equivalent to that with the least mutual
information decrease.

Note further that in (\ref{eq:deltaLdat}), since $\hat{P}(C_v)$ is
unchanged before and after the merge, it can be canceled
out. Replacing the probabilities with their maximum likelihood
estimates, we obtain
\[
 \begin{array}{lll}
  \frac{1}{m} \cdot \delta L_{dat} & = & \frac{1}{m} \cdot \biggl(
  - \sum_{C_v \in \Pi_v} (f(C_i,C_v)+f(C_j,C_v))\cdot \log\frac{f(C_i,C_v)+f(C_j,C_v)}{f(C_i)+f(C_j)} \\
 & & + \sum_{C_v \in \Pi_v}f(C_i,C_v)\cdot\log\frac{f(C_i,C_v)}{f(C_i)}
 + \sum_{C_v \in \Pi_v} f(C_j,C_v)\cdot\log\frac{f(C_j,C_v)}{f(C_j)} \biggr). \\ 
\end{array}
\]
We need calculate only this quantity for each possible merge at Step 1
in Merge.

In an implementation of the algorithm, we first load the co-occurrence
data into a matrix, with nouns corresponding to rows, verbs to
columns. When merging a noun class in row $i$ and that in row $j$
($i < j$), for each $C_v$, we add $f(C_i,C_v)$ and $f(C_j,C_v)$,
obtaining $f(C_{ij},C_v)$; then write $f(C_{ij},C_v)$ on row $i$; move
$f(C_{last},C_v)$ to row $j$. This reduces the matrix by one row.

With the above implementation, the worst case time complexity
of the algorithm turns out to be $O(N^3 \cdot V + V^3 \cdot
N)$, where $N$ denotes the size of the set of nouns, and $V$
that of verbs. If we can merge $b_n$ and $b_v$ classes at each
step, the algorithm will become slightly more efficient, with a
time complexity of $O(\frac{N^3}{b_n} \cdot V + \frac{V^3}{b_v}
\cdot N)$.

The method proposed in this chapter is an extension of that proposed
by Brown et al. Their method iteratively merges the word class pair
having the least reduction in mutual information until the number of
word classes created equals a certain designated number. This method
is based on MLE, but it only employs MLE {\em locally}.

In general, MLE is not able to select the best model from a class of
models having different numbers of parameters because MLE will always
suggest selecting the model having the largest number of parameters,
which would have a better fit to the given data. In Brown et al's
case, MLE is used to iteratively select the model with the maximum
likelihood from a class of {\em models that have the same number of
  parameters}. Such a model class is repeatedly obtained by merging
any word class pair in the current situation. The number of word
classes within the models in the final model class, therefore, has to
be designated in advance. There is, however, no guarantee at all the
designated number will be optimal.

The method proposed here resolves this problem by employing MDL. This
is reflected in use of the threshold (\ref{eq:threshold}) in
clustering, which will result in automatic selection of the optimal
number of word classes to be created.

\section{Experimental Results}

\subsection{Experiment 1: qualitative evaluation} In this experiment,
I used heuristic rules to extract verbs and their arg2 slot values
(direct objects) from the {\em tagged} texts of the WSJ corpus
(ACL/DCI CD-ROM1) which consists of 126,084 sentences.

\begin{figure}[htb]
\begin{center}
\epsfxsize7cm\epsfysize7cm\epsfbox{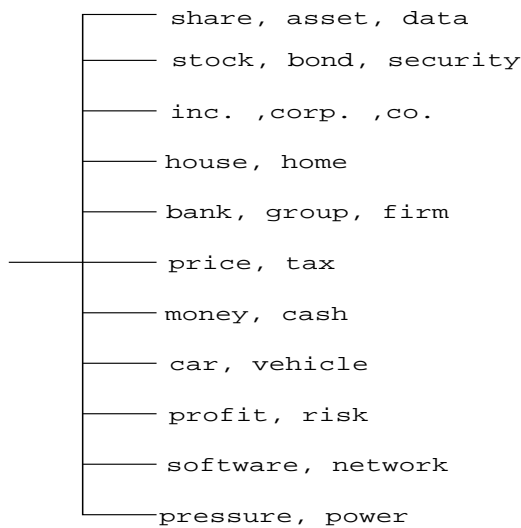}
\end{center}
\caption{A part of a constructed thesaurus.}
\label{fig:exthesau}
\end{figure}

I then constructed a number of thesauruses based on these data, using
the method proposed in this chapter. Figure~\ref{fig:exthesau} shows a
part of a thesaurus for 100 randomly selected nouns, that serve as
direct objects of 20 randomly selected verbs. The thesaurus seems to
agree with human intuition to some degree. The words `stock,'
`security,' and `bond' are classified together, for example, despite
the fact that their absolute frequencies are quite different (272, 59,
and 79, respectively). The results seem to demonstrate one desirable
feature of the proposed method: it classifies words solely on the
basis of the similarities in co-occurrence data and is not affected by
the absolute frequencies of the words.

\subsection{Experiment 2: compound noun disambiguation} 

I tested the effectiveness of the clustering method by using the
acquired word classes in compound noun disambiguation. This would
determine, for example, the word `base' or `system' to which `data'
should be attached in the compound noun triple (data, base, system).

To conduct compound noun disambiguation, we can use here the
probabilities
\begin{equation}\label{eq:compoundprob1}
  \hat{P}({\rm data} | {\rm base}),
\end{equation}
\begin{equation}\label{eq:compoundprob2}
\hat{P}({\rm data} | {\rm system}).
\end{equation}
If the former is larger, we attach `data' to `base;' if the latter is
larger we attach it to `system;' otherwise, we make no decision.

I first randomly selected 1000 nouns from the corpus, and extracted
from the corpus compound noun doubles (e.g., (data, base)) containing
the nouns as training data and compound noun triples containing the
nouns as test data. There were 8604 training data and 299 test data. I
also labeled the test data with disambiguation `answers.'

I conducted clustering on the nouns in the left position in
the training data, and also on the nouns in the right position, by
using, respectively, both the method proposed in this chapter, denoted
as `2D-Clustering,' and Brown et al's, denoted as `Brown.' I actually
implemented an extended version of their method, which separately
conducts clustering for nouns on the left and those on the right
(which should only improve the performance).

\begin{figure}[htb] \begin{center}
\epsfxsize8cm\epsfysize6cm\epsfbox{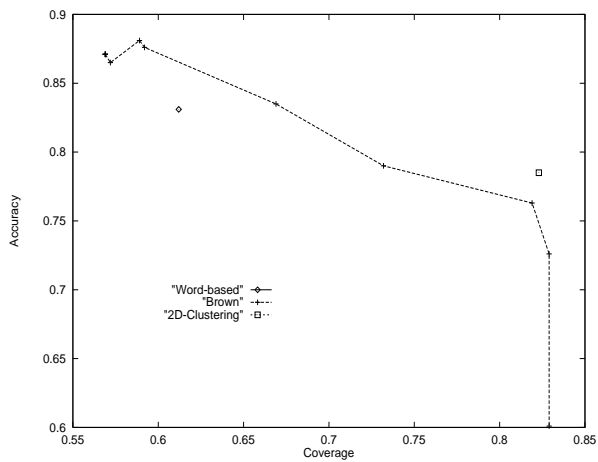}
\caption{Accuracy-coverage plots for 2D-Clustering, Brown, and Word-based.} 
\label{fig:compound} 
\end{center}
\end{figure}

I conducted structural disambiguation on the test data, using the
probabilities like those in (\ref{eq:compoundprob1}) and
(\ref{eq:compoundprob2}), estimated on the basis of 2D-Clustering and
Brown, respectively. I also tested the method of using probabilities
estimated based on word occurrences, denoted here as `Word-based.'

Figure~\ref{fig:compound} shows the results in terms of accuracy and
coverage, where `coverage' refers to the percentage of test data for
which the disambiguation method was able to make a decision. Since for
Brown the number of word classes finally created has to be designed in
advance, I tried a number of alternatives and obtained results for
them (for 2D-Clustering, the optimal number of word classes is
automatically selected). We see that, for Brown, when the number of
word classes finally to be created is small, though the coverage will
be large, the accuracy will deteriorate dramatically, indicating that
in word clustering it is preferable to introduce a mechanism to
automatically determine the final number of word classes.

\begin{table}[htb]
\caption{Compound noun disambiguation results.}
\label{tab:compound}
\begin{center}
\begin{tabular}{|lc|} \hline
Method & Accuracy($\%$) \\ \hline
Default & $59.2$ \\
Word-based + Default & $73.9$ \\
Brown + Default & $77.3$ \\
2D-Clustering + Default & $78.3$ \\ \hline
\end{tabular}
\end{center}
\end{table}

Table~\ref{tab:compound} shows final results for the above methods
combined with `Default' in which we attach the first noun to the
neighboring noun when a decision cannot be made by an individual
method.

We can see here that 2D-Clustering performs the best. These results
demonstrate one desirable aspect of 2D-Clustering: its ability
to {\em
  automatically} select the most appropriate level of clustering,
i.e., it results in neither over-generalization nor
under-generalization. (The final result of 2D-Clustering is still not
completely satisfactory, however. I think that this is partly due to
insufficient training data.)

\subsection{Experiment 3: pp-attachment disambiguation}

I tested the effectiveness of the proposed method by using the
acquired classes in pp-attachment disambiguation involving
quadruples ($v, n_1, p, n_2$).

As described in Chapter 3, in disambiguation of (eat, ice-cream, with,
spoon), we can perform disambiguation by comparing the probabilities
\begin{equation}\label{eq:pp-prob1} 
\hat{P}_{\rm with}({\rm spoon} | {\rm eat}),
\end{equation} 
\begin{equation}\label{eq:pp-prob2} 
\hat{P}_{\rm with}({\rm spoon} | {\rm ice\_cream}).  
\end{equation}
If the former is larger, we attach `with spoon' to `eat;' if the
latter is larger we attach it to `ice-cream;' otherwise, we make no
decision.

I used the ten sets used in Experiment 2 in Chapter 4, and conducted
experiments through `ten-fold cross validation,' i.e., all of the
experimental results reported below were obtained from averages taken
over ten trials.

\begin{table}[htb]
\caption{PP-attachment disambiguation results.}
\label{tab:pp-attach}
\begin{center}
\begin{tabular}{|lcc|} \hline
Method & Coverage($\%$) & Accuracy($\%$) \\ \hline
Default & $100$ & $56.2$ \\
Word-based & $32.3$ & $95.6$ \\
Brown & $51.3$ & $98.3$ \\
2D-Clustering & $51.3$ & $98.3$ \\
WordNet & $74.3$ & $94.5$ \\
1D-Thesaurus & $42.6$ & $97.1$ \\ 
\hline
\end{tabular}
\end{center}
\end{table}

I conducted word clustering by using the method proposed in
this chapter, denoted as `2D-Clustering,' and the method proposed in
\cite{Brown92}, denoted as `Brown.' In accord with the proposal
offered by \cite{Tokunaga95}, for both methods, I separately conducted
clustering with respect to each of the 10 most frequently occurring
prepositions (e.g., `for,' `with,' etc). I did not cluster words with
respect to rarely occurring prepositions. I then performed
disambiguation by using probabilities estimated based on 2D-Clustering
and Brown. I also tested the method of using the probabilities
estimated based on word co-occurrences, denoted here as `Word-based.'

Next, rather than using the co-occurrence probabilities estimated by
2D-Clustering, I only used the noun thesauruses constructed by
2D-Clustering, and applied the method of estimating the best tree cut
models within the thesauruses in order to estimate conditional
probabilities like those in (\ref{eq:pp-prob1}) and
(\ref{eq:pp-prob2}). I call this method `1D-Thesaurus.'

Table~\ref{tab:pp-attach} shows the results for all these methods in
terms of coverage and accuracy.  It also shows the results obtained in
the experiment described in Chapter2, denoted here as `WordNet.'

I then enhanced each of these methods by using a default decision of
attaching $(p,n_2)$ to $n_1$ when a decision cannot be made. This is
indicated as `Default.' Table~\ref{tab:pp-attach-final} shows the
results of these experiments.

\begin{table}[htb]
\caption{PP-attachment disambiguation results.}
\label{tab:pp-attach-final}
\begin{center}
\begin{tabular}{|lc|} \hline
Method & Accuracy($\%$) \\ \hline
Word-based + Default & $69.5$ \\
Brown + Default & $76.2$ \\
2D-Clustering + Default & $76.2$ \\
WordNet + Default & $82.2$ \\  
1D-Thesaurus + Default & $73.8$ \\ \hline
\end{tabular}
\end{center}
\end{table}

We can make a number of observations from these results. (1)
2D-Clustering achieves broader coverage than does 1D-Thesaurus. This
is because, in order to estimate the probabilities for disambiguation,
the former exploits more information than the latter. (2) For Brown, I
show here only its best result, which happens to be the same as the
result for 2D-Clustering, but in order to obtain this result I had to
take the trouble of conducting a number of tests to find the best
level of clustering. For 2D-Clustering, this needed to be done only
once and could be done automatically. (3) 2D-Clustering outperforms
WordNet in term of accuracy, but not in terms of coverage.  This seems
reasonable, since an automatically constructed thesaurus is more
domain dependent and therefore captures the domain dependent features
better, thus helping achieve higher accuracy.  On the other hand, with
the relatively small size of training data we had available, its
coverage is smaller than that of a general purpose hand-made
thesaurus. The result indicates that it makes sense to combine the use
of automatically constructed thesauruses with that of a hand-made
thesaurus. I will describe such a method and the experimental results
with regard to it in Chapter 7.

\section{Summary}

I have described in this chapter a method of clustering
words. That is a natural extension of Brown et al's
method. Experimental results indicate that it is superior to theirs.

The proposed clustering algorithm, 2D-Clustering, can be used in
practice so long as the data size is at the level of the current Penn
Tree Bank. It is still relatively computationally demanding, however,
and the important work of improving its efficiency remains to be
performed.

The method proposed in this chapter is not limited to word clustering;
it can be applied to other tasks in natural language processing and
related fields, such as, document classification (cf.,
\cite{Iwayama95}).

\chapter{Structural Disambiguation}

\begin{tabular}{p{5.5cm}r}
 &
\begin{minipage}{10cm}
\begin{tabular}{p{9cm}}
  {\em To have good fruit you must have a healthy tree; if you have a poor tree you will have bad fruit.} \\ \multicolumn{1}{r}{- The Gospel according to Matthew} \\ 
\end{tabular}
\end{minipage}
\end{tabular}
\vspace{0.5cm}

In this chapter, I propose a practical method for pp-attachment
disambiguation. This method combines the use of the hard co-occurrence
model with that of the tree cut model.

\section{Procedure}\label{sec:disam}

Let us consider here the problem of structural disambiguation, in
particular, the problem of resolving pp-attachment ambiguities
involving quadruples ($v, n_1, p, n_2$), such as (eat, ice-cream,
with, spoon).

As described in Chapter 6, we can resolve such an ambiguity by using
probabilities estimated on the basis of hard co-occurrence models. I
denote them as
\[
\hat{P}_{\rm hcm}({\rm spoon}|{\rm eat},{\rm with}),
\]
\[
\hat{P}_{\rm hcm}({\rm spoon}|{\rm ice\_cream},{\rm with}).
\]
Further, as described in Chapter 4, we can also resolve the ambiguity
by using probabilities estimated on the basis of tree cut models {\em
with respect to a hand-made thesaurus}, denoted as
\[
\hat{P}_{\rm tcm}({\rm spoon}|{\rm eat},{\rm with}),
\]
\[
\hat{P}_{\rm tcm}({\rm spoon}|{\rm ice\_cream},{\rm with}).
\]

Both methods are a class-based approach to disambiguation, and thus
can help to handle the data sparseness problem. The former method is
based on corpus data and thus can capture domain specific features and
achieve higher accuracy. At the same time, since corpus data is never
sufficiently large, coverage is bound to be less than satisfactory. By
way of contrast, the latter method is based on human-defined knowledge
and thus can bring about broader coverage. At the same time, since the
knowledge used is not domain-specific, accuracy might be expected to
be less than satisfactory. Since both methods have pros and cons, it
would seem be better to combine the two, and I propose here a back-off
method to do so.

In disambiguation, we first use probabilities estimated based on hard
co-occurrence models; if the probabilities are equal (particularly both
of them are 0), we use probabilities estimated based on tree cut
models with respect to a hand-made thesaurus; if the probabilities are
still equal, we make a default
decision. Figure~\ref{fig:disambiguation} shows the procedure of this
method.

\begin{figure}[htb]
\begin{tabbing}
{\bf Procedure:} \\
1. Take $(v, n_1, p, n_2)$ as input; \\
2. {\bf if} \\
3. \tab $\hat{P}_{\rm hcm}(n_2|v,p) > \hat{P}_{\rm hcm}(n_2|n_1,p)$ \\
4. {\bf then} \\
5. \tab attach $(p,n_2)$ to $v$; \\
6. {\bf else if} \\
7. \tab $\hat{P}_{\rm hcm}(n_2|v,p) < \hat{P}_{\rm hcm}(n_2|n_1,p)$ \\
8. {\bf then} \\
9. \tab attach $(p,n_2)$ to $n_1$; \\
10. {\bf else} \\
11.  \tab {\bf if} \\
12. \tab \tab $\hat{P}_{\rm tcm}(n_2|v,p) > \hat{P}_{\rm tcm}(n_2|n_1,p)$ \\
13. \tab {\bf then} \\
14. \tab \tab attach $(p,n_2)$ to $v$; \\
15. \tab {\bf else if} \\
16. \tab \tab $\hat{P}_{\rm tcm}(n_2|v,p) < \hat{P}_{\rm tcm}(n_2|n_1,p)$ \\
17. \tab {\bf then} \\
18. \tab \tab attach $(p,n_2)$ to $n_1$; \\
19. \tab {\bf else} \\
20. \tab \tab make a default decision. \\
\end{tabbing}
\caption{The disambiguation procedure.}
\label{fig:disambiguation}
\end{figure}

\section{An Analysis System}

Let us consider this disambiguation method in more general terms. The
natural language analysis system that implements the method operates
on the basis of two processes: a learning process and an analysis
process.

During the learning process, the system takes natural language
sentences as input and acquires lexical semantic knowledge. First, the
POS (part-of-speech) tagging module uses a probabilistic tagger
(cf., Chapter 2) to assign the most likely POS tag to each word in the
input sentences. The word sense disambiguation module then employs a
probabilistic model (cf., Chapter 2) to resolve word sense
ambiguities. Next, the case frame extracting module employs a
heuristic method (cf., Chapter 2) to extract case frame
instances. Finally, the learning module acquires lexical semantic
knowledge (case frame patterns) on the basis of the case frame
instances.

During the analysis process, the system takes a sentence as input and
outputs a most likely interpretation (or several most likely
interpretations). The POS tagging module assigns the most likely tag
to each word in the input sentence, as is in the case of learning. The
word sense disambiguation module then resolves word sense ambiguities,
as is in the case of learning. The parsing module then analyzes the
sentence. When ambiguity arises, the structural disambiguation module
refers to the acquired knowledge, calculates the likelihood values of
the ambiguous interpretations (case frames) and selects the most
likely interpretation as the analysis result.

\begin{figure}[htb]
\begin{center}
\epsfxsize9cm\epsfysize9cm\epsfbox{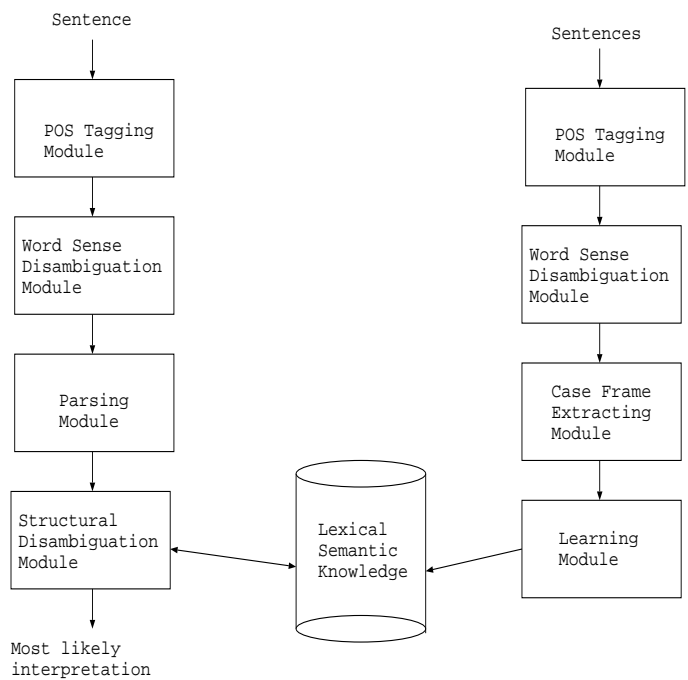}
\end{center}
\caption{Outline of the natural language analysis system.}
\label{fig:outline} 
\end{figure}

Figure~\ref{fig:outline} shows an outline of the system. Note that
while for simplicity the parsing process and the disambiguation
process are separated into two modules, they can (and usually should)
be unified into one module. Furthermore, for simplicity some other
knowledge necessary for natural language analysis, e.g., a grammar,
has also been omitted from the figure.

The learning module consists of two submodules: a thesaurus
construction submodule, and a case slot generalization submodule. The
thesaurus construction submodule employs the hard co-occurrence model
to calculate probabilities. The case slot generalization submodule
then employs the tree cut model to calculate probabilities.

The structural disambiguation module refers to the probabilities, and
calculates likelihood for each interpretation.  The likelihood values
based on the hard co-occurrence model for the two interpretations of the
sentence (\ref{eq:icecream}) are calculated as follows
\[
L_{\rm hcm} (1) =
\hat{P}_{\rm hcm}({\rm I}|{\rm eat},{\rm arg1}) \cdot \hat{P}_{\rm hcm}({\rm ice\_cream}|{\rm eat},{\rm arg2}) \cdot \hat{P}_{\rm hcm}({\rm spoon}|{\rm eat},{\rm with})
\]
\[
L_{\rm hcm} (2) = \hat{P}_{\rm hcm}({\rm I}|{\rm eat},{\rm arg1}) \cdot \hat{P}_{\rm hcm}({\rm ice\_cream}|{\rm eat},{\rm arg2}) \cdot \hat{P}_{\rm hcm}({\rm spoon}|{\rm girl},{\rm with}).
\]
The likelihood values based on the tree cut model can be calculated
analogously. Finally, the disambiguation module selects the most
likely interpretation on the basis of a back-off procedure like that
described in Section 1.

Note that in its current state of development, the disambiguation
module is still unable to exploit syntactic knowledge. As described in
Chapter 2, disambiguation decisions may not be made solely on the
basis of lexical knowledge; it is necessary to utilize syntactic
knowledge as well. Further study is needed to determine how to define
a unified model which combines both lexical knowledge and syntactic
knowledge. In terms of syntactic factors, we need to consider
psycholinguistic principles, e.g., the `right association principle.' 
I have found in my study that using a probability model embodying
these principles helps improve disambiguation results
\cite{Li96a}. Another syntactic factor we need to take into
consideration is the likelihood of the phrase structure of an
interpretation (cf., \cite{Charniak97,Collins97,Shirai98}).

\section{Experimental Results}

I tested the proposed disambiguation method by using the data used in
Chapters 4 and 6. Table~\ref{tab:pp-attach-comp} shows the results;
here the method is denoted as `2D-Clustering+WordNet+Default.' 
Table~\ref{tab:pp-attach-comp} also shows the results of
WordNet+Default and TEL which were described in Chapter 4, and the
result of 2D-Clustering+Default which was described in Chapter 6. We
see that the disambiguation method proposed in this chapter performs
the best of four.

\begin{table}[htb]
\caption{PP-attachment disambiguation results.}
\label{tab:pp-attach-comp}
\begin{center}
\begin{tabular}{|lc|} \hline
TEL & $82.4$ \\ 
2D-Clustering + Default & $76.2$ \\
WordNet + Default & $82.2$ \\
2D-Clustering + WordNet + Default & $85.2$ \\
\hline
\end{tabular}
\end{center}
\end{table}

Table~\ref{tab:pp-attach-comp2} shows the disambiguation results
reported in other studies. Since the data sets used in the respective
studies were different, a straightforward comparison of the various
results would have little significance, we may say that the method
proposed in this chapter appears to perform relatively well with
respect to other state-of-the-art methods.

\begin{table}[htb]
\caption{Results reported in previous work.}
\label{tab:pp-attach-comp2}
\begin{center}
\begin{tabular}{|llc|} \hline
Method & Data & Accuracy ($\%$) \\ \hline
\cite{Hindle91} & AP News & $78.3$ \\
\cite{Resnik93a} & WSJ & $82.2$ \\
\cite{Brill94} & WSJ & $81.8$ \\
\cite{Ratnaparkhi94} & WSJ & $81.6$ \\
\cite{Collins95} & WSJ & $84.5$ \\ \hline
\end{tabular}
\end{center}
\end{table}

\chapter{Conclusions}

\begin{tabular}{p{5.5cm}r}
 &
\begin{minipage}{10cm}
\begin{tabular}{p{9cm}}
{\em If all I know is a fraction, then my only fear is of losing the thread.} \\
\multicolumn{1}{r}{- Lao Tzu} \\
\end{tabular}
\end{minipage}
\end{tabular}
\vspace{0.5cm}

\section{Summary}

The problem of acquiring lexical semantic knowledge is an important
issue in natural language processing, especially with regard to
structural disambiguation. The approach I have adopted here to this
problem has the following characteristics: (1) dividing the problem
into three subproblems: case slot generalization, case dependency
learning, and word clustering, (2) viewing each subproblem as that of
statistical estimation and defining probability models (probability
distributions) for each subproblem, (3) adopting MDL as a learning
strategy, (4) employing efficient learning algorithms, and (5) viewing
the disambiguation problem as that of statistical prediction.

Major contributions of this thesis include: (1) formalization of the
lexical knowledge acquisition problem, (2) development of a number of
learning methods for lexical knowledge acquisition, and (3)
development of a high-performance disambiguation method.

Table~\ref{tab:modellist} shows the models I have proposed, and
Table~\ref{tab:algolist} shows the algorithms I have employed. The
overall accuracy achieved by the pp-attachment disambiguation method
is $85.2\%$, which is better than that of state-of-the-art methods.

\begin{table}[htb]
\caption{Models proposed.}
\label{tab:modellist}
\begin{center}
\begin{tabular}{|lll|} \hline
Purpose & Basic model & Restricted model \\ \hline
case slot generalization & case slot model & tree cut model \\ 
 & (hard, soft) & \\
case dependency learning & case frame model & dependency forest model \\ 
 & (word-based, class-based, slot-based) & \\
word clustering & co-occurrence model & hard co-occurrence model \\ 
 & (hard, soft) & \\ \hline
\end{tabular}
\end{center}
\end{table}

\begin{table}[htb]
\caption{Algorithm employed.}
\label{tab:algolist}
\begin{center}
\begin{tabular}{|llc|} \hline
Purpose & Algorithm & Time complexity \\ \hline
case slot generalization & Find-MDL & $O(N)$ \\ 
case dependency learning & Suzuki's algorithm & $O(n^2(k^2+n^2))$ \\ 
word clustering & 2D-Clustering & $O(N^3 \cdot V + V^3 \cdot N)$ \\ \hline
\end{tabular}
\end{center}
\end{table}

\section{Open Problems}

Lexical semantic knowledge acquisition and structural disambiguation
are difficult tasks. Although I think that the investigations reported
in this thesis represent some significant progress, further research
on this problem is clearly still needed.

Other issues not investigated in this thesis and some possible
solutions include:

\begin{description} \item[More complicated models] In the discussions
so far, I have restricted the class of hard case slot models to that
of tree cut models for an existing thesaurus {\em tree}. Under this
restriction, we can employ an efficient dynamic-programming-based
learning algorithm which can provablely find the optimal MDL model. In
practice, however, the structure of a thesaurus may be a directed
acyclic graph (DAG) and straightforwardly extending the algorithm to a
DAG may no longer guarantee that the optimal model will be found. The
question now is whether there exist sub-optimal algorithms for more
complicated model classes. The same problem arises in case dependency
learning, for which I have restricted the class of case frame models
to that of dependency forest models. It would be more appropriate,
however, to restrict the class to, for example, the class of normal
Bayesian Networks.  How to learn such a complicated model, then,
needs further investigation.
\item[Unified model] 
I have divided the problem of learning lexical
  knowledge into three subproblems for easy examination. It would be
  more appropriate to define a single unified model. How
to define such a model, as well as how to learn it, are issues for
  future research. (See \cite{Miyata97,Utsuro97} for some recent
progress on this issue; see also discussions in Chapter 3.)
\item[Combination with extraction] We have seen that the amount of
data currently available is generally far less than that necessary for
accurate learning, and the problem of how to collect sufficient data
may be expected to continue to be a crucial issue. One solution might
be to employ bootstrapping, i.e., to conduct extraction and
generalization, iteratively. How to combine the two processes needs
further examination.
\item[Combination with word sense disambiguation] I have not addressed
the word sense ambiguity problem in this thesis, simply proposing to
conduct word sense disambiguation in pre-processing. (See
\cite{McCarthy97} for her proposal on word sense disambiguation.) In
order to improve the disambiguation results, however, it would be
better to employ the soft case slot model to perform structural and
word sense disambiguation at the same time. How to effectively learn
such a model requires further work.
\item[Soft clustering] I have formalized the problem of 
constructing a thesaurus into that of learning a double mixture model.
How to efficiently learn such a model is still an open problem.
\item[Parsing model] 
The use of lexical knowledge alone in
disambiguation might result in the resolving of most of the
ambiguities in sentence parsing, but not all of them. As has been described,
one solution to
the problem might be to define a unified model combining both lexical
knowledge and syntactic knowledge.
The problem still requires further work.
\end{description}

\newpage
\addcontentsline{toc}{chapter}{\numberline{}References}

\appendix

\chapter{}

\section{Derivation of Description Length: Two-stage Code}\label{append:mdl}

We consider here 
\begin{equation}\label{eq:minimization}
\begin{array}{l}
  \min_{\delta} \left( \log \frac{V}{\delta_1\cdots\delta_k} - \log
  P_{\tilde{\theta}}(x^n) \right). \\
\end{array}
\end{equation}

We first make Taylor's expansion of $- \log P_{\tilde{\theta}}(x^n)$
around $\hat{\theta}$:
\[
\begin{array}{l}
  - \log P_{\tilde{\theta}}(x^n) = - \log
  P_{\hat{\theta}}(x^n) + \frac{\partial (-\log
    P_{\theta}(x^n))}{\partial \theta}|_{\hat{\theta}} \cdot
  \delta \\ + \frac{1}{2} \cdot 
  \delta^T \cdot
  \left \{ \frac{\partial^2 (-\log P_{\theta}(x^n))}{\partial
    \theta^2}|_{\hat{\theta}} \right \} \cdot
  \delta + O(n \cdot \delta^3),
\end{array}
\] where $\delta^T$ denotes a transpose of $\delta$. The second term
equals 0 because $\hat{\theta}$ is the MLE estimate, and we ignore the fourth
term. Furthermore, the third term can be written as
\[ \frac{1}{2} \cdot \log e \cdot n \cdot \delta^T \cdot 
\left \{
\frac{\partial^2 (- \frac{1}{n} \cdot \ln P_{\theta}(x^n))}{\partial
  \theta^2}|_{\hat{\theta}} \right \} 
\cdot \delta,
\] where `$\ln$' denotes the natural logarithm (recall that `$\log$'
denotes the logarithm to the base 2). Under certain suitable conditions, when
$n \rightarrow \infty$, $\left \{ \frac{\partial^2 (- \frac{1}{n} \ln
  P_{\theta}(x^n))}{\partial \theta^2}|_{\hat{\theta}} \right \}$ can
be approximated as a $k$-dimensional matrix of constants $I(\theta)$
known as the `Fisher information matrix.'

Let us next consider
\[
\min_{\delta} 
\left( 
\log \frac{V}{\delta_1\cdots\delta_k}
- \log P_{\hat{\theta}}(x^n) 
+ \frac{1}{2} \cdot \log e \cdot n \cdot \delta^T \cdot
I(\theta) \cdot \delta
\right).
\] 

Differentiating this formula with each $\delta_i$ and having the
results equal 0, we obtain the following equations:
\begin{equation}\label{eq:delta}
(n \cdot I(\theta) \cdot \delta)_i - \frac{1}{\delta_i} = 0, \ \ \ \ (i=1,\cdots,k).
\end{equation}
Suppose that the eigenvalues of $I(\theta)$ are $\lambda_1.\cdots,
\lambda_k$, and the eigenvectors are $(u_1,\cdots,u_k)$. If we
consider only the case in which the axes of a cell ($k$-dimensional
rectangular solid) in the discretized vector space are in parallel
with $(u_1,\cdots,u_k)$, then (\ref{eq:delta}) becomes \[ n \cdot
\left ( \begin{array}{lll} \lambda_1 & & 0 \\
 & \ddots & \\
0 & & \lambda_k \\
\end{array}
\right )
\left (
\begin{array}{l}
\delta_1 \\
\vdots \\
\delta_k \\
\end{array}
\right )
= 
\left (
\begin{array}{l}
\frac{1}{\delta_1} \\
\vdots \\
\frac{1}{\delta_k} \\
\end{array}
\right ).
\]
Hence, we have
\[
\delta_i = \frac{1}{\sqrt{n\cdot \lambda_i}}
\]
and
\[ n \cdot \delta^T \cdot I(\theta) \cdot \delta = k.
\] Moreover, since $\lambda_1\cdots\lambda_k = |I(\theta)|$ where
`$|{\bf A}|$' stands for the determinant of ${\bf A}$, we have
\[
\frac{1}{\delta_1\cdots\delta_k} = {\sqrt n}^{k} \cdot \sqrt{
|I(\theta)|}.
\] 

Finally, (\ref{eq:minimization}) becomes
\[
\begin{array}{l}
\min_{\delta} \left( \log \frac{V}{\delta_1\cdots\delta_k} - \log
  P_{\tilde{\theta}}(x^n) \right) \\
\approx \log ( V \cdot {\sqrt n}^{k} \cdot \sqrt{|I(\theta)|} ) - \log
P_{\hat{\theta}}(x^n) + \frac{1}{2}\cdot\log e \cdot k + O(\frac{1}{\sqrt{n}}) \\
= -\log P_{\hat{\theta}}(x^n) + \frac{k}{2} \cdot \log n 
+ \frac{k}{2} \cdot \log e + \log V + \frac{1}{2} \cdot \log (|I(\theta)|) + O(\frac{1}{\sqrt{n}}) \\
= -\log P_{\hat{\theta}}(x^n) + \frac{k}{2} \cdot \log n + O(1).
\end{array}
\]

\section{Learning a Soft Case Slot Model}\label{append:fmm}

I describe here a method of learning the soft case slot model defined
in (\ref{eq:soft-csmodel}).

We can first adopt an existing soft clustering of words and estimate
the word probability distribution $P(n|C)$ within each class by
employing a heuristic method (cf., \cite{Li97b}). We can next estimate
the coefficients (parameters) $P(C|v,r)$ in the finite mixture model.

There are two common methods for statistical estimation, Maximum
Likelihood Estimation and Bayes Estimation. In their implementation
for estimating the above coefficients, however, both of
them suffer from computational intractability. The EM algorithm
\cite{Dempster77} can be used to approximate the maximum likelihood
estimates of the coefficients. The Markov Chain Monte-Carlo technique
\cite{Hastings70,Geman84,Tanner87,Gelfand90} can be used to
approximate the Bayes estimates of the coefficients.

We consider here the use of an extended version of the EM algorithm
\cite{Helmbold95}. For the sake of notational simplicity, for some
fixed $v$ and $r$, let us write $P(C|v,r), C \in \Gamma$ as
$\theta_{i} (i=1,\cdots,m)$ and $P(n|C)$ as $P_{i}(n)$. Then the
soft case slot model in (\ref{eq:soft-csmodel}) may be written as \[
  P(n|v,r)=\sum ^{m}_{i=1}\theta_{i}\cdot P_{i}(n).
\]
Letting $\theta =(\theta_{1},\cdots,\theta_{m})$, for a given training
sequence $n_{1}\cdots n_{S},$ the maximum likelihood estimate of
$\theta$, denoted as $\hat{\theta}$, is defined as the value that
maximizes the following log likelihood function
\[
L(\theta) = \frac{1}{S} \sum_{t=1}^{S} \log \left(\sum_{i=1}^{m}\theta_i \cdot P_i(n_t) \right).
\]

The EM algorithm first arbitrarily sets the initial value of $\theta$,
which we denote as $\theta^{(0)}$, and then successively calculates
the values of $\theta$ on the basis of its most recent values. Let $s$
be a predetermined number. At the $l$th iteration ($l=1,\cdots,s$), we
calculate $\theta^{(l)}=(\theta^{(l)}_{1},\cdots,\theta^{(l)}_m)$ by
\[
\theta^{(l)}_{i} = \theta^{(l-1)}_i \left(\eta (\bigtriangledown
L(\theta^{(l-1)})_i-1) + 1\right),
\]
where $\eta > 0$ (when $\eta=1$, Helmbold et al's version simply
becomes the standard EM algorithm), and $\bigtriangledown L(\theta)$
denotes
\[
  \bigtriangledown L(\theta) = \left( \frac{\partial L}{\partial \theta_1} \cdots \frac{\partial L}{\partial \theta_m} \right).
\]
After $s$ numbers of calculations, the EM algorithm outputs
$\theta^{(s)}=(\theta^{(s)}_{1},\cdots,\theta^{(s)}_m)$ as an
approximate of $\hat{\theta}$. It is theoretically guaranteed that the
EM algorithm converges to a {\em local} maximum of the likelihood
function \cite{Dempster77}.

\section{Number of Tree Cuts}\label{append:cut}

If we write $F(i)$ for the number of tree cuts in a complete $b$-ary
tree of depth $i$, we can show by mathematical induction that the
number of tree cuts in a complete $b$-ary tree of depth $d$ satisfies
\[ F(d) = \Theta(2^{b^{d-1}}), \]since
clearly \[ F(1) = 1 + 1 \] and \[ F(i) = (F(i-1))^b + 1 \ \ \ \
(i=2,\cdots,d).\]

Since the number of leaf nodes $N$ in a complete $b$-ary tree equals
$b^d$, we conclude that the number of tree cuts in a complete $b$-ary
tree is of order $\Theta(2^{\frac{N}{b}})$.

\begin{figure}[htb] \begin{center}
\epsfxsize5cm\epsfysize5cm\epsfbox{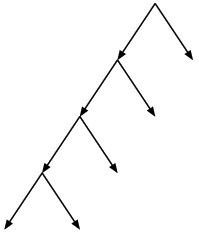}
\end{center}
\caption{An example one-way branching binary tree.} \label{fig:owb-tree}
\end{figure}

Note that the number of tree cuts in a tree depends on the structure
of that tree. If a tree is what I call a `one-way branching $b$-ary
tree,' then it is easy to verify that the number of tree cuts in that
tree is only of order
$\Theta(\frac{N-1}{b-1})$. Figure~\ref{fig:owb-tree} shows an example
one-way branching $b$-ary tree.  Note that a thesaurus tree is an
unordered tree (for the definition of an unordered tree, see, for
example, \cite{Knuth73}).

\section{Proof of Proposition 1}\label{append:prop1}

For an arbitrary subtree $T'$ of a thesaurus tree $T$ and an arbitrary
tree cut model $M=(\Gamma,\theta)$ in $T$, let
$M_{T'}=(\Gamma_{T'},\theta_{T'})$ denote the submodel of $M$ that is
contained in $T'$. Also, for any sample $S$, let $S_{T'}$ denote the
subsample of $S$ contained in $T'$. (Note that $M_T= M$, $S_T=S$.) 
Then, in general for any submodel $M_{T'}$ and subsample $S_{T'}$,
define $L(S_{T'}|\Gamma_{T'},\hat{\theta}_{T'})$ to be the data
description length of subsample $S_{T'}$ using submodel $M_{T'}$,
define $L(\hat{\theta}_{T'}|\Gamma_{T'})$ to be the parameter
description length for the submodel $M_{T'}$, and define $L'(M_{T'},
S_{T'})$ to be $L(S_{T'}|\Gamma_{T'},\hat{\theta}_{T'}) +
L(\hat{\theta}_{T'}|\Gamma_{T'})$.

First for any (sub)tree $T$, for any (sub)model
$M_T=(\Gamma_T,\hat{\theta}_T)$ which is contained in $T$ except the
(sub)model consisting only of the root node of $T$, and for any
(sub)sample $S_T$ contained in $T$, we have
\begin{equation}
\label{eq:prov1} L(S_T|\Gamma_T,\hat{\theta}_T) = \sum_{i=1}^{k}
L(S_{T_i}|\Gamma_{T_i},\hat{\theta}_{T_i}), \end{equation} where $T_i,
(i=1,\cdots,k)$ denote the child subtrees of $T$.

For any (sub)tree $T$, for any (sub)model
$M_T=(\Gamma_T,\hat{\theta}_T)$ which is contained in $T$ except the
(sub)model consisting only of the root node of $T$, we have
\begin{equation} \label{eq:prov2} L(\hat{\theta}_T|\Gamma_T) =
\sum_{i=1}^{k} L(\hat{\theta}_{T_i}|\Gamma_{T_i}), \end{equation}
where $T_i, (i=1,\cdots,k)$ denote the child subtrees of $T$.

When $T$ is the entire thesaurus tree, the parameter description
length for a tree cut model in $T$ should be
\begin{equation}\label{eq:prov3} L(\hat{\theta}_T|\Gamma_T) =
\sum_{i=1}^{k} L(\hat{\theta}_{T_i}|\Gamma_{T_i}) - \frac{\log
|S|}{2}, \end{equation} where $|S|$ is the size of the entire
sample. Since the second term $- \frac{\log |S|}{2}$ in
(\ref{eq:prov3}) is common to each model in the entire thesaurus tree,
it is irrelevant for the purpose of finding a model with the minimum
description length.

We will thus use identity (\ref{eq:prov2}) both when $T$ is a proper
subtree and when it is the entire tree. (This allows us to use the
same recursive algorithm (Find-MDL) in all cases.)

It follows from (\ref{eq:prov1}) and (\ref{eq:prov2}) that the
minimization of description length can be done essentially
independently for each subtree. Namely, if we let $L'_{min}(M_T,S_T)$
denote the minimum description length (as defined by (\ref{eq:prov1})
and (\ref{eq:prov2})) achievable for (sub)model $M_T$ on (sub)sample
$S_T$ contained in (sub)tree $T$, $\hat{P}(\eta)$ the MLE estimate of
the probability for node $\eta$, and ${\rm root}(T)$ the root node of
$T$, then we have
\begin{equation}\label{eq:prov4}
\begin{array}{l}
  L'_{min}(M_T,S_T) = \min \{ \sum_{i=1}^{k} L'_{min}(M_{T_i}, S_{T_i}),
L'(([{\rm root}(T)], [\hat{P}({\rm root}(T))]), S_T)\}. \\ 
\end{array}
\end{equation}
Here, $T_i, (i=1,\cdots,k)$ denote the child subtrees of $T$.

The rest of the proof proceeds by induction. First, if $T$ is a
subtree having a single node, then there is only one submodel in $T$,
and it is clearly the submodel with the minimum description
length. Next, inductively assume that {\rm Find-MDL}$(T')$ correctly
outputs a submodel with the minimum description length for any subtree
$T'$ of size less than $n$.  Then, given a (sub)tree $T$ of size $n$
whose root node has at least two child subtrees, say $T_i : i=1,\cdots,k$,
for each $T_i$, {\rm
  Find-MDL}$(T_i)$ returns a submodel with the minimum description
length by inductive hypothesis. Then, since (\ref{eq:prov4}) holds, in
whichever way the if-clause on lines 8, 9 of {\rm Find-MDL} is
evaluated, what is returned on line 11 or line 13 will still be a
(sub)model with the minimum description length, completing the
inductive step.

It is easy to see that the time complexity of the algorithm is linear
in the number of leaf nodes of the input thesaurus tree.

\section{Equivalent Dependency Tree Models}\label{append:dend}

We prove here that the dependency tree models based on a labeled free
tree are equivalent to one another. Here, a labeled free tree means a
tree in which each node is associated with one unique label and in
which any node can be the root.

\begin{figure}[htb] \begin{center}
\epsfxsize12cm\epsfysize5cm\epsfbox{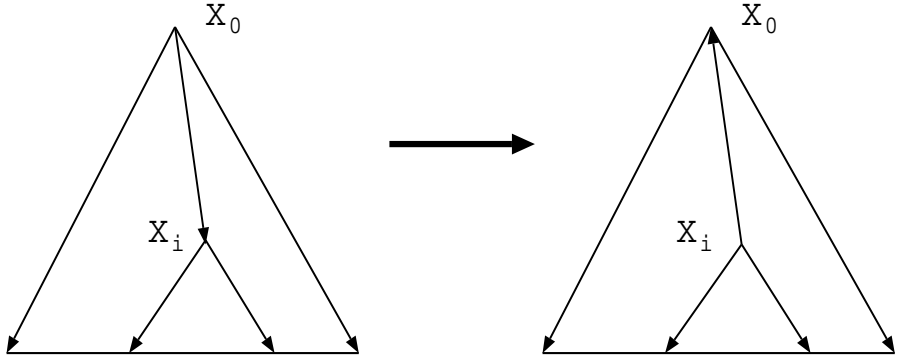}
\end{center}
\caption{Equivalent dependency tree models.} \label{fig:free-tree}
\end{figure}

Suppose that the free tree we have is now rooted at $X_0$
(Figure~\ref{fig:free-tree}). The dependency tree model based on this
rooted tree will then be uniquely determined. Suppose that we randomly
select one other node $X_i$ from this tree. If we reverse the
directions of the links from $X_0$ to $X_i$, we will obtain another
tree rooted at $X_i$. Another dependency tree model based on this tree
will also be determined. It is not difficult to see that these two
distributions are equivalent to one another, since \[
\begin{array}{ll}
 & P(X_0) \cdot P(X_1|X_0) \cdots P(X_i|X_{i-1}) \\
= & P(X_0|X_1) \cdot P(X_1) \cdots P(X_i|X_{i-1}) \\
= & \cdots \\
= & P(X_0|X_1) \cdots P(X_{i-1}|X_i) \cdot P(X_i). \\
\end{array}
\]
Thus we complete the proof.

\section{Proof of Proposition 2}\label{append:prop2}

We can represent any dependency forest model as
\[
\begin{array}{c}
P(X_1,\cdots,X_n) = P(X_1|X_{q(1)}) \cdots P(X_i|X_{q(i)}) \cdots P(X_n|X_{q(n)}) \\
0 \le q(i) \le n, q(i)\not=i, (i=1,\cdots,n)\\
\end{array}
\]
where $X_{q(i)}$ denotes a random variable which $X_i$ depends on. We
let $P(X_i|X_0) = P(X_i)$. Note that there exists a $j,
(j=1,\cdots,n)$ for which $P(X_j|X_{q(j)})=P(X_j|X_0)=P(X_j)$.

The sum of parameter description length and data description length
for any dependency forest model equals
\[
\begin{array}{l}
\sum_{i=1}^{n} \frac{k_i-1}{2} k_{q(i)} \cdot\log N - \sum_{x_1,\cdots,x_n} f(x_1,\cdots,x_n) \cdot\log \biggl(\hat{P}(x_1|x_{q(1)}) \cdots \hat{P}(x_i|x_{q(i)}) \cdots \hat{P}(x_n|x_{q(n)})\biggr) \\
= \sum_{i=1}^{n} \frac{k_i-1}{2} k_{q(i)} \cdot\log N - \sum_{i=1}^{n} \sum_{x_i,x_{q(i)}} f(x_i,x_{q(i)}) \cdot\log \hat{P}(x_i|x_{q(i)}) \\
= \sum_{i=1}^{n} \biggl(\frac{k_i-1}{2} k_{q(i)} \cdot\log N - \sum_{x_i,x_{q(i)}} f(x_i,x_{q(i)}) \cdot\log \hat{P}(x_i|x_{q(i)})\biggr),
\end{array}
\]
where $N$ denotes data size, $x_i$ the possible values of $X_i$, and
$k_i$ the number of possible values of $X_i$. We let $k_0 = 1$ and
$f(x_i,x_0)=f(x_i)$.

Furthermore, 
the sum of parameter description length and data description length
for the independent model (i.e., $P(X_1,\cdots,X_n)=\prod_{i=1}^n P(X_i)$) 
equals
\[
\begin{array}{l}
\sum_{i=1}^{n} \frac{k_i-1}{2} \cdot\log N - \sum_{x_1,\cdots,x_n} f(x_1,\cdots,x_n) \cdot\log \biggl( \prod_{i=1}^n \hat{P}(x_i)\biggr) \\
= \sum_{i=1}^{n} \frac{k_i-1}{2} \cdot\log N - \sum_{i=1}^{n} \sum_{x_i} f(x_i) \cdot\log \hat{P}(x_i) \\
= \sum_{i=1}^{n} \biggl(\frac{k_i-1}{2} \cdot\log N - \sum_{x_i} f(x_i) \cdot\log \hat{P}(x_i)\biggr).
\end{array}
\]

Thus, the difference between the description length of the independent model
and the description length of any dependency forest model becomes
\[
\begin{array}{l}
\sum_{i=1}^n \biggl( \sum_{x_i,x_q(i)} f(x_i,x_{q(i)}) \cdot
(\log \hat{P}(x_i|x_{q(i)}) - \log \hat{P}(x_i) ) - \frac{(k_i-1)\cdot (k_{q(i)}-1)}{2} \cdot\log N \biggr) \\
= \sum_{i=1}^n \biggl( N \cdot \hat{I}(X_i,X_{q(i)}) - \frac{(k_i-1)\cdot (k_{q(i)}-1)}{2} \cdot\log N \biggr) \\
= \sum_{i=1}^{n} N \cdot \theta(X_i,X_{q(i)}).
\end{array}
\]

Any dependency forest model for which there exists an $i$ satisfying
$\theta(X_i,X_{q(i)}) < 0$ is not favorable from the viewpoint of MDL
because the model for which the corresponding $i$ satisfying
$\theta(X_i,X_{q(i)}) = 0$ always exists and is clearly more favorable.

We thus need only select the MDL model from those models for which for
any $i$, $\theta(X_i,X_{q(i)}) \ge 0$ is satisfied. Obviously, the
model for which $\sum_{i=1}^{n} \theta(X_i,X_{q(i)})$ is maximized is
the best model in terms of MDL. What Suzuki's algorithm outputs is
exactly this model, and this completes the proof.

\newpage
\addcontentsline{toc}{chapter}{\numberline{}Publication List}

\chapter*{Publication List}

\section*{Reviewed Journal Papers}

\begin{enumerate}

\item Li, H.: A Probabilistic Disambiguation Method based on
  Psycholinguistic Principles, (in Japanese) {\em Computer Software},
  Vol.13, No. 6, (1996) pp.~53--65.

\item
  Li, H. and Abe, N.: Clustering Words with the MDL Principle, {\em
    Journal of Natural Language Processing}, Vol.4, No.  2, (1997),
  pp.~71--88.

\item
Li, H. and Abe, N.: Generalizing Case Frames Using a Thesaurus and the MDL
  Principle, {\em Computational Linguistics}, Vol.24, No.2
(1998), pp.~217-244.

\end{enumerate}

\section*{Reviewed Conference Papers}

\begin{enumerate}

\item
Li, H. and Abe, N.: Generalizing Case Frames Using a Thesaurus and the MDL
  Principle, {\em Proceedings of Recent Advances in Natural Language
  Processing}, (1995), pp.~239--248.

\item Abe, N. and Li, H.: On-line Learning of Binary Lexical Relations
  Using Two-dimensional Weighted Majority Algorithms, {\em Proceedings
    of the 12th International Conference on Machine Learning
    (ICML'95)}, (1995), pp.~71--88.

\item
Li, H.: A Probabilistic Disambiguation Method based on Psycholinguistic
  Principles, {\em Proceedings of the 4th Workshop on Very Large Corpora},
  (1996), pp.~141--154.

\item Li, H. and Abe, N.: Clustering Words with the MDL Principle,
  {\em Proceedings of the 16th International Conference on
    Computational Linguistics (COLING'96)}, (1996), pp.~4--9.

\item
Li, H. and Abe, N.: Learning Dependencies between Case Frame Slots, {\em
  Proceedings of the 16th International Conference on Computational
  Linguistics (COLING'96)}, (1996), pp.~10--15.

\item
  Abe, N. and Li, H.:Learning Word Association Norms Using Tree Cut
  Pair Models , {\em Proceedings of the 13th International Conference
    on Machine Learning (ICML'96)}, (1996),
  pp.~71--88.

\item
Li, H. and Yamanishi, K.: Document Classification Using a Finite Mixture Model,
  {\em Proceedings of the 35th Annual Meeting of Association for Computational
  Linguistics (ACL/EACL'97) }, (1997).

\item Li, H. and Abe, N.: Word Clustering and Disambiguation Based on
Co-occurrence Data, {\em
  Proceedings of the 18th International Conference on Computational
  Linguistics and the 36th Annual Meeting of Association for
Computational Linguistics (COLING-ACL'98)}, (1998), to appear.

\end{enumerate}

\end{document}